\documentclass{article}

\usepackage{microtype}
\usepackage{graphicx}
\usepackage{subcaption}
\usepackage{makecell}

\usepackage{booktabs} % for professional tables

\usepackage{hyperref}

\usepackage[accepted]{icml2026}

\usepackage{amsmath}
\usepackage{amssymb}
\usepackage{mathtools}
\usepackage{amsthm}

\usepackage[capitalize,noabbrev]{cleveref}

\usepackage{tabularx} 

\usepackage{enumitem}

\newcolumntype{Y}{>{\centering\arraybackslash}X}

\theoremstyle{plain}

\theoremstyle{definition}

\theoremstyle{remark}

\usepackage[textsize=tiny]{todonotes}

\icmltitlerunning{Predicting Future Utility: Global Combinatorial Optimization for Task-Agnostic KV Cache Eviction}

\usepackage{booktabs}
\usepackage{multirow}
\usepackage{graphicx}

\usepackage{fontawesome5}

\usepackage{xurl}
\usepackage[table]{xcolor} % For \rowcolor

\definecolor{color_ours}{RGB}{210, 225, 215}
\newcommand{\camblue}[1]{\textcolor{black}{#1}}

\begin{document}

\twocolumn[
  \icmltitle{Predicting Future Utility: Global Combinatorial Optimization \\ for Task-Agnostic KV Cache Eviction}



  \icmlsetsymbol{equal}{*}

  \begin{icmlauthorlist}
    % \icmlauthor{Ziyao Tang}{equal,fudan,baidu,intern}
    \icmlauthor{Ziyao Tang}{equal,fudan,intern}
    \icmlauthor{Pengkun Jiao}{equal,fudan}
    \icmlauthor{Xinhang Chen}{baidu}
    \icmlauthor{Wei Liu}{baidu}
    \icmlauthor{Shiyong Li}{baidu}
    \icmlauthor{Jingjing Chen}{fudan}

  \end{icmlauthorlist}

  \icmlaffiliation{fudan}{Fudan University}
  \icmlaffiliation{baidu}{Baige AI Team, Baidu inc}
  \icmlaffiliation{intern}{Work done during an internship at Baidu}

  \icmlcorrespondingauthor{Jingjing Chen}{chenjingjing@fudan.edu.cn}

  \icmlkeywords{Machine Learning, KV Cache, Optimization}

  \vskip 0.3in
]



\printAffiliationsAndNotice{\icmlEqualContribution}  % no special notice (required even if empty)

% \begin{abstract}

% Key-Value (KV) cache eviction is critical for mitigating memory bottlenecks during Large Language Model inference; however, existing cross-head budget allocation strategies typically rely on directly comparing proxy scores (e.g., attention weights) across different attention heads. We identify a significant ``Optimality Gap'' where high proxy scores do not consistently correspond to high Oracle Importance across heads, causing raw-score-based competition to misallocate resources away from truly informative regions. To address this, we propose the Metric-Alignment Profiling framework under the query-agnostic setting, which quantifies the marginal utility of specific metrics in retrieving ground-truth information via offline profiling and formulates allocation as a combinatorial optimization problem to maximize global utility. Extensive experiments demonstrate that our approach significantly improves compression performance across varying proxy metrics on both the LongBench (16 tasks) and RULER (13 tasks) benchmarks. Notably, under an 80\% compression ratio, our method boosts the accuracy of SnapKV on the challenging RULER multi-needle retrieval task from 1.00\% (under uniform allocation) to 67.40\%, substantially enhancing the robustness of existing compression methods in extreme scenarios.
% \end{abstract}

\begin{abstract}
Given the quadratic complexity of attention, KV cache eviction is vital to accelerate model inference.
Current KV cache eviction methods typically rely on instantaneous heuristic metrics, implicitly assuming that score magnitudes are consistent proxies for importance across all heads. 
However, this overlooks the heterogeneity in predictive fidelity across attention heads. While certain heads prioritize the \textit{instantaneous contribution} of tokens, others are dedicated to capturing \textit{long-horizon utility}.
In this paper, we propose that optimal budget allocation should be governed by the marginal utility in preserving long-term semantic information. 
Building on this insight, we propose \textbf{LU-KV}, a novel framework that formulates head-level budget allocation as a global combinatorial optimization problem to maximize the long-horizon marginal contribution of reserved tokens. To solve this non-convex problem, we employ a convex-hull relaxation and a marginal-utility-based greedy solver, achieving near-optimal solutions. Furthermore, we implement a data-driven offline profiling protocol to facilitate the practical deployment of LU-KV.
Evaluations on LongBench and RULER benchmarks demonstrate that LU-KV reduces KV cache size by 80\% with minimal performance degradation, while also decreasing inference latency and GPU memory footprint.
% Extensive evaluations on LongBench and RULER benchmarks demonstrate that \textbf{LU-KV} achieves an 80\% reduction in KV cache size with minimal performance degradation, while simultaneously reducing inference latency and GPU memory footprint.
\small \faGithub~\href{https://github.com/baidu-baige/LU-KV}{Project Code}
\end{abstract}

\section{Introduction}
\label{sec:intro}

% The advent of Large Language Models (LLMs) has revolutionized long-context processing; however, the Key-Value (KV) cache presents a formidable bottleneck due to the  quadratic complexity of attentio . As sequence lengths reach million-token scales, the linear growth of cache memory limits inference throughput and complicates scalable deployment. To address this, KV cache eviction~\cite{zhenyu_h2o_2023,feng_adakv_2024,jang_kvzip_2025} has become a standard necessity, traditionally operating via a two-stage paradigm: \textit{intra-head scoring} to identify critical tokens and \textit{cross-head budget allocation} to distribute available storage across the model's architecture.

% \textcolor{black}{
The recent push towards long-context processing in Large Language Models (LLMs) is fundamentally bottlenecked by the Key-Value (KV) cache. As sequence lengths scale to millions of tokens, the linear explosion in memory consumption severely throttles inference throughput and hinders scalable deployment. Consequently, KV cache eviction~\cite{zhenyu_h2o_2023,feng_adakv_2024,jang_kvzip_2025} has emerged as an essential technique. It conventionally operates under a two-stage paradigm: \textit{intra-head scoring} to filter critical tokens, and \textit{cross-head budget allocation} to distribute the memory budget across attention heads.
% }

While significant progress has been made in designing sophisticated scoring metrics \citep{zhenyu_h2o_2023, yuhong_SnapKV_2024}, budget allocation strategies remain a critical yet underdeveloped frontier. Existing methods largely rely on instantaneous heuristic scoring, assuming that current attention magnitudes serve as reliable proxies for future importance. However, we identify a fundamental flaw in this magnitude-based paradigm: it ignores the inherent heterogeneity in predictive fidelity across different attention heads. Specifically, high-magnitude scores in certain heads often align poorly with \textbf{Oracle Importance}---the true long-term contribution to the KV cache---capturing transient noise rather than enduring semantic anchors. Blindly biasing budgets toward regions with high-magnitude heuristic scores, without accounting for their long-term utility, inevitably leads to suboptimal cache retention.

% While significant progress has been made in designing sophisticated scoring metrics \citep{zhenyu_h2o_2023, yuhong_SnapKV_2024}, budget allocation strategies remain a critical yet underdeveloped frontier. Existing methods largely rely on instantaneous heuristic scoring, assuming that current attention magnitudes serve as reliable proxies for future importance. However, we identify a fundamental flaw in this magnitude-based paradigm: it ignores the inherent heterogeneity in predictive fidelity across different attention heads. Specifically, high-magnitude scores in certain heads often align poorly with \textbf{Oracle Importance}---the true long-term contribution to the KV cache---capturing transient noise rather than enduring semantic anchors. Blindly biasing budgets toward regions with high-magnitude heuristic scores, without accounting for their long-term utility, inevitably leads to suboptimal cache retention.

We posit that optimal budget allocation should be governed not by absolute scores, but by the marginal utility of a metric in preserving future information. In this view, memory allocation is treated as a strategic investment: if a metric exhibits poor alignment with the Oracle Importance in a specific head, increasing its budget yields rapidly diminishing returns. Conversely, in heads where the metric is precise, a unit investment in budget effectively preserves the model's long-horizon generative quality. 
\textcolor{black}{Therefore, the crux of an efficient allocation strategy lies in quantifying and optimizing the long-term cost-effectiveness of each head as guided by a given metric.}
% Therefore, the crux of an efficient allocation strategy lies in quantifying and optimizing the long-term cost-benefit ratio of each head under a specific metric.

\textcolor{black}{Based on this insight, we propose \textbf{Long-horizon Utility KV (LU-KV)}, a novel framework for head-wise KV cache budget allocation. We introduce a data-driven offline calibration mechanism to profile the marginal contribution curves of individual attention heads. To construct these profiles, we formulate the global budget distribution as a combinatorial optimization problem aimed at maximizing the expected long-horizon utility retention across all heads. To solve this efficiently, we employ a convex-hull relaxation and a greedy solver, ensuring near-optimal budget allocation with minimal computational overhead.}

% Based on this insight, we propose \textbf{Long-horizon Utility KV (LU-KV)}, a novel framework for head-wise KV cache budget allocation. We introduce a data-driven offline calibration mechanism to profile the marginal contribution curves of individual attention heads. Based on this, we formulate the global budget distribution as a combinatorial optimization problem aimed at maximizing expected long-horizon utility retention across all heads. To solve this efficiently, we employ a convex-hull relaxation and a greedy solver, ensuring near-optimal budget allocation with minimal computational overhead.

% Based on this insight, we propose \textbf{L}ong-horizon \textbf{U}tility \textbf{KV} \textbf{(LU-KV)}, a novel framework for head-wise KV cache budget allocation. We introduce a data-driven offline calibration mechanism to profile the \textbf{marginal contribution curves} of individual heads. By simulating the long-horizon decoding process, we formulate the problem of how the global budget translates into the expected retention of long-horizon utility whthin each heads in a combinatorial optimization problem. To resolve the resulting combinatorial optimization problem, we employ a \textbf{convex-hull relaxation} and a greedy solver, ensuring near-optimal budget distribution with minimal computational overhead.

Extensive experiments on the LongBench and RULER benchmarks demonstrate the effectiveness of LU-KV. Our method achieves an 80\% reduction in KV cache size with minimal performance degradation.
% , while simultaneously reducing inference latency and GPU memory footprint.

Our contributions are summarized as follows:
\begin{itemize}[leftmargin=*, nosep]
    \item \textcolor{black}{We identify a critical gap between heuristic importance metrics and long-horizon marginal utility in head-wise KV eviction, as these metrics fail to reflect the true long-horizon marginal contribution of tokens in certain heads.}
    \item \textcolor{black}{We formulate budget allocation as a long-term utility maximization problem and introduce an efficient solver using convex-hull relaxation and a marginal-utility-based greedy strategy. Furthermore, we propose an offline profiling protocol for practical deployment.}
    \item We conduct extensive evaluations across diverse long-context benchmarks to validate the effectiveness and robustness of our proposed methods.

    % \item We define the \textbf{Optimality Gap} between heuristic metrics and long-horizon utility importance, demonstrating that heuristic score magnitudes are insufficient for cross-head budget allocation.
    % \item We formulate budget allocation as a long-term utility maximization problem and introduce an efficient solver using convex-hull relaxation and marginal-utility-based greedy allocation.
    % \item We propose an offline profiling protocol that leverages the structural stability of LLMs, enabling zero-overhead online execution via a pre-computed lookup table.
    % \item We conduct extensive evaluations across diverse long-context benchmarks to validate the effectiveness and robustness of our proposed methods.
\end{itemize}
\normalcolor

\section{Related Work}
\label{sec:related_work}

Existing research on KV cache eviction can be broadly categorized into two synergistic streams: \textit{intra-head eviction policies}, which identify informative tokens within individual heads, and \textit{cross-head budget allocation}, which manages resource distribution across the entire model.

\paragraph{Intra-head Eviction Policies}
\label{subsec:intra_head_policies}

Intra-head strategies focus on designing high-fidelity proxy metrics to distinguish critical tokens from noise. Early heuristics, such as StreamingLLM \citep{guang_streamingllm_2024}, identified the ``attention sink'' phenomenon, showing that retaining initial tokens is crucial for maintaining model stability. Subsequently, methods like H2O \citep{zhenyu_h2o_2023} and SnapKV \citep{yuhong_SnapKV_2024} utilized accumulated attention scores or observation windows to dynamically cluster and retain salient tokens. Beyond raw attention weights, recent literature has explored geometric and perturbation-based indicators to mitigate inherent biases. For instance, KeyDiff \citep{jun_keydiff_2025} leverages the geometric features of Key vectors, while CriticalKV \citep{feng_criticalkv_2025} explicitly measures potential output perturbations by considering Value magnitudes and projection weights. \camblue{DefensiveKV~\citep{feng_defensivekv_2026} adopts a worst-case view for robust eviction by aggregating scores within individual heads.} KVZip
\citep{jang_kvzip_2025} investigates query-agnostic KV cache
compression via context reconstruction, using a reconstruction
objective to guide cache eviction without relying on the current
query.  
Despite these advances, existing methods primarily optimize
token selection within a fixed per-head budget, overlooking how that budget should be allocated across heads.

\paragraph{Cross-Head Budget Allocation}
\label{subsec:cross_head_allocation}
Recognizing the heterogeneity of information density across layers, recent studies have shifted toward non-uniform allocation strategies. Static and rule-based methods often rely on structural priors; for example, PyramidKV \citep{cai_pyramidkv_2024} employs a fixed pyramidal shape based on the ``information funneling'' hypothesis, while HeadKV \citep{Fu_headkv_2025} and CAKE \citep{Qin_cakekv_2025} incorporate task-specific priors or spatial dispersion to formulate cascading rules. In contrast, dynamic allocation strategies like Ada-KV \citep{feng_adakv_2024} attempt to distribute resources based on real-time statistics, such as attention entropy. However, these approaches inherently assume that proxy scores are well-calibrated and comparable across different heads—an assumption that often fails in practice due to varying score scales and metric inaccuracies.

% \paragraph{Relation to our work.} Unlike previous methods that focus on either designing more accurate metrics or estimating head importance via raw statistics, we propose an allocation framework specifically designed to capture Long-Horizon Utility. Rather than directly comparing uncalibrated proxy scores, our approach profiles the \textbf{budget-loss curve} offline to explicitly quantify the \textbf{marginal utility} of a specific metric within each head. This grants our framework \textbf{metric-universality}: for any chosen proxy (e.g., SnapKV or KeyDiff), we can derive its unique alignment characteristics to solve for the theoretically optimal budget configuration. Consequently, our method minimizes the compound loss arising from both capacity constraints and the intrinsic fidelity gaps of the underlying compression metrics.

% \paragraph{Relation to our work.} 
\textcolor{black}{
Unlike previous methods that rely on instantaneous heuristic 
scoring and consequently overlook the long-horizon importance 
of tokens, we propose an allocation framework governed by 
long-horizon utility. Rather than directly comparing 
uncalibrated proxy scores, our approach profiles the 
budget-utility relationship to explicitly quantify the 
marginal gain of retaining tokens within each attention head. 
Furthermore, our framework is metric-agnostic: given any 
chosen proxy metric (\textit{e.g.}, SnapKV), we can derive the optimal 
budget configuration to maximize long-horizon information 
retention.}

\section{Preliminaries}
\label{sec:preliminaries}

We consider decoder-only LLMs with $L$ layers and $H$ attention heads per layer. Inference proceeds in two phases: parallel prefill and autoregressive decoding. 

\subsection{Attention Mechanism and KV Cache}
At decoding step $k$, for a fixed layer $\ell$ and head $h$, the model generates a query $\mathbf{q}_{\ell,h,k} \in \mathbb{R}^{d_h}$ to attend to historical keys $\mathbf{k}_{\ell,h,j}$ and values $\mathbf{v}_{\ell,h,j}$ ($j \le k$). The attention weights $A$ and the head output are computed as:
\begin{align}
    A_{\ell,h,k,j} &= \text{Softmax} \left( \frac{\mathbf{q}_{\ell,h,k}^\top \mathbf{k}_{\ell,h,j}}{\sqrt{d_h}} \right), \\
    \mathbf{o}_{\ell,k} &= \sum_{h=1}^{H} \left( \sum_{j=1}^{T} A_{\ell,h,k,j} \mathbf{v}_{\ell,h,j} \right) \mathbf{W}_{O}^{(\ell,h)}.
\end{align}
where $\mathbf{W}_{O}^{(\ell,h)} \in \mathbb{R}^{d_h \times d_{\text{model}}}$ is the head-specific output projection. To avoid redundant computation, previously computed $(\mathbf{k}, \mathbf{v})$ pairs are stored in a \textbf{KV Cache}. As the sequence length grows, the linear increase in cache size poses a significant memory bottleneck.

\subsection{KV Cache Eviction}
\label{subsec:kv_eviction}

\camblue{For readability, the main symbols used below are summarized before the formal setup.}
\begin{center}
\resizebox{0.48\textwidth}{!}{%
    \begin{tabular}{@{}>{\bfseries}l l@{}}
        \toprule
        Symbol & Description \\
        \midrule
        $L, H$ & Number of layers and number of attention heads per layer, respectively. \\
        $(\ell, h)$ & The $h$-th attention head in the $\ell$-th layer. \\
        $T$ & The total number of input tokens during the prefill phase. \\
        $\pi$ & The metric used to evaluate token importance. \\
        $\pi^*$ & The oracle metric of token importance. \\
        $\sigma$ & \camblue{Target compression ratio, i.e., the fraction of token positions evicted.} \\
        $B_{\text{total}}$ & \camblue{Global all-head memory budget; for length $T$, $B_{\text{total}}=(1-\sigma)L\times H\times T$.} \\
        $b_{\ell,h}$ & \camblue{Memory budget allocated to a specific attention head $(\ell, h)$.} \\
        $r_{\ell,h}$ & \camblue{Local head-level compression ratio, with $b_{\ell,h}=\lfloor(1-r_{\ell,h})T\rfloor$.} \\
        $\mathcal{M}_{\ell,h}^\pi$ & Token indices at head $(\ell, h)$, sorted by $\pi$ (descending). \\
        $\mathcal{M}_{\ell,h}^\pi(k)$ & \camblue{Retained subset containing the top-$k$ elements of $\mathcal{M}_{\ell,h}^\pi$.} \\
        \bottomrule
    \end{tabular}
}
\end{center}
To maintain a manageable memory footprint, \textit{KV Cache Eviction} strategies~\cite{feng_adakv_2024, jun_keydiff_2025} impose a fixed budget $B$ per head. These methods typically use attention scores as an importance proxy, retaining a subset of indices $\mathcal{I}_{\ell,h} \subset \{1, \dots, T\}$ such that $|\mathcal{I}_{\ell,h}| \le B$. The attention output is then approximated by re-normalizing weights over the retained set:
\begin{equation}
    \tilde{\mathbf{o}}_{\ell,h,k} = \sum_{j \in \mathcal{I}_{\ell,h}} \tilde{A}_{\ell,h,k,j} \mathbf{v}_{\ell,h,j}.
\end{equation}
Current policies often prioritize ``heavy hitters'' with the highest cumulative attention scores, assuming their dominance in preserving model performance.

\section{Methodology}
\label{sec:methodology}

\subsection{Long-horizon KV Cache Eviction}
\label{subsec:formulation}

KV Cache eviction inherently entails a risk of information loss. Traditional eviction methods (e.g., H2O, SnapKV) rely on \textit{instantaneous} attention weights, such as those calculated during the prefill stage. We term these methods \textbf{Heuristic Metric}. However, these methods overlook the potential for shifts in attention patterns during future decoding steps. 
To address this, we propose \textbf{L}ong-horizon \textbf{U}tility \textbf{KV} \textbf{(LU-KV)}, a framework that evaluates KV utility over extended sequences.
We formulate the cache eviction problem as a Global Combinatorial Optimization of long-horizon utility, which we solve to determine the optimal head-wise budget allocation.
% In reality, the true contribution of a token (i.e., its attention weight) is dynamic. 

\paragraph{Oracle Importance.}
\textcolor{black}{
To rigorously quantify the long-horizon marginal contribution 
of each token, we introduce the concept of \textbf{Oracle 
Importance}. 
Inspired by the output perturbation bound analysis in AdaKV~\citep{feng_adakv_2024} and the criticality definition in CriticalKV~\citep{feng_criticalkv_2025}, we define the oracle importance score $I_{\ell,h,j}$ of a cached position $j$ in head $(\ell,h)$ as its maximum potential contribution over a future decoding window:}
\begin{equation}
\label{eq:oracle_importance}
I_{\ell,h,j}
\triangleq
\max_{k\in\{1,\dots,K_{\max}\}}
\left(
A_{\ell,h,k,j}\cdot
\Bigl\|\mathbf{v}_{\ell,h,j}\mathbf{W}_{O}^{(\ell,h)}\Bigr\|
\right).
\end{equation}
This score captures the true utility of a token in a long-horizon view: whether it constitutes a major component of the output vector at any future step $k$.
Based on this, we theoretically construct an \textbf{Oracle Metric} ($\pi^*$) that yields the set $\mathcal{M}_{\ell,h}^{\pi^*}$, which perfectly aligns with the descending ranking of the oracle importance score $I_{\ell,h,:}$.
% \camblue{Because this oracle ranking depends on future queries, Oracle Importance is used only as an offline analytical upper bound and not as a deployable online scoring metric.}

\begin{figure}[t]
    \centering
    \includegraphics[width=0.85\linewidth]{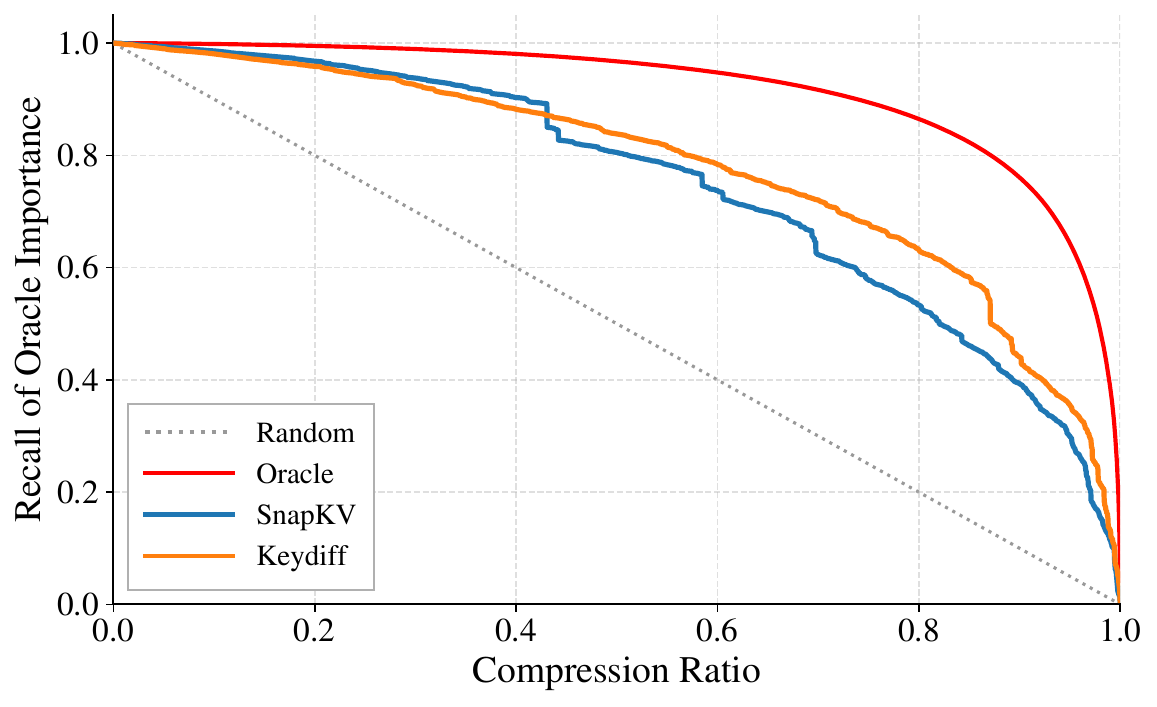}
    \caption{
    Recall of oracle importance for oracle metric and several heuristic metrics across varying compression ratios ($\sigma$) on Mistral-7B-v0.3, where $1$ implies full compression and $0$ implies no compression. 
    }
    \label{fig:misalignment_gap}
\end{figure}

\paragraph{Limitation of Heuristic Metric.}
\camblue{For a head budget $b_{\ell,h}$, $\mathcal{M}_{\ell,h}^{\pi}(b_{\ell,h})$ denotes the KV positions retained under metric $\pi$; positions outside this set are evicted.}
% Considering a head KV cache budget $b_{\ell,h}$.
In practice, $\mathcal{M}_{\ell,h}^{\pi}(b_{\ell,h})$ determined by a heuristic metric $\pi$ often deviates from the optimal set $\mathcal{M}_{\ell,h}^{\pi^*}(b_{\ell,h})$, due to the short-horizon of $\pi$. 

To analyze this discrepancy, we decompose the relationship between $\mathcal{M}_{\ell,h}^{\pi^*}$ and $\mathcal{M}_{\ell,h}^\pi$ into 3 classes:
\begin{itemize}[nosep]
    \item \textbf{Hits:} $\mathcal{M}_{\ell,h,\mathrm{hit}}=\mathcal{M}_{\ell,h}^{\pi^*}(b_{\ell,h})\cap \mathcal{M}_{\ell,h}^{\pi}(b_{\ell,h})$ (Correctly retained high oracle importance tokens)
    \item \textbf{Misses:} $\mathcal{M}_{\ell,h,\mathrm{miss}}=\mathcal{M}_{\ell,h}^{\pi^*}(b_{\ell,h})\setminus \mathcal{M}_{\ell,h}^{\pi}(b_{\ell,h})$ (High Oracle importance, wrongly evicted)
    \item \textbf{False Positives:} $\mathcal{M}_{\ell,h,\mathrm{fp}}=\mathcal{M}_{\ell,h}^{\pi}(b_{\ell,h})\setminus \mathcal{M}_{\ell,h}^{\pi^*}(b_{\ell,h})$ (Low oracle importance, wrongly retained)
\end{itemize}

% Consequently, kept cache set $\mathcal{M}_{\ell,h}^{\pi}(b_{\ell,h})$ determined by $\pi$ can also be expressed as:
Consequently, \camblue{the retained cache set} $\mathcal{M}_{\ell,h}^{\pi}(b_{\ell,h})$ determined by $\pi$ can also be expressed as:
\begin{equation}
\label{eq:set_compose}
\begin{aligned}
\mathcal{M}_{\ell,h}^\pi(b_{\ell,h}) = (\mathcal{M}_{\ell,h}^{\pi^*}(b_{\ell,h})\setminus \mathcal{M}_{\ell,h,\mathrm{miss}}) \cup \mathcal{M}_{\ell,h,\mathrm{fp}}.
\end{aligned}
\end{equation}

% Let \textbf{eviction loss} $\mathcal{L}_{\ell,h}(\cdot )$ denote the Oracle Importance mass lost by head $(\ell,h)$ due to the removal of kv cache. 
Let eviction loss $\mathcal{L}_{\ell,h}(\cdot )$ denote the Oracle Importance mass lost by head $(\ell,h)$ due to \camblue{removed KV-cache positions:}
% \camblue{Since compression is applied to the prefill cache, $j$ denotes a cached token position in $\{1,\dots,T\}$, and all set complements in this subsection are taken with respect to this prefill-position set.}
\begin{equation}
\label{eq:head_loss_def}
% \mathcal{L}_{\ell,h}(\mathcal{M}_{\ell,h})
% \triangleq
% \sum_{j\notin \mathcal{M}_{\ell,h}}I_{\ell,h,j}.
\mathcal{L}_{\ell,h}(\mathcal{M}_{\ell,h})
\triangleq
\sum_{\camblue{j\in \{1,\dots,T\}\setminus \mathcal{M}_{\ell,h}}}I_{\ell,h,j}.
\end{equation}

% \camblue{Here the argument $\mathcal{M}_{\ell,h}$ denotes the retained set of cached positions for head $(\ell,h)$.}

% Then the eviction loss of metric $\pi$ can be rigorously decomposed as follows:
Substituting the set formulation from Eq.~\eqref{eq:set_compose} into the loss definition in Eq.~\eqref{eq:head_loss_def}, we can rigorously decompose the eviction loss of policy $\pi$ into two distinct components:
\begin{equation}
\label{eq:loss_decomp}
\resizebox{0.49\textwidth}{!}{ % 0.9表示缩放到页面宽度的90%，!表示高度按比例缩放
$\begin{aligned}
\mathcal{L}_{\ell,h}(\mathcal{M}_{\ell,h}^{\pi}(b_{\ell,h})) 
&= \mathcal{L}_{\ell,h}(\mathcal{M}_{\ell,h}^{\pi^*}(b_{\ell,h})) + \!\!\!\!\!\sum_{j \in \mathcal{M}_{\ell,h,\mathrm{miss}}} \!\!\!\!\!\!I_{\ell,h,j} - \!\!\!\!\sum_{j \in \mathcal{M}_{\ell,h,\mathrm{fp}}} \!\!\!\!I_{\ell,h,j}  \\[-2pt]
&= \underbrace{\mathcal{L}_{\ell,h}(\mathcal{M}_{\ell,h}^{\pi^*}(b_{\ell,h}))}_{\text{Oracle Metric Loss}}
+
\underbrace{\Delta_{\ell,h}(\pi, \pi^*,\camblue{b_{\ell,h}})}_{\text{Optimality Gap Loss}}.
\end{aligned}$
}
\end{equation}
where $\mathcal{L}_{\ell,h}(\mathcal{M}_{\ell,h}^*)$ is the Oracle Metric Loss, which is fixed due to compression ratio; $\Delta_{\ell,h}(\pi,I)$ is defined as the \textbf{Optimality Gap} between the oracle metric and the used metric $\pi$ in long-horizon view, which relevant to $\pi$.

Figure~\ref{fig:misalignment_gap} validates the decomposition in Eq.~\eqref{eq:loss_decomp} by visualizing the total loss as the vertical distance to $Recall=1.0$. Specifically, the loss of a heuristic metric $\mathcal{L}_{\ell,h}(\mathcal{M}_{\ell,h}^{\pi}(b_{\ell,h}))$ is the sum of the inherent oracle metric loss(red curve to 1.0) and the optimality gap loss(vertical gap between heuristic metric curve and oracle curve). 
% This observation suggests two optimization paths: (1) refining budget allocation cross head to lower the total Oracle loss, and (2) improving the selection metric to bridge the optimality gap.

\subsection{Global Optimization of Head-Level KV Cache Budget Allocation}
\label{subsec:allocation}

The formulation above characterizes the loss within a single attention head; however, modern LLMs operate through a complex multi-head, multi-layer architecture. Existing head-wise allocation approaches, e.g, AdaKV~\cite{feng_adakv_2024}, attempt to address this by employing a global greedy strategy that pools candidate tokens from all heads and retains the top-$K$ elements based on surrogate scores. Nevertheless, this strategy remains suboptimal due to the existence of the optimality gap \camblue{$\Delta_{\ell,h}(\pi, \pi^*, b_{\ell,h})$} defined in Eq.~\ref{eq:loss_decomp}.

\paragraph{Global Optimization Objective.}
We now consider the problem of allocating a global cache budget $B_{\text{total}}$ across all attention heads to minimize the aggregate eviction loss across the entire model. 

Let $b_{\ell,h}$ denote the cache budget allocated to head $(\ell,h)$, subject to the global constraint $\sum_{\ell,h} b_{\ell,h} = B_{\text{total}}$. For a given metric $\pi$, we define $\mathcal{M}_{\ell,h}^{\pi}(b_{\ell,h})$ as the set of top-$b_{\ell,h}$ token positions selected by $\pi$ within that head.

The global optimization objective aims to minimize the aggregate eviction loss across all layers and heads by optimizing the budget distribution $\{b_{\ell,h}\}$:
\begin{equation}
\label{eq:combinatorial_optimization}
\begin{aligned}
\min_{\{b_{\ell,h}\}} \quad & \sum_{\ell=1}^{L}\sum_{h=1}^{H} \mathcal{L}_{\ell,h}\!\left(\mathcal{M}_{\ell,h}^{\pi}(b_{\ell,h})\right), \\
\text{s.t.} \quad & \sum_{\ell=1}^{L}\sum_{h=1}^{H} b_{\ell,h} = B_{\text{total}}.
\end{aligned}
\end{equation}

% Since $\mathcal{M}_{\ell,h}^{\pi}(b_{\ell,h})$ lacks strict monotonicity with respect to the oracle importance, and given that the parameter space for $\{b_{\ell,h}\}$ constitutes a high-dimensional combinatorial domain, rendering the global optimization problem NP-hard. The proof of the non-convex of $\mathcal{L}_{\ell,h}\!\left(\mathcal{M}_{\ell,h}^{\pi}(b_{\ell,h})\right)$ is provided in Appendix~\ref{app:prove_non_convex_comb_opt}. 

Since the ordering induced by $\pi$ can make the discrete loss sequence
$\{\mathcal{L}_{\ell,h}(\mathcal{M}_{\ell,h}^{\pi}(i))\}_{i=0}^{T}$
non-convex with respect to the integer budget, Eq.~\ref{eq:combinatorial_optimization}
forms a high-dimensional discrete combinatorial allocation problem. Although exact dynamic programming is possible when all per-head loss tables are available, it is costly at profiling scale. The proof of the non-convexity of $\mathcal{L}_{\ell,h}\!\left(\mathcal{M}_{\ell,h}^{\pi}(b_{\ell,h})\right)$ is provided in Appendix~\ref{app:prove_non_convex_comb_opt}.

\begin{figure}[t]
    \centering
    % --- 第一个子图 ---
    \begin{subfigure}[b]{0.4\textwidth} % 宽度调大一点更美观
        \centering
        \includegraphics[width=\linewidth]{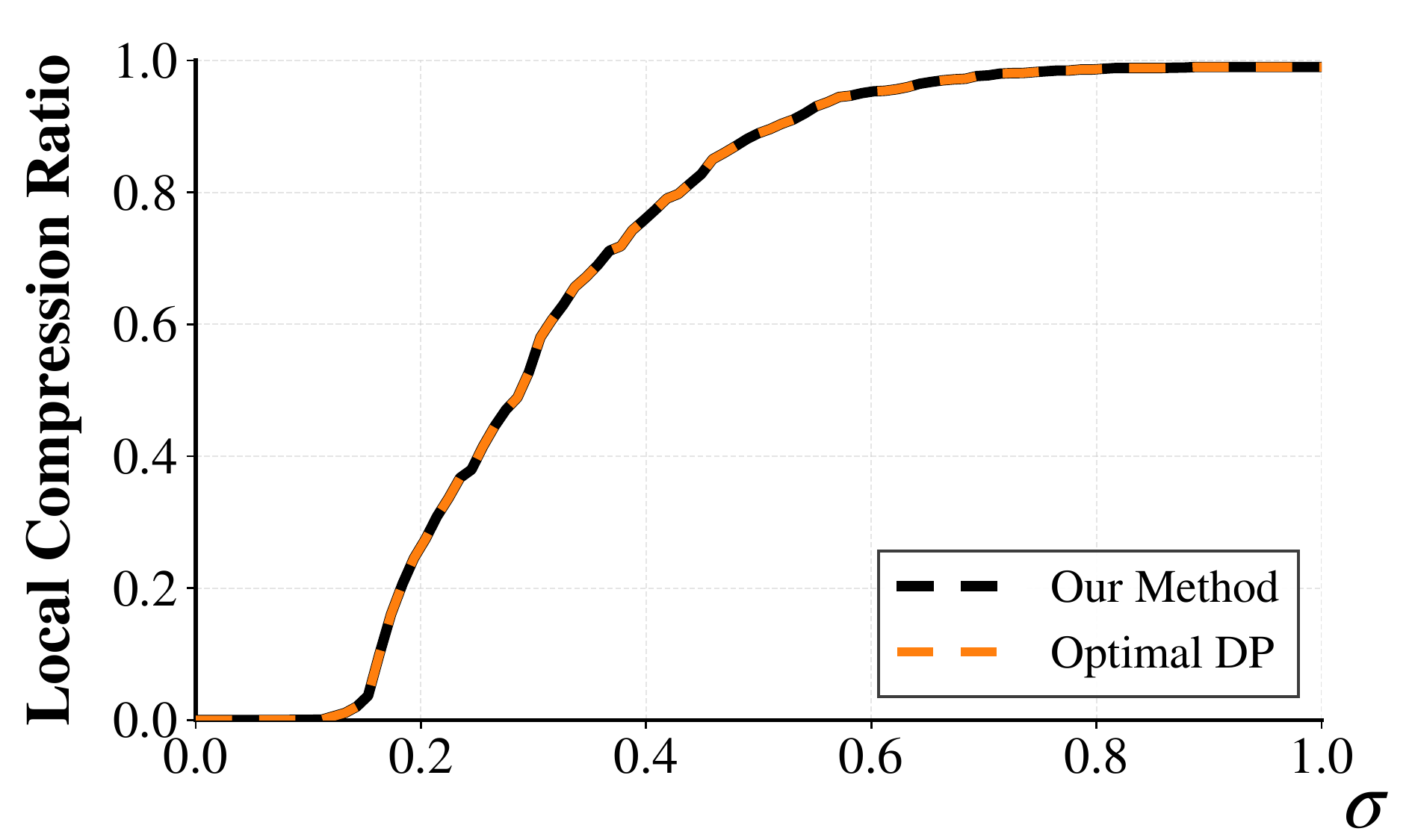}
        \vspace{-0.3in}
        \caption{} %{Solver Comparison}
        \label{fig:compare_dp}
    \end{subfigure}
    \hfill % 使用 \hfill 撑开中间间距，确保左右对齐
    % --- 第二个子图（注意：上一行 subfigure 结束后立即接 \hfill，不要留空行） ---
    \begin{subfigure}[b]{0.4\textwidth}
        \centering
        \includegraphics[width=\linewidth]{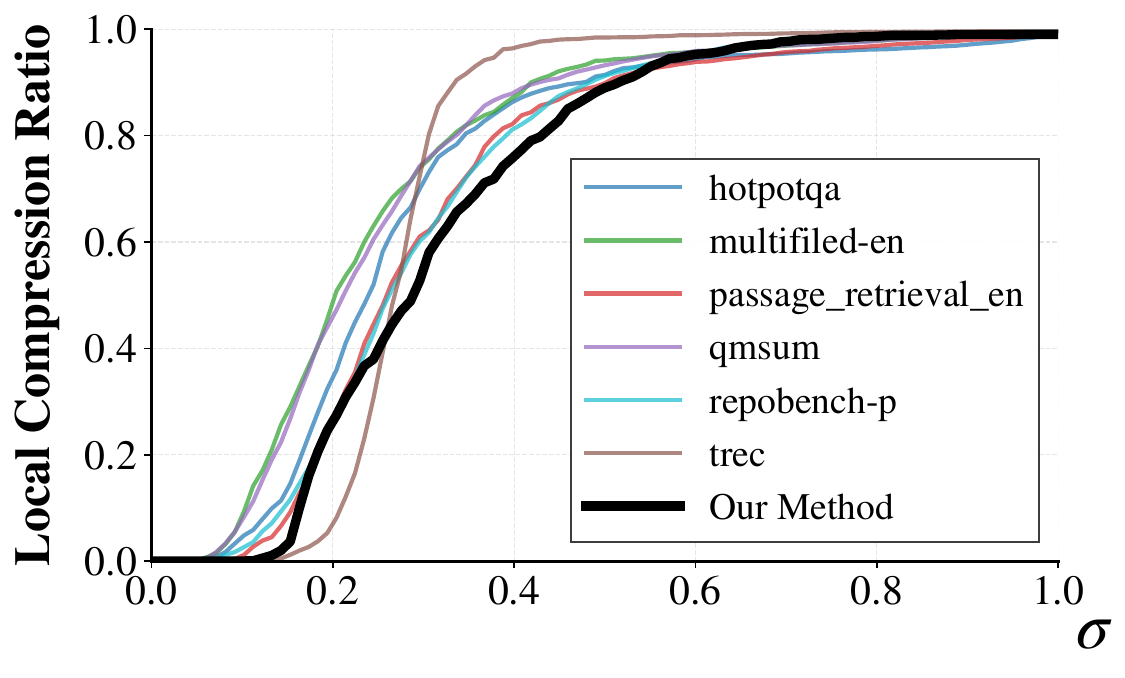}
        \vspace{-0.3in}
        \caption{} %{Consistent Trend of Optimal Cache Allocation}
        \label{fig:compare_longbench_tasks}
    \end{subfigure}
    
    \caption{
        (a) Comparison between our greedy solver based on convex-hull relaxation (solving Eq.~\ref{eq:combinatorial_optimization_relaxd}) and DP solution (solving Eq.~\ref{eq:combinatorial_optimization}). (b) Shows the consistent trend of optimal local compression ratio \camblue{(the head-level ratio $r_{\ell,h}$)}  across different downstream tasks under the same global compression ratio $\sigma$. Evaluated on Mistral-7B-v0.3.
    }
    \label{fig:allocation_analysis}
\end{figure}

\paragraph{Efficient Optimization via Convex Hull Relaxation.}
To facilitate an efficient solution to the objective in Equation~\ref{eq:combinatorial_optimization}, we propose a convex relaxation approach that transforms the discrete loss $\mathcal{L}_{\ell,h}$ into a tractable surrogate.
% By applying Isotonic Regression via the Pool Adjacent Violators Algorithm (PAVA) to the raw  loss sequence $ \left \{\mathcal{L}_{\ell,h}\!\left( \mathcal{M}_{\ell,h}^{\pi}(i) \right) | 1\le i\le T \right  \} $, we derive a convex, non-increasing surrogate sequence, denoted as
% $ \left \{\breve{\mathcal{L}}_{\ell,h}\!\left( \mathcal{M}_{\ell,h}^{\pi}(i) \right) | 1\le i\le T \right  \} $.

We first compute the raw marginal decreases of the loss sequence
$ \left \{\mathcal{L}_{\ell,h}\!\left( \mathcal{M}_{\ell,h}^{\pi}(i) \right) | 0\le i\le T \right  \} $
and apply Isotonic Regression via the Pool Adjacent Violators Algorithm (PAVA) to project them onto a non-negative, non-increasing sequence. Reconstructing the loss from these projected marginal decreases yields a convex, non-increasing surrogate sequence, denoted as
$ \left \{\breve{\mathcal{L}}_{\ell,h}\!\left( \mathcal{M}_{\ell,h}^{\pi}(i) \right) | 0\le i\le T \right  \}$.
We define the effective marginal gain of allocating $i$-th token in $\mathcal{M}_{\ell,h}^{\pi}$:
\begin{equation}
\label{eq:marginal_gain}
g_{\ell,h}^{\pi}(i) = \breve{\mathcal{L}}_{\ell,h}\!\left(\mathcal{M}_{\ell,h}^{\pi}(i-1)\right) - \breve{\mathcal{L}}_{\ell,h}\!\left(\mathcal{M}_{\ell,h}^{\pi}(i)\right) \ge 0.
\end{equation}
The marginal gain $g_{\ell,h}^{\pi}(i)$ is monotonically non-increasing while $i$ increases. This property allows a global greedy strategy to achieve the global optimum for the \textit{relaxed} objective:
\begin{equation}
\label{eq:combinatorial_optimization_relaxd}
\begin{aligned}
\min_{\{b_{\ell,h}\}} \quad &
\sum_{\ell=1}^{L}\sum_{h=1}^{H}
\breve{\mathcal{L}}_{\ell,h}\!\left(\mathcal{M}_{\ell,h}^{\pi}(b_{\ell,h})\right), \\
\text{s.t.} \quad &
\sum_{\ell=1}^{L}\sum_{h=1}^{H} b_{\ell,h} = B_{\text{total}}.
\end{aligned}
\end{equation}
Specifically, we iteratively allocate the $i$-th token from the attention head $(\ell, h)$ that yields the maximum effective marginal gain $g_{\ell,h}^{\pi}(i)$, continuing until the global budget $B_{\text{total}}$ is exhausted. As illustrated in Figure~\ref{fig:compare_dp}, our approach achieves an exact match with the results of the optimal Dynamic Programming (DP) solver.
Details of the convex relaxation and allocation process are in Appendix~\ref{app:convex_relaxation}.

\subsection{Practical Implementation: Offline Profiling}
\label{subsec:calibration}

The optimization problem formulated in Eq.~\ref{eq:combinatorial_optimization_relaxd} requires future decoding results to compute the oracle importance $I$, which is inherently inaccessible during real-time inference. 

However, we identify a key structural property of LLMs: individual attention heads exhibit a \textbf{consistent trend} in their optimal local-to-global compression ratios across diverse tasks. As illustrated in Figure~\ref{fig:compare_longbench_tasks}, these compression profiles remain remarkably stable across various scenarios, e.g, question answering and long-context retrieval. This empirical consistency allows us to characterize the optimal global-local budget allocation in an offline manner.

\paragraph{Offline Optimal Budget Estimation.}
To construct this offline allocation, we employ a data-driven probing protocol consisting of three phases:

\begin{itemize}[leftmargin=*, nosep]
    \item \textbf{Context Generation:} We construct a long-context input $C_{\text{syn}}$ ($\approx$ 4,000 tokens) with a coherent narrative structure to simulate realistic KV cache states. \camblue{In our default profile, $C_{\text{syn}}$ is an AI-generated Chinese novel excerpt and is disjoint from all evaluation benchmarks. We provide a transferability analysis of the offline profiling data in Appendix~\ref{append:Trans_ana_of_offline_profile_data}. }
    \item \textbf{Oracle Computation:} We generate a diverse set of queries $\mathcal{Q} = \{q_1, \dots, q_M\}$ targeting different information segments. \camblue{We use $M=30$ generated questions in the default profile.} For each $q_i$, the ground-truth oracle importance is computed via full-attention decoding.
    \item \textbf{Profile Aggregation:} We solve Eq.~\ref{eq:combinatorial_optimization_relaxd} for each query across a dense grid of global compression ratios $\rho \in [0, 1]$ to obtain the query-specific optimal local ratios $r^{*(\pi)}_{\ell,h}(q_i; \rho)$.
\end{itemize}

We aggregate these solutions into a final static profile $\Phi^{(\pi)}$ by averaging the optimal local ratios across the calibration set:
\begin{equation}
\label{eq:calibration_avg}
\Phi^{(\pi)}(\rho)_{\ell,h} \triangleq \frac{1}{M}\sum_{i=1}^{M} r^{*(\pi)}_{\ell,h}(q_i;\rho).
\end{equation}
The resulting $\Phi^{(\pi)}$ serves as a lookup table mapping any target global sparsity $\rho$ to a precise configuration for every head, effectively capturing the \textbf{expected utility} of each head across the general data distribution.

\paragraph{Online Execution.}
During inference, our method introduces negligible computational overhead through three steps:
\begin{enumerate}[leftmargin=*, nosep]
    \item \textbf{Lookup:} Given a target global compression ratio $\sigma_{\text{target}}$, the system retrieves the pre-computed local ratios $\{r_{\ell,h}\} \leftarrow \Phi^{(\pi)}(\sigma_{\text{target}})$.
    \item \textbf{Budgeting:} These ratios are translated into integer budgets: $b_{\ell,h} = \lfloor (1 - r_{\ell,h}) \cdot T \rfloor$.
    \item \textbf{Eviction:} Each head independently applies the heuristic metric $\pi$ to retain the top-$b_{\ell,h}$ tokens.
\end{enumerate}
This strategy successfully bridges the gap between theoretical oracle performance and practical runtime constraints without requiring online optimization.

% ==========================================
% TABLE 1: LongBench Results
% ==========================================
\begin{table*}[t!]
\centering
% 调整列间距
\footnotesize
% \small
\setlength{\tabcolsep}{5pt} 
\caption{
\textbf{Main Results on LongBench.} Comparison of KV cache eviction strategies using the SnapKV metric ($\pi_1$) and the KeyDiff metric ($\pi_2$) at an 80\% compression ratio.}

\label{tab:longbench_0.8}

% 使用 resizebox 强制表格填满整个文本宽度
\resizebox{\textwidth}{!}{%
\begin{tabularx}{\linewidth}{ll YYYY YYY}   
\toprule
\multirow{2}{*}{\textbf{Model}} & \multirow{2}{*}{\textbf{Method}} & \multicolumn{2}{c}{\textbf{QA}} & \multirow{2}{*}{\textbf{Summ.}} & \multirow{2}{*}{\textbf{Few-Shot}} & \multirow{2}{*}{\textbf{Synth.}} & \multirow{2}{*}{\textbf{Code}} & \multirow{2}{*}{\textbf{Avg.}} \\
\cmidrule(lr){3-4} 
 & & \textbf{Single} & \textbf{Multi} & & & & & \\
\midrule

% =============================================
% Mistral-7B-Instruct-v0.3
% =============================================
\multirow{10}{*}{\rotatebox[origin=c]{90}{\textbf{Mistral-7B-v0.3}}} 
 & Full-KV  & 38.36 & 37.83 & 28.86 & 70.86 & 51.25 & 63.26 & 47.30 \\
 \cmidrule(lr){2-9}
 % --- SnapKV Variants ---
 % & \textit{Metric SnapKV ($\pi_1$)} & & & & & & & \\
 & \hspace{1em} Uniform-$\pi_1$ & 24.70 & 31.78 & 24.15 & 63.95 & 47.50 & 60.98 & 40.67 \\
 & \hspace{1em} Pyramid-$\pi_1$ & 25.36 & 31.95 & 23.98 & 63.35 & 48.75 & 61.80 & 40.94 \\
 & \hspace{1em} Ada-$\pi_1$ & 26.21 & 31.88 & 24.42 & 66.52 & 49.50 & \textbf{62.05} & 41.89 \\
 \rowcolor{color_ours} \cellcolor{white} & \hspace{1em} \textbf{LU-KV-$\pi_1$ (Ours)} & \textbf{37.16} & \textbf{36.59} & \textbf{27.97} & \textbf{69.81} & \textbf{51.35} & 57.69 & \textbf{45.79} \\
 % \addlinespace[4pt]
 % --- KeyDiff Variants ---
 % & \textit{Metric KeyDiff ($\pi_2$)} & & & & & & & \\
  \cmidrule(lr){2-9}
 & \hspace{1em} Uniform-$\pi_2$ & 28.53 & 30.86 & 24.71 & 59.00 & 33.08 & 46.89 & 36.83 \\
 & \hspace{1em} Pyramid-$\pi_2$ & 29.44 & 30.53 & 24.50 & 59.33 & 34.06 & 42.95 & 36.59 \\
 & \hspace{1em} Ada-$\pi_2$ & 31.94 & 33.78 & 25.23 & 59.60 & 40.79 & \textbf{57.71} & 40.54 \\
 \rowcolor{color_ours} \cellcolor{white} & \hspace{1em} \textbf{LU-KV-$\pi_2$ (Ours)} & \textbf{39.80} & \textbf{38.09} & \textbf{28.22} & \textbf{68.32} & \textbf{52.02} & 56.04 & \textbf{46.21} \\
\midrule

% =============================================
% Qwen2.5-32B-Instruct
% =============================================
\multirow{10}{*}{\rotatebox[origin=c]{90}{\textbf{Qwen2.5-32B}}} 
 & Full-KV   & 42.91 & 54.15 & 27.33 & 68.91 & 55.75 & 42.35 & 48.51 \\
 \cmidrule(lr){2-9}
 % --- SnapKV Variants ---
 % & \textit{Metric SnapKV ($\pi_1$)} & & & & & & & \\
 & \hspace{1em} Uniform-$\pi_1$ & 25.02 & 44.50 & 23.33 & 64.45 & 48.75 & 44.99 & 41.21 \\
 & \hspace{1em} Pyramid-$\pi_1$ & 19.44 & 40.23 & 21.84 & 60.24 & 50.46 & 46.34 & 38.68 \\
 & \hspace{1em} Ada-$\pi_1$ & 25.75 & 43.60 & 23.34 & 65.70 & 50.13 & 44.97 & 41.58 \\
 \rowcolor{color_ours} \cellcolor{white} & \hspace{1em} \textbf{LU-KV-$\pi_1$ (Ours)} & \textbf{39.84} & \textbf{53.55} & \textbf{26.19} & \textbf{67.32} & \textbf{54.75} & \textbf{48.46} & \textbf{47.95} \\
 % \addlinespace[4pt]
 % --- KeyDiff Variants ---
 % & \textit{Metric KeyDiff ($\pi_2$)} & & & & & & & \\
  \cmidrule(lr){2-9}
 & \hspace{1em} Uniform-$\pi_2$ & 26.85 & 44.30 & 22.32 & 65.82 & 38.36 & 29.32 & 38.33 \\
 & \hspace{1em} Pyramid-$\pi_2$ & 22.00 & 35.87 & 20.64 & 60.05 & 24.71 & 26.85 & 32.42 \\
 & \hspace{1em} Ada-$\pi_2$ & 32.16 & 46.44 & 23.68 & 66.33 & 48.92 & 36.71 & 42.32 \\
 \rowcolor{color_ours} \cellcolor{white} & \hspace{1em} \textbf{LU-KV-$\pi_2$ (Ours)} & \textbf{41.58} & \textbf{54.23} & \textbf{26.61} & \textbf{67.86} & \textbf{53.75} & \textbf{46.91} & \textbf{48.26} \\
\bottomrule
\end{tabularx}
}
\end{table*}

% ==========================================
% TABLE 2: RULER Results
% ==========================================
\begin{table*}[t!]
\centering
\small 
\setlength{\tabcolsep}{2.5pt} 
\renewcommand{\arraystretch}{1.15}
% \caption{\textbf{Results on RULER-16K.} Comparison of different KV cache eviction strategies under \textbf{80\% compression ratio}. We evaluate standard baselines and their \textit{KeyDiff} variants. ``\textbf{Ours}'' denotes our proposed method. Results are reported as accuracy (\%).}
% \label{tab:ruler_16k_0.8}
\caption{
\textbf{Main Results on RULER-16K.} Comparison of KV cache eviction strategies using the SnapKV metric ($\pi_1$) and the KeyDiff metric ($\pi_2$) at an 80\% compression ratio.}
\label{tab:ruler_16k_0.8}

\resizebox{\textwidth}{!}{%
\begin{tabular}{llcccccccccccccc}
\toprule
% ================= HEADER =================
\multirow{2}{*}{\textbf{Model}} & \multirow{2}{*}{\textbf{Method}} & 
\multicolumn{14}{c}{\textbf{RULER Tasks (16K)}} \\
\cmidrule(lr){3-16}
& & 
\rotatebox{30}{single1} & \rotatebox{30}{single2} & \rotatebox{30}{single3} & 
\rotatebox{30}{multikey1} & \rotatebox{30}{multikey2} & \rotatebox{30}{multikey3} & 
\rotatebox{30}{multivalue} & \rotatebox{30}{multiquery} & 
\rotatebox{30}{vt} & \rotatebox{30}{cwe} & \rotatebox{30}{fwe} & 
\rotatebox{30}{qa-1} & \rotatebox{30}{qa-2} & \rotatebox{30}{\textbf{Avg}} \\

\midrule

% =============================================
% Mistral-7B-Instruct-v0.3
% =============================================
\multirow{10}{*}{\rotatebox[origin=c]{90}{\textbf{Mistral-7B-v0.3}}} 
 & Full-KV  & 94.20 & 96.40 & 99.60 & 97.40 & 95.60 & 76.80 & 89.50 & 88.65 & 96.28 & 82.22 & 87.93 & 71.60 & 50.00 & 86.63 \\
 \cmidrule(lr){2-16}
 
 % --- SnapKV Variants ---
 % & \textit{Metric SnapKV ($\pi_1$)} & & & & & & & & & & & & & & \\
 & \hspace{1em} Uniform-$\pi_1$ & 40.40 & 16.20 & 2.40 & 14.20 & 6.20 & 1.00 & 9.65 & 11.00 & 66.92 & 66.96 & 85.53 & 29.80 & 33.60 & 29.53 \\
 & \hspace{1em} Pyramid-$\pi_1$ & 50.00 & 57.00 & 2.40 & 28.00 & 4.80 & 0.20 & 16.15 & 21.55 & 62.32 & 31.94 & 82.20 & 32.00 & 33.00 & 32.43 \\
 & \hspace{1em} Ada-$\pi_1$ & 58.00 & 38.80 & 2.40 & 20.20 & 12.40 & 5.60 & 12.85 & 16.80 & 92.08 & 71.36 & \textbf{86.13} & 33.60 & 37.00 & 37.48 \\
 \rowcolor{color_ours} \cellcolor{white} & \hspace{1em} \textbf{LU-KV-$\pi_1$ (Ours)} & \textbf{70.80} & \textbf{78.80} & \textbf{18.20} & \textbf{83.60} & \textbf{79.20} & \textbf{67.40} & \textbf{67.80} & \textbf{76.25} & \textbf{95.88} & \textbf{78.32} & 84.47 & \textbf{62.00} & \textbf{47.00} & \textbf{69.98} \\
 
 % \addlinespace[4pt] 
 
 % --- KeyDiff Variants ---
 % & \textit{Metric KeyDiff ($\pi_2$)} & & & & & & & & & & & & & & \\
 \cmidrule(lr){2-16}
 & \hspace{1em} Uniform-$\pi_2$ & \textbf{94.60} & 72.80 & 100.00 & 78.80 & 7.40 & 0.80 & 94.80 & 86.10 & 94.16 & 65.56 & \textbf{90.87} & 32.40 & 35.80 & 65.70 \\
 & \hspace{1em} Pyramid-$\pi_2$ & 93.20 & \textbf{96.20} & 99.60 & \textbf{88.20} & 6.60 & 0.60 & 92.00 & 89.75 & \textbf{94.36} & 36.92 & 88.73 & 31.40 & 34.80 & 65.57 \\
 & \hspace{1em} Ada-$\pi_2$ & 92.60 & 91.20 & 97.40 & 87.80 & 6.80 & 1.20 & 88.00 & 86.45 & 91.28 & 75.44 & 86.47 & 36.40 & 36.60 & 67.51 \\
 \rowcolor{color_ours} \cellcolor{white} & \hspace{1em} \textbf{LU-KV-$\pi_2$ (Ours)} & 85.60 & 76.60 & 100.00 & 87.00 & \textbf{90.80} & \textbf{35.20} & \textbf{96.45} & \textbf{92.85} & 92.16 & \textbf{80.78} & 86.80 & \textbf{64.60} & \textbf{46.80} & \textbf{79.66} \\
\midrule

% =============================================
% Qwen2.5-32B-Instruct
% =============================================
\multirow{10}{*}{\rotatebox[origin=c]{90}{\textbf{Qwen2.5-32B}}} 
 & Full-KV  & 100.00 & 100.00 & 100.00 & 100.00 & 99.80 & 100.00 & 99.85 & 99.95 & 100.00 & 97.70 & 96.20 & 79.40 & 62.40 & 95.02 \\
 \cmidrule(lr){2-16}
 
 % --- SnapKV Variants ---
 % & \textit{Metric SnapKV ($\pi_1$)} & & & & & & & & & & & & & & \\
 & \hspace{1em} Uniform-$\pi_1$ & 97.40 & 55.60 & 3.80 & 25.80 & 4.80 & 2.00 & 14.40 & 19.60 & 99.28 & 87.14 & 94.00 & 28.00 & 39.00 & 43.91 \\
 & \hspace{1em} Pyramid-$\pi_1$ & 83.80 & 36.00 & 2.40 & 19.20 & 2.00 & 0.00 & 13.15 & 14.95 & 93.68 & 56.84 & \textbf{95.73} & 26.40 & 34.60 & 36.83 \\
 & \hspace{1em} Ada-$\pi_1$ & 98.80 & 52.60 & 4.40 & 21.80 & 7.00 & 4.20 & 14.75 & 18.25 & 99.32 & 88.48 & 94.53 & 29.40 & 39.00 & 44.04 \\
 \rowcolor{color_ours} \cellcolor{white} & \hspace{1em} \textbf{LU-KV-$\pi_1$ (Ours)} & \textbf{99.80} & \textbf{99.20} & \textbf{32.00} & \textbf{84.20} & \textbf{71.80} & \textbf{78.40} & \textbf{84.60} & \textbf{85.80} & \textbf{99.72} & \textbf{95.66} & 93.13 & \textbf{65.00} & \textbf{56.80} & \textbf{80.47} \\
 
 % \addlinespace[4pt]
 
 % --- KeyDiff Variants ---
 % & \textit{Metric KeyDiff ($\pi_2$)} & & & & & & & & & & & & & & \\
 \cmidrule(lr){2-16}
 & \hspace{1em} Uniform-$\pi_2$ & 100.00 & 100.00 & 100.00 & 100.00 & 8.00 & 1.00 & 99.40 & 99.95 & 98.92 & 90.36 & \textbf{99.33} & 36.40 & 41.40 & 74.98 \\
 & \hspace{1em} Pyramid-$\pi_2$ & 100.00 & 100.00 & 99.80 & 99.60 & 1.00 & 0.20 & \textbf{99.55} & 99.95 & 84.52 & 69.26 & 98.93 & 30.20 & 35.80 & 70.68 \\
 & \hspace{1em} Ada-$\pi_2$ & 100.00 & 100.00 & 100.00 & 99.80 & 44.20 & 23.00 & 99.00 & 99.90 & 99.88 & 95.34 & 99.07 & 44.00 & 46.40 & 80.81 \\
 \rowcolor{color_ours} \cellcolor{white} & \hspace{1em} \textbf{LU-KV-$\pi_2$ (Ours)} & 100.00 & 100.00 & 100.00 & 100.00 & \textbf{86.60} & \textbf{46.80} & 99.25 & 99.95 & \textbf{100.00} & \textbf{96.02} & 96.20 & \textbf{71.60} & \textbf{59.00} & \textbf{88.88} \\
\bottomrule
\end{tabular}%
}
\end{table*}

%%%%%%%%%%%%%%%%%%%%%%%%%%%%%%%%%%%%%%
% Loss
%%%%%%%%%%%%%%%%%%%%%%%%%%%%%%%
\begin{figure*}[t]
    \centering
    % 第一张图
    \begin{minipage}{0.48\textwidth}
        \centering
        \includegraphics[width=1\linewidth]{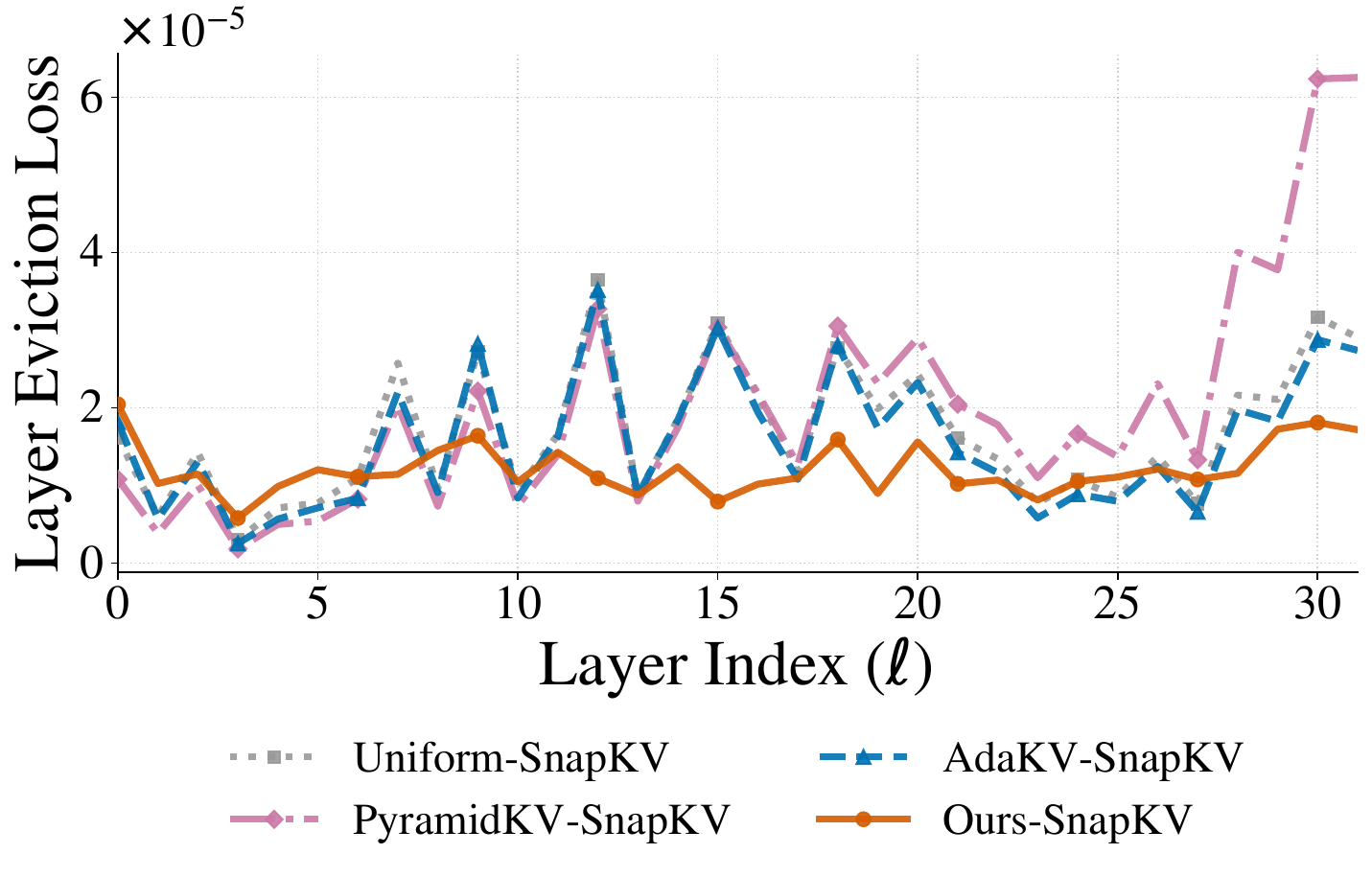}
        \caption{Comparison of aggregated layer-wise eviction loss. \textbf{Ours} consistently achieves the lowest and most stable loss across all layers, whereas baselines like AdaKV and PyramidKV exhibit severe loss spikes.}
        \label{fig:alloc_trend}
    \end{minipage}
    \hfill 
    % 第二张图
    \begin{minipage}{0.48\textwidth}
        \centering
        % \vspace{-0.0in}
        \includegraphics[width=0.98\linewidth]{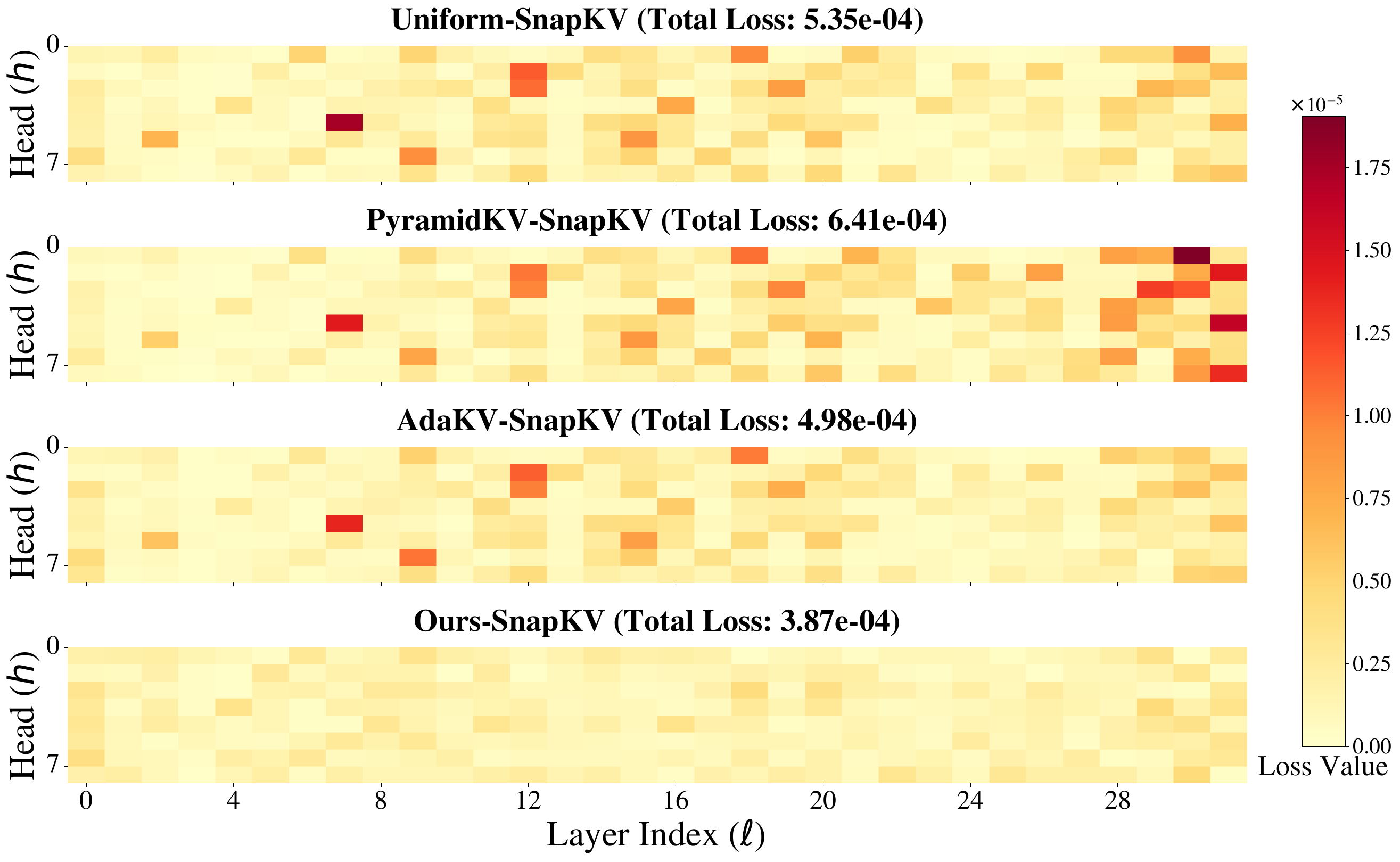}
        \caption{Heatmap visualization of per-head loss distribution $\mathcal{L}_{\ell,h}$. Baselines suffer from intense "loss bursts" (dark red blocks) in specific heads due to optimality gap, while our method effectively suppresses these spikes across the entire model.}
        \label{fig:alloc_heatmap}
    \end{minipage}
\end{figure*}

\begin{figure*}[t]
    \centering
    % 左图：显存
    \begin{subfigure}{0.46\textwidth}
        \includegraphics[width=\linewidth]{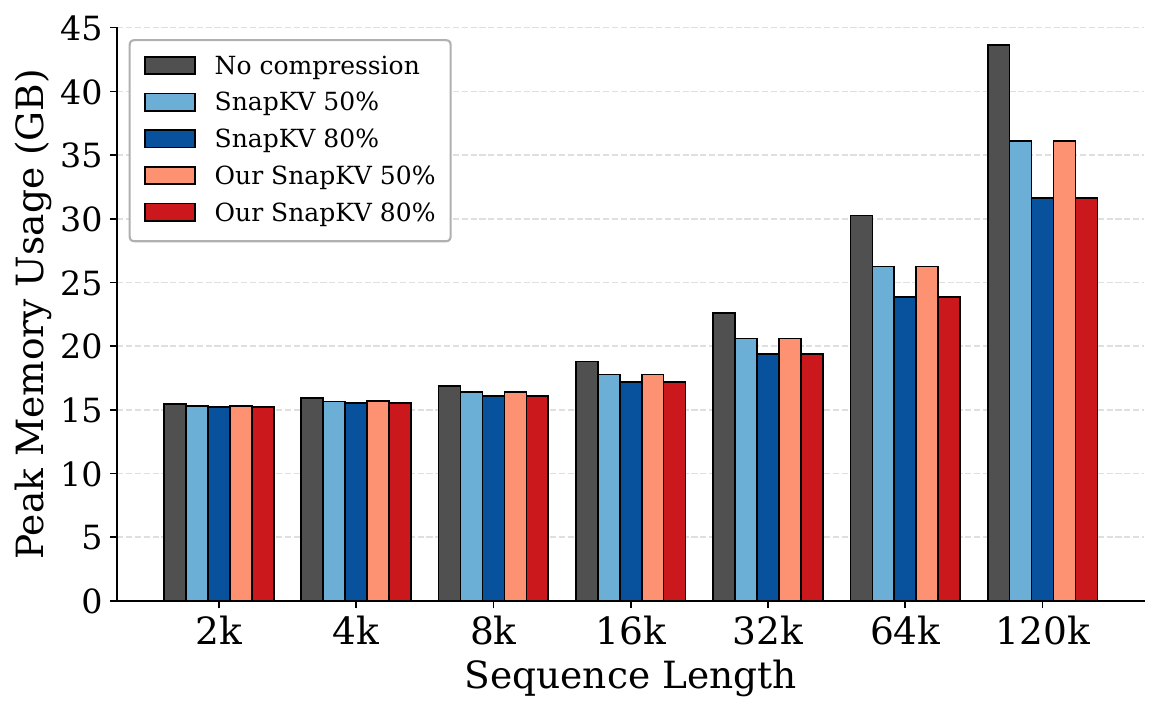}
        \caption{Peak memory usage}
        \label{fig:memory}
    \end{subfigure}
    \;\;\;\;\;\;\;\;\;\;\;\;
    % 右图：时延
    \begin{subfigure}{0.46\textwidth}
        \includegraphics[width=\linewidth]{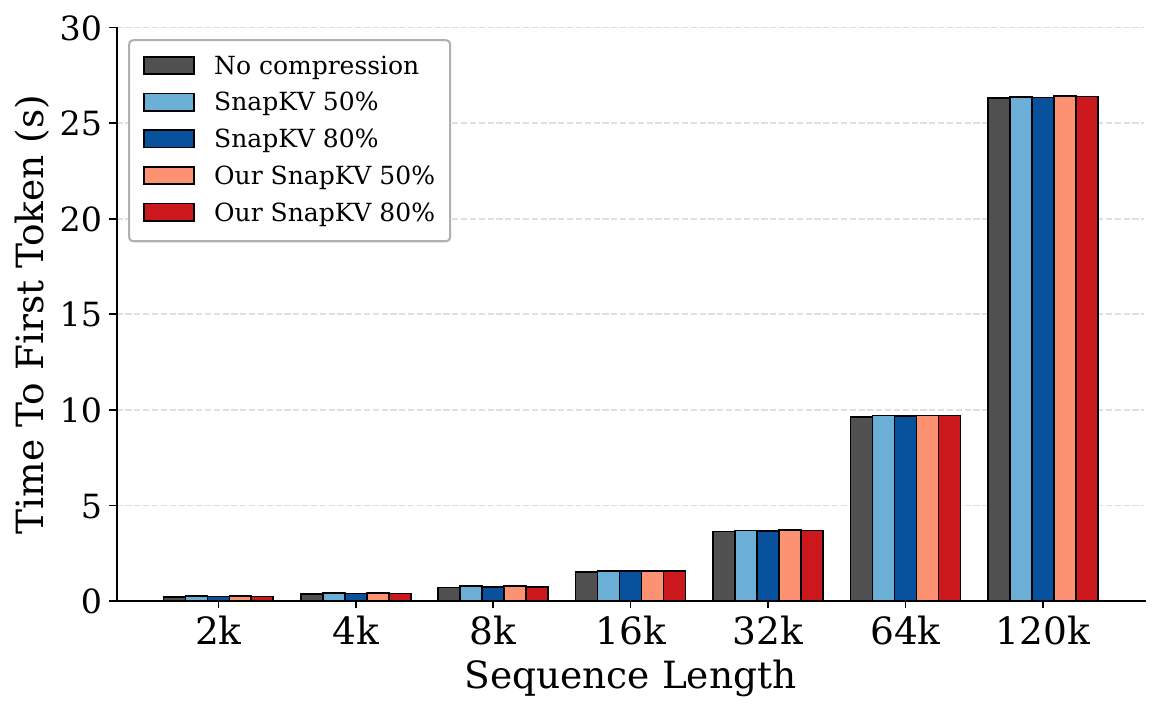}
        \caption{Time To First Token (TTFT) latency}
        \label{fig:ttft}
    \end{subfigure}
    
    \caption{Efficiency comparison on Llama-3.1-8b. Our method maintains comparable latency to baselines while significantly reducing memory usage in long-context scenarios.}
    \label{fig:efficiency}
\end{figure*}

\section{Experiments}
\label{sec:experiments}

\subsection{Experimental Setup}
\label{subsec:setup}

\noindent \textbf{Benchmarks.}
We assess general long-context generation capabilities using \textbf{LongBench}~\citep{bai_longbench_2024}, which consists of 16 diverse datasets covering various long-form tasks. Additionally, we utilize \textbf{RULER}~\citep{hsieh_ruler_2024} to evaluate retrieval robustness across expanding context windows, ranging from 4k to 128k tokens. Further details regarding the benchmarks are provided in Appendix~\ref{app:benchmark}.

\noindent \textbf{Base Models and Baseline Methods.}
We evaluate our method using LLMs of varying scales and context window capacities: Llama-3.1-8B-Instruct~\citep{dubey_llama3_2024}, Mistral-7B-Instruct-v0.3~\citep{jiang_mistral_2023}, and Qwen2.5-32B-Instruct~\citep{qwen2.5}.
We consider two KV cache importance metrics: the \textbf{ SnapKV}~\citep{yuhong_SnapKV_2024} metric, denoted as $\pi_1$, which relies on accumulated attention scores; and the \textbf{KeyDiff}~\citep{jun_keydiff_2025} metric, denoted as $\pi_2$, which utilizes the geometric features of key vectors.
Under these two metrics, we compare our approach against three allocation strategies: 
\begin{itemize}[leftmargin=*, align=left, nosep]
    \item \textbf{Uniform}: A static allocation that distributes the KV budget evenly across all layers.
    \item \textbf{PyramidKV}~\citep{cai_pyramidkv_2024}: A static allocation based on the \textit{Information Funneling} hypothesis, which progressively prunes the budget in deeper layers.
    \item \textbf{AdaKV}~\citep{feng_adakv_2024}: A dynamic allocation employing \textbf{Global Top-$k$} selection, based on the assumption that heads with higher importance scores warrant larger budgets.
\end{itemize}

In all experiments, \textbf{Full-KV} denotes the no-eviction upper bound retaining all KV pairs. Detailed baseline specifications are deferred to Appendix~\ref{app:baseline}.

\paragraph{Experimental Settings.}
Our evaluations are conducted using the KVPress framework \citep{kvpress}, adopting a \textit{question-agnostic} compression protocol. In this setting, the context is compressed and the KV cache is evicted solely based on the input document, strictly before the arrival of any query. This paradigm closely mirrors real-world production environments where the prompt or context is pre-filled and cached for future unknown user interactions. Furthermore, by precluding access to query-specific attention, this setup imposes a significantly more rigorous test on the method's ability to retain salient information compared to query-aware approaches.

\subsection{Main Results on KV Cache Eviction}

\paragraph{Results on LongBench.}
Table~\ref{tab:longbench_0.8} summarizes the performance under an 80\% compression ratio. 
Consistent with our global optimization objective in Eq.~\ref{eq:combinatorial_optimization}, our method effectively minimizes the aggregate eviction loss, translating into significant accuracy gains. 
On \texttt{Mistral-7B-v0.3} with $\pi_2$ (KeyDiff), our method improves the average accuracy from 40.54 (AdaKV) to 46.21, recovering 84\% of the performance gap between the compressed model and the Full-KV upper bound. 
Crucially, these gains are robust across diverse domains, ranging from summarization to synthetic tasks, demonstrating that our learned compression profiles successfully capture the intrinsic \textit{Oracle Importance} distribution across varying data densities.

%%%%%%%%%%%%%%%%%%%%%%%
% Abla
%5%%%%%%%%%%%%%%%%%%%%%
\begin{table*}[t]
\centering
\small 
\setlength{\tabcolsep}{1.8pt} % 紧凑的列间距
\renewcommand{\arraystretch}{1.15}
% \caption{\textbf{Ablation Study on LongBench.} We evaluate the impact of the optimality gap by comparing strategies \textbf{w/o Optimality Gap ($\bigtriangleup $)} against \textbf{w/ Optimality Gap}. Evaluated on \textbf{Mistral-7B-v0.3} and \textbf{Qwen2.5-32B} at an 80\% compression ratio.}
% \camblue{
\caption{\textbf{Ablation Study on Optimality Gap Loss.} We evaluate whether explicitly considering the optimality gap loss ($\Delta$) improves budget allocation, comparing variants w/o $\Delta$ and w/ $\Delta$. Evaluated on Mistral-7B-v0.3 and Qwen2.5-32B at an 80\% compression ratio.}
% }
\label{tab:longbench_ablation_optim_gap}
\resizebox{\textwidth}{!}{%
\begin{tabular}{llccccccccccccccccc}
\toprule
% ================= HEADER =================
\multirow{2}{*}{\textbf{Model}} & \multirow{2}{*}{\textbf{Method}} & 
\multicolumn{3}{c}{Single-Doc QA} & \multicolumn{3}{c}{Multi-Doc QA} & \multicolumn{3}{c}{Summarization} & \multicolumn{3}{c}{Few-shot} & \multicolumn{2}{c}{Synthetic} & \multicolumn{2}{c}{Code} & \multirow{2}{*}{\textbf{Avg}} \\
\cmidrule(lr){3-5} \cmidrule(lr){6-8} \cmidrule(lr){9-11} \cmidrule(lr){12-14} \cmidrule(lr){15-16} \cmidrule(lr){17-18}
& & 
\rotatebox{45}{\scriptsize NrtvQA} & \rotatebox{45}{\scriptsize Qasper} & \rotatebox{45}{\scriptsize MF-en} & 
\rotatebox{45}{\scriptsize Hotpot} & \rotatebox{45}{\scriptsize 2WikiQA} & \rotatebox{45}{\scriptsize Musique} & 
\rotatebox{45}{\scriptsize GovRep} & \rotatebox{45}{\scriptsize QMSum} & \rotatebox{45}{\scriptsize MultiNews} & 
\rotatebox{45}{\scriptsize TREC} & \rotatebox{45}{\scriptsize TriviaQA} & \rotatebox{45}{\scriptsize SAMSum} & 
\rotatebox{45}{\scriptsize PCount} & \rotatebox{45}{\scriptsize PR-en} & 
\rotatebox{45}{\scriptsize Lcc} & \rotatebox{45}{\scriptsize RB-P} & 
 \\
\midrule

% =============================================
% Mistral-7B-Instruct-v0.3
% =============================================
\multirow{6}{*}{\rotatebox[origin=c]{90}{\textbf{Mistral-7B}}} 
 & Full-KV  & 27.04 & 38.30 & 49.75 & 49.11 & 36.68 & 27.69 & 34.64 & 25.55 & 26.40 & 76.50 & 88.96 & 47.11 & 5.50 & 97.00 & 65.60 & 60.92 & 47.30 \\
 \cmidrule(lr){2-19}
 
 % --- SnapKV Variants ---
 % & \textit{SnapKV Selection} & & & & & & & & & & & & & & & & & \\
 & \hspace{1em} LU-KV-$\pi_1$ (w/o $\bigtriangleup $) & 23.03 & 27.14 & 42.68 & \textbf{49.69} & 31.27 & 21.20 & 30.34 & 23.15 & 23.64 & 66.00 & 89.36 & \textbf{47.13} & 5.00 & 96.50 & \textbf{66.72} & 60.33 & 43.95 \\
 & \hspace{1em} LU-KV-$\pi_1$ (w/ $\bigtriangleup $) & \textbf{25.25} & \textbf{34.91} & \textbf{51.32} & 48.87 & \textbf{38.10} & \textbf{22.80} & \textbf{33.57} & \textbf{25.02} & \textbf{25.31} & \textbf{71.00} & \textbf{91.32} & 47.12 & \textbf{5.19} & \textbf{97.50} & 53.76 & \textbf{61.62} & \textbf{45.79} \\
 
 % \addlinespace[4pt] 
 
 % --- KeyDiff Variants ---
 % & \textit{KeyDiff Selection} & & & & & & & & & & & & & & & & & \\
 \cmidrule(lr){2-19}
 & \hspace{1em} LU-KV-$\pi_2$ (w/o $\bigtriangleup $) & 24.82 & 29.97 & 47.07 & 44.72 & 34.68 & 23.10 & 30.62 & 24.15 & 24.53 & 47.50 & \textbf{89.06} & 46.80 & 4.68 & 92.50 & 46.63 & 60.89 & 41.98 \\
 & \hspace{1em} LU-KV-$\pi_2$ (w/ $\bigtriangleup $) & \textbf{25.80} & \textbf{39.78} & \textbf{53.82} & \textbf{48.26} & \textbf{41.33} & \textbf{24.69} & \textbf{33.49} & \textbf{25.52} & \textbf{25.65} & \textbf{69.00} & 88.81 & \textbf{47.14} & \textbf{6.53} & \textbf{97.50} & \textbf{51.18} & 60.89 & \textbf{46.21} \\
\midrule

% =============================================
% Qwen2.5-32B-Instruct
% =============================================
\multirow{6}{*}{\rotatebox[origin=c]{90}{\textbf{Qwen2.5-32B}}} 
 & Full-KV & 30.68 & 45.93 & 52.13 & 63.00 & 60.75 & 38.71 & 32.43 & 24.51 & 25.06 & 72.00 & 88.71 & 46.01 & 11.50 & 100.00 & 50.72 & 33.98 & 48.51 \\
 \cmidrule(lr){2-19}
 
 % --- SnapKV Variants ---
 % & \textit{SnapKV Selection} & & & & & & & & & & & & & & & & & \\
 & \hspace{1em} LU-KV-$\pi_1$ (w/o $\bigtriangleup $) & 27.50 & 26.07 & 36.90 & 59.58 & 53.91 & 37.27 & 29.37 & 20.76 & 22.32 & 65.50 & 88.57 & \textbf{45.45} & 9.50 & 99.25 & \textbf{60.36} & \textbf{37.77} & 45.00 \\
 & \hspace{1em} LU-KV-$\pi_1$ (w/ $\bigtriangleup $) & \textbf{29.41} & \textbf{39.16} & \textbf{50.95} & \textbf{62.82} & \textbf{58.00} & \textbf{39.84} & \textbf{31.34} & \textbf{23.12} & \textbf{24.10} & \textbf{71.00} & \textbf{88.89} & 42.07 & 9.50 & \textbf{100.00} & 60.21 & 36.71 & \textbf{47.95} \\
 
 % \addlinespace[4pt]
 
 % --- KeyDiff Variants ---
 % & \textit{KeyDiff Selection} & & & & & & & & & & & & & & & & & \\
 \cmidrule(lr){2-19}
 & \hspace{1em} LU-KV-$\pi_2$ (w/o $\bigtriangleup $) & 26.65 & 26.11 & 42.18 & 57.12 & 52.08 & 30.64 & 28.25 & 22.33 & 21.16 & 73.50 & \textbf{88.69} & \textbf{43.64} & \textbf{8.50} & 91.54 & 42.14 & 36.31 & 43.18 \\
 & \hspace{1em} LU-KV-$\pi_2$ (w/ $\bigtriangleup $) & \textbf{31.30} & \textbf{42.88} & \textbf{50.55} & \textbf{61.61} & \textbf{59.67} & \textbf{41.41} & \textbf{31.56} & \textbf{24.01} & \textbf{24.25} & \textbf{74.00} & 88.31 & 41.28 & 7.50 & \textbf{100.00} & \textbf{55.45} & \textbf{38.37} & \textbf{48.26} \\
\bottomrule
\end{tabular}%
}
\end{table*}

\paragraph{Results on RULER.}
The RULER benchmark serves as a stress test for retrieval robustness in extreme contexts. 
Focusing on \texttt{Mistral-7B-v0.3} using the SnapKV metric ($\pi_1$) in Table~\ref{tab:ruler_16k_0.8}, conventional strategies struggle significantly: Uniform allocation collapses to 29.53\% average accuracy, and AdaKV provides only marginal relief at 37.48\%. 
In contrast, our approach achieves a remarkable 69.98\% average accuracy under the same 80\% compression ratio. 
Notably, on the challenging \texttt{multikey-3} task, our method boosts performance from 1.00\% (Uniform) to 67.40\%, demonstrating substantial robustness in preserving sparse yet critical information.

\subsection{LU-KV Achieves Optimal Global Allocation}
To validate our core hypothesis, we visualize the eviction loss 
distribution under an 80\% compression ratio. 
As illustrated in Figures~\ref{fig:alloc_trend} 
and~\ref{fig:alloc_heatmap}, the failure modes of existing 
strategies reveal fundamental limitations in both optimization 
granularity and deployment constraints.

% To validate our core hypothesis, we visualize the eviction loss distribution under an 80\% compression ratio. As illustrated in Figure~\ref{fig:alloc_trend} and Figure~\ref{fig:alloc_heatmap}, the failure modes of different strategies reveal their fundamental limitations in both optimization granularity and deployment constraints.

\textbf{Limitations of PyramidKV.}
PyramidKV optimizes along the layer-wise dimension 
based on fixed structural priors.
% PyramidKV primarily attempts to optimize in the layer-wise dimension based on fixed priors. 
While it adjusts the budget distribution across layers, Figure~\ref{fig:alloc_trend} (pink line) shows that this rigid heuristic induces a sharp escalation in loss within deeper layers (layers 27--32), failing to adapt to the high semantic density of these regions. 
Consequently, the aggressive pruning in deep layers—based on an ill-suited pyramidal hypothesis—causes irreversible context loss that outweighs the minor gains in shallow layers, ultimately yielding a higher aggregated Oracle Loss than the Uniform baseline.

\textbf{Limitations of AdaKV.}
AdaKV focuses on the head-wise dimension, allocating budgets dynamically based on proxy score magnitudes. 
However, it faces a critical engineering trade-off: performing a true global cross-layer sort requires buffering all KV states during prefill, which causes unacceptable peak memory spikes. 
Consequently, AdaKV is often practically constrained to layer-wise uniform (or locally dynamic) budgets while competing only within layers.
This explains why its layer-wise loss curve (Figure~\ref{fig:alloc_trend}, blue dashed line) closely mirrors the Uniform baseline, failing to rebalance resources across layers.
Furthermore, within layers, Figure~\ref{fig:alloc_heatmap} reveals that distinct ``loss bursts'' (dark red blocks) persist. 
This confirms the ranking discordance: simply prioritizing heads with high proxy scores fails to capture true Oracle importance, leading to suboptimal intra-layer allocation.

\textbf{Superiority of LU-KV via Optimality-Gap-Aware Allocation.}
LU-KV integrates the advantages of both dimensions 
while circumventing their respective drawbacks.
As an offline static strategy, our method retrieves the optimal configuration from a pre-computed profile. This allows us to execute true global optimization (cross-layer and cross-head) without incurring the runtime memory overhead that limits online dynamic methods.
Results show distinct improvements in two aspects:
(1) Cross-Layer: Figure~\ref{fig:alloc_trend} (orange solid line) shows that our method effectively homogenizes the eviction loss across all layers, preventing the surge in deeper layers observed in PyramidKV.
(2) Cross-Head: Figure~\ref{fig:alloc_heatmap} confirms that the localized ``loss bursts'' characteristic of AdaKV are successfully eliminated.
By globally optimizing the Effective Marginal Gain 
($g_{\ell,h}^{\pi}$), LU-KV achieves superior resource 
utilization, significantly reducing the total Oracle 
Eviction Loss relative to AdaKV 
($3.87\times10^{-4}$ vs.\ $4.98\times10^{-4}$).

% By optimizing the Effective Marginal Gain ($g_{\ell,h}^{\pi}$) globally, we achieve superior resource utilization, significantly reducing the total Oracle Eviction Loss compared to AdaKV ($3.87\times10^{-4}$ vs. $4.98\times10^{-4}$).

\subsection{Ablation Study} \label{subsec:ablation}

We conduct an ablation study on the optimality gap loss (Eq.~\ref{eq:loss_decomp}) to assess its contribution to LU-KV. As shown in Table~\ref{tab:longbench_ablation_optim_gap}, removing
the optimality gap allocation leads to substantial
performance degradation across various metrics and base models.
This result validates our theoretical derivation:
the proxy metric for allocation is inherently suboptimal
due to its deviation from the oracle importance (\textit{i.e.}, the maximum potential contribution of a cached token over a future decoding window), and explicitly incorporating this gap into the optimization
objective effectively improves KV cache allocation.

\normalcolor

\subsection{Efficiency Analysis}
Figure~\ref{fig:efficiency} presents an efficiency comparison 
in terms of peak memory usage and Time To First Token (TTFT) 
latency. LU-KV achieves peak memory reduction and TTFT latency comparable to the SnapKV baseline, confirming that our method introduces negligible computational overhead. This demonstrates that LU-KV effectively operates within strict resource constraints while significantly mitigating the performance degradation caused by KV cache compression.

\section{Conclusion}

In this paper, we theoretically analyze the limitations of existing heuristic KV eviction methods through the lens of long-horizon inference, revealing their inability to capture the long-term cumulative contribution of tokens. 
To bridge this gap, we introduce a novel paradigm: KV cache retention should be determined not only by instantaneous importance but also by future utility. We formulate the head-level budget allocation as a global combinatorial optimization problem and propose an efficient convex-hull relaxation and a greedy solver algorithm to solve it.
Extensive evaluation across highly demanding benchmarks, such as LongBench and RULER, demonstrates the efficacy of our proposed approach.

\section*{Acknowledgement}
This work was supported by the Baige AI Team, Baidu Inc., through the Pinecone University Collaboration Program (Grant No. ZPJ2025001271-RPT1), and by the National Natural Science Foundation of China (NSFC) under Grant No. 62522206.

% Our approach achieves state-of-the-art performance on demanding benchmarks, including LongBench and RULER.

% \subsection{Ablation Study: The Role of Alignment Awareness}
% \label{subsec:ablation}

% To deconstruct the sources of performance gain, we compare two allocation logics (Table~\ref{tab:ablation}):
% \begin{enumerate}[leftmargin=*,align=left]
%     \item \textbf{Capacity-Only Allocation:} Allocates budget based solely on total Oracle importance (assuming perfect metric ranking, ignoring the $\Delta$ correction).
%     \item \textbf{Alignment-Aware Allocation (Ours):} Explicitly incorporates the Optimality Gap correction term.
% \end{enumerate}
% Results show that ignoring alignment leads to a significant performance drop (average score decrease of \textbf{X} points). This strongly supports our theoretical derivation: budget allocation should not depend on where the metric \textit{believes} importance lies (Method Confidence), but on where the metric \textit{actually} retrieves information correctly (Method Correctness).


% \clearpage

% \section{Impact Statement}
% This paper presents work whose goal is to advance the field of Machine
% Learning. There are many potential societal consequences of our work, none
% which we feel must be specifically highlighted here.
\section*{Impact Statement}
% \camblue{
This work aims to make long-context LLM inference more memory- and latency-efficient by improving KV-cache budget allocation. More efficient cache use can lower serving cost and enable longer-context applications on constrained hardware.
The main deployment risk is that aggressive compression can discard rare but important context, especially under domain shifts, context lengths, or decoding settings that differ from the profiling setup. Practitioners should validate the chosen compression ratio and profiling data for their target workload, and should avoid using cache compression as a substitute for safety, privacy, or reliability checks in downstream systems.

\bibliography{example_paper}
\bibliographystyle{icml2026}

%%%%%%%%%%%%%%%%%%%%%%%%%%%%%%%%%%%%%%%%%%%%%%%%%%%%%%%%%%%%%%%%%%%%%%%%%%%%%%%
%%%%%%%%%%%%%%%%%%%%%%%%%%%%%%%%%%%%%%%%%%%%%%%%%%%%%%%%%%%%%%%%%%%%%%%%%%%%%%%
% APPENDIX
%%%%%%%%%%%%%%%%%%%%%%%%%%%%%%%%%%%%%%%%%%%%%%%%%%%%%%%%%%%%%%%%%%%%%%%%%%%%%%%
%%%%%%%%%%%%%%%%%%%%%%%%%%%%%%%%%%%%%%%%%%%%%%%%%%%%%%%%%%%%%%%%%%%%%%%%%%%%%%%
\newpage
\appendix
\onecolumn
%%%%%%%%%%%%%%%%%%%%%%%%%%%%%%%%%%%%%%%%%%%%%%%%%%%%%%%%%%%%%%%%%%%%%%%%%%%%%%%
%%%%%%%%%%%%%%%%%%%%%%%%%%%%%%%%%%%%%%%%%%%%%%%%%%%%%%%%%%%%%%%%%%%%%%%%%%%%%%%
% \section{Prove}

% \subsection{The Non-convexity property of Eviction Loss in Equation~\ref{eq:combinatorial_optimization}} 

\section{Theoretical Proofs}

\subsection{Non-convexity of Eviction Loss in Eq.~\ref{eq:combinatorial_optimization}}
\label{app:prove_non_convex_comb_opt}

% Fix an attention head $(\ell,h)$ and consider the discrete loss sequence
% $\left\{\mathcal{L}_{\ell,h}\!\left(\mathcal{M}_{\ell,h}^{\pi}(i)\right)\right\}_{i=0}^{T}$,
% where $\mathcal{M}_{\ell,h}^{\pi}(i)$ denotes the top-$i$ positions selected by $\pi$ in this head.
Fix an attention head $(\ell,h)$ and consider the discrete loss sequence
$\left\{\mathcal{L}_{\ell,h}\!\left(\mathcal{M}_{\ell,h}^{\pi}(i)\right)\right\}_{i=0}^{T}$,
where $\mathcal{M}_{\ell,h}^{\pi}(i)$ denotes the top-$i$ positions selected by $\pi$ in this head.
Assume $I_{\ell,h,j}\ge 0$ for all cached positions $j$, and that ties in $\pi$ are resolved by a fixed deterministic rule.
Since $\mathcal{M}_{\ell,h}^{\pi}(i-1)\subset \mathcal{M}_{\ell,h}^{\pi}(i)$ and
$\left|\mathcal{M}_{\ell,h}^{\pi}(i)\setminus \mathcal{M}_{\ell,h}^{\pi}(i-1)\right|=1$ for all $i\ge 1$,
by Eq.~\eqref{eq:head_loss_def} we have
\begin{align}
\mathcal{L}_{\ell,h}\!\left(\mathcal{M}_{\ell,h}^{\pi}(i)\right)
&=
\sum_{j\notin \mathcal{M}_{\ell,h}^{\pi}(i)} I_{\ell,h,j} \nonumber\\
&=
\sum_{j\notin \mathcal{M}_{\ell,h}^{\pi}(i-1)} I_{\ell,h,j}
-
\sum_{j\in \mathcal{M}_{\ell,h}^{\pi}(i)\setminus \mathcal{M}_{\ell,h}^{\pi}(i-1)} I_{\ell,h,j} \nonumber\\
&=
\mathcal{L}_{\ell,h}\!\left(\mathcal{M}_{\ell,h}^{\pi}(i-1)\right)
-
\sum_{j\in \mathcal{M}_{\ell,h}^{\pi}(i)\setminus \mathcal{M}_{\ell,h}^{\pi}(i-1)} I_{\ell,h,j}.
\label{eq:proof_first_diff_setsum}
\end{align}
Therefore, the discrete first difference is non-positive:
\begin{equation}
\label{eq:proof_first_diff_nonpos}
\mathcal{L}_{\ell,h}\!\left(\mathcal{M}_{\ell,h}^{\pi}(i)\right)
-
\mathcal{L}_{\ell,h}\!\left(\mathcal{M}_{\ell,h}^{\pi}(i-1)\right)
=
-
\sum_{j\in \mathcal{M}_{\ell,h}^{\pi}(i)\setminus \mathcal{M}_{\ell,h}^{\pi}(i-1)} I_{\ell,h,j}
\le 0.
\end{equation}
The discrete second difference satisfies
\begin{align}
&\mathcal{L}_{\ell,h}\!\left(\mathcal{M}_{\ell,h}^{\pi}(i+1)\right)
-2\mathcal{L}_{\ell,h}\!\left(\mathcal{M}_{\ell,h}^{\pi}(i)\right)
+\mathcal{L}_{\ell,h}\!\left(\mathcal{M}_{\ell,h}^{\pi}(i-1)\right) \nonumber\\
&=
\Bigl(
\mathcal{L}_{\ell,h}\!\left(\mathcal{M}_{\ell,h}^{\pi}(i+1)\right)
-\mathcal{L}_{\ell,h}\!\left(\mathcal{M}_{\ell,h}^{\pi}(i)\right)
\Bigr)
-
\Bigl(
\mathcal{L}_{\ell,h}\!\left(\mathcal{M}_{\ell,h}^{\pi}(i)\right)
-\mathcal{L}_{\ell,h}\!\left(\mathcal{M}_{\ell,h}^{\pi}(i-1)\right)
\Bigr) \nonumber\\
&=
\sum_{j\in \mathcal{M}_{\ell,h}^{\pi}(i)\setminus \mathcal{M}_{\ell,h}^{\pi}(i-1)} I_{\ell,h,j}
-
\sum_{j\in \mathcal{M}_{\ell,h}^{\pi}(i+1)\setminus \mathcal{M}_{\ell,h}^{\pi}(i)} I_{\ell,h,j}.
\label{eq:proof_second_diff_setsum}
\end{align}
% Hence, $\mathcal{L}_{\ell,h}\!\left(\mathcal{M}_{\ell,h}^{\pi}(i)\right)$ is (discretely) convex in $i$
% if and only if the increment added at step $i$ has no smaller oracle importance than the increment added at step $i+1$,
% i.e., the oracle importances are non-increasing along the ordering induced by $\pi$.
% For a heuristic metric $\pi$, this monotonicity generally fails (there exist inversions w.r.t.\ $I_{\ell,h,:}$),
% which implies that there exists some $i$ such that the right-hand side of Eq.~\eqref{eq:proof_second_diff_setsum} is negative.
% Consequently, $\mathcal{L}_{\ell,h}\!\left(\mathcal{M}_{\ell,h}^{\pi}(b_{\ell,h})\right)$ is non-convex as a function of $b_{\ell,h}$ in general,
% and Eq.~\eqref{eq:combinatorial_optimization} is a non-convex combinatorial optimization problem.
Hence, $\mathcal{L}_{\ell,h}\!\left(\mathcal{M}_{\ell,h}^{\pi}(i)\right)$ is (discretely) convex in $i$
if and only if the oracle importances are non-increasing along the ordering induced by $\pi$.
For a heuristic metric $\pi$, this condition is not guaranteed. Whenever the ordering induced by $\pi$ contains an adjacent inversion with respect to $I_{\ell,h,:}$, the right-hand side of Eq.~\eqref{eq:proof_second_diff_setsum} becomes negative for that $i$.
Consequently, the raw loss sequence $\mathcal{L}_{\ell,h}\!\left(\mathcal{M}_{\ell,h}^{\pi}(b_{\ell,h})\right)$ is non-convex in general, and Eq.~\eqref{eq:combinatorial_optimization} is a non-convex discrete combinatorial allocation problem.

\subsection{Convex Relaxation Optimization of Equation~\ref{eq:combinatorial_optimization}}
\label{app:convex_relaxation}

Eq.~\eqref{eq:combinatorial_optimization} is a discrete multi-head budget allocation problem.
As shown in Appendix~\ref{app:prove_non_convex_comb_opt}, for a heuristic metric $\pi$,
$\mathcal{L}_{\ell,h}\!\left(\mathcal{M}_{\ell,h}^{\pi}(b_{\ell,h})\right)$ is generally non-convex in $b_{\ell,h}$.
We adopt the convex-hull relaxation described in Section~\ref{subsec:allocation} to obtain a tractable surrogate objective.

\paragraph{Convex surrogate loss by PAVA.}
% For each head $(\ell,h)$, consider the raw discrete loss sequence
% $\left\{\mathcal{L}_{\ell,h}\!\left(\mathcal{M}_{\ell,h}^{\pi}(i)\right)\right\}_{i=0}^{T}$.
% Applying isotonic regression via PAVA yields a convex, non-increasing surrogate sequence
% $\left\{\breve{\mathcal{L}}_{\ell,h}\!\left(\mathcal{M}_{\ell,h}^{\pi}(i)\right)\right\}_{i=0}^{T}$,
% as defined in Section~\ref{subsec:allocation}.
For each head $(\ell,h)$, consider the raw discrete loss sequence
$\left\{\mathcal{L}_{\ell,h}\!\left(\mathcal{M}_{\ell,h}^{\pi}(i)\right)\right\}_{i=0}^{T}$.
We first compute its raw marginal decreases and apply isotonic regression via PAVA to project them onto a non-negative, non-increasing sequence. Reconstructing the loss by cumulative subtraction then yields a convex, non-increasing surrogate sequence
$\left\{\breve{\mathcal{L}}_{\ell,h}\!\left(\mathcal{M}_{\ell,h}^{\pi}(i)\right)\right\}_{i=0}^{T}$,
as defined in Section~\ref{subsec:allocation}.
We further define the effective marginal gain (Eq.~\eqref{eq:marginal_gain}) as
\begin{equation}
\label{eq:app_gain_restate}
g_{\ell,h}^{\pi}(i)
=
\breve{\mathcal{L}}_{\ell,h}\!\left(\mathcal{M}_{\ell,h}^{\pi}(i-1)\right)
-
\breve{\mathcal{L}}_{\ell,h}\!\left(\mathcal{M}_{\ell,h}^{\pi}(i)\right)
\ge 0,
\end{equation}
which is monotonically non-increasing in $i$.

\paragraph{Equivalent maximization form.}
By telescoping Eq.~\eqref{eq:app_gain_restate}, for any integer budget $b_{\ell,h}\in\{0,1,\dots,T\}$ we have
\begin{equation}
\label{eq:app_telescope}
\breve{\mathcal{L}}_{\ell,h}\!\left(\mathcal{M}_{\ell,h}^{\pi}(b_{\ell,h})\right)
=
\breve{\mathcal{L}}_{\ell,h}\!\left(\mathcal{M}_{\ell,h}^{\pi}(0)\right)
-
\sum_{i=1}^{b_{\ell,h}} g_{\ell,h}^{\pi}(i).
\end{equation}
Substituting Eq.~\eqref{eq:app_telescope} into Eq.~\eqref{eq:combinatorial_optimization_relaxd},
the relaxed minimization is equivalent to the following maximization:
\begin{equation}
\label{eq:app_relaxed_max}
\max_{\{b_{\ell,h}\}}
\ \sum_{\ell=1}^{L}\sum_{h=1}^{H}\sum_{i=1}^{b_{\ell,h}} g_{\ell,h}^{\pi}(i)
\quad
\text{s.t.}\quad
\sum_{\ell=1}^{L}\sum_{h=1}^{H} b_{\ell,h}=B_{\text{total}}.
\end{equation}

\paragraph{Optimality of greedy allocation.}
Since $g_{\ell,h}^{\pi}(i)$ is non-increasing in $i$ for every head,
Eq.~\eqref{eq:app_relaxed_max} is a separable diminishing-returns allocation problem.
An optimal solution is obtained by iteratively allocating one unit of budget to the head $(\ell,h)$
that maximizes the next available gain $g_{\ell,h}^{\pi}(b_{\ell,h}+1)$, until the budget constraint is met.
Equivalently, the greedy procedure selects the $B_{\text{total}}$ largest feasible marginal gains across all heads
under the prefix constraint induced by $\{g_{\ell,h}^{\pi}(i)\}_{i}$, which yields the global optimum of
Eq.~\eqref{eq:combinatorial_optimization_relaxd}.

\section{Additional Experimental Results}

\subsection{Visualizing the Eviction Loss across Different Metrics}
\label{app:visualizing_optimality_gap}

\begin{figure}[htbp]
    \centering
    % --- 第一行: SnapKV ---
    \begin{subfigure}[b]{0.48\textwidth}
        \centering
        \includegraphics[width=\linewidth]{img/Mistral-v0.3_HotpotQA_r80_layer_SnapKV.pdf}
        \vspace{-0.2in}
        \caption{SnapKV: Layer-wise Aggregation}
        \label{fig:SnapKV_layer}
    \end{subfigure}
    \hfill
    \begin{subfigure}[b]{0.48\textwidth}
        \centering
        \includegraphics[width=\linewidth]{img/Mistral-v0.3_HotpotQA_r80_heatmap_SnapKV.pdf}
        \vspace{-0.2in}
        \caption{SnapKV: Per-head Heatmap}
        \label{fig:SnapKV_heatmap}
    \end{subfigure}

    \vspace{0.4cm} 

    % --- 第二行: KeyDiff ---
    \begin{subfigure}[b]{0.48\textwidth}
        \centering
        \includegraphics[width=\linewidth]{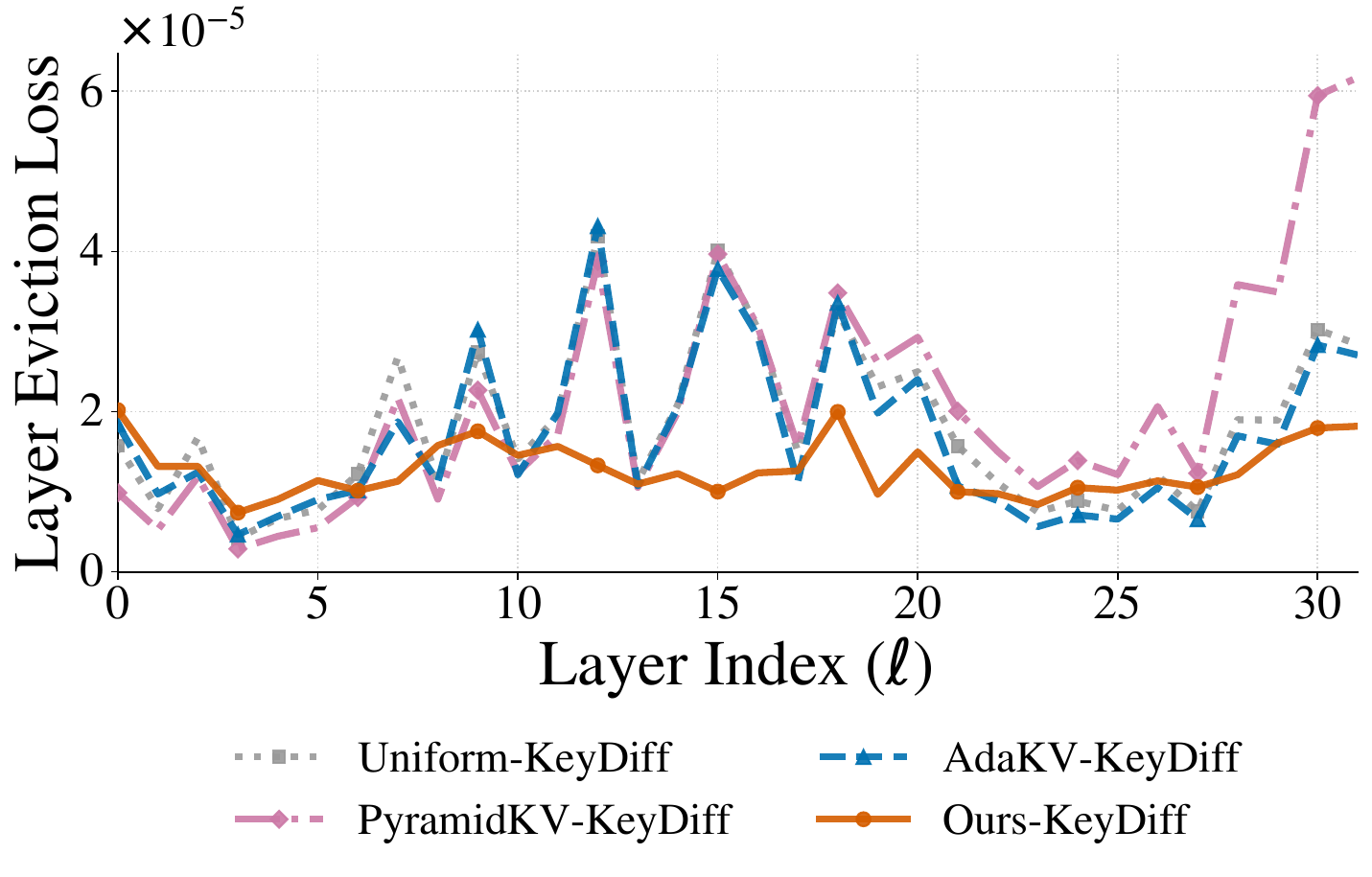}
        \vspace{-0.2in}
        \caption{KeyDiff: Layer-wise Aggregation}
        \label{fig:keydiff_layer}
    \end{subfigure}
    \hfill
    \begin{subfigure}[b]{0.48\textwidth}
        \centering
        \includegraphics[width=\linewidth]{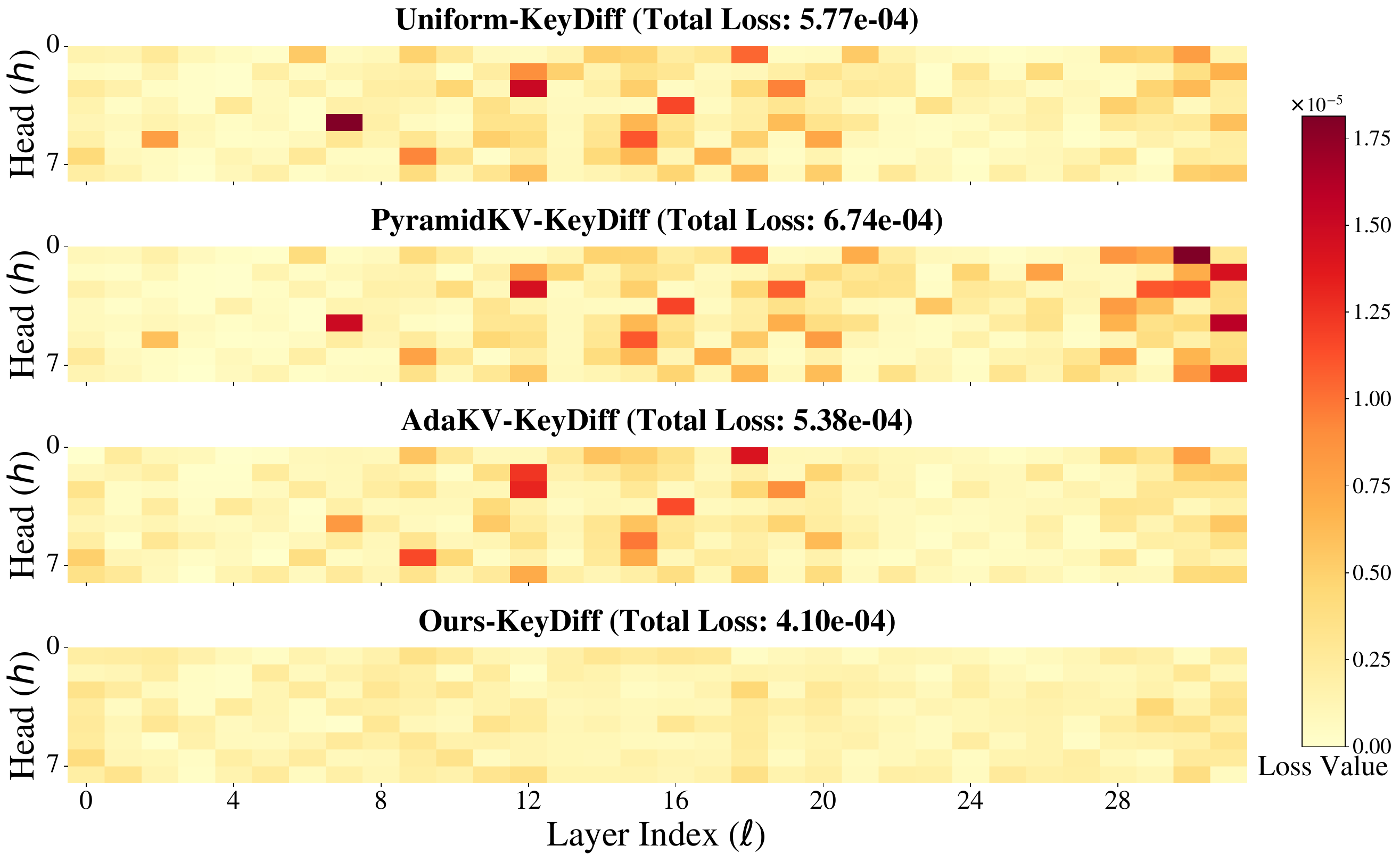}
        \vspace{-0.2in}
        \caption{KeyDiff: Per-head Heatmap}
        \label{fig:keydiff_heatmap}
    \end{subfigure}

    \vspace{0.4cm}

    % --- 第三行: EA ---
    \begin{subfigure}[b]{0.48\textwidth}
        \centering
        \includegraphics[width=\linewidth]{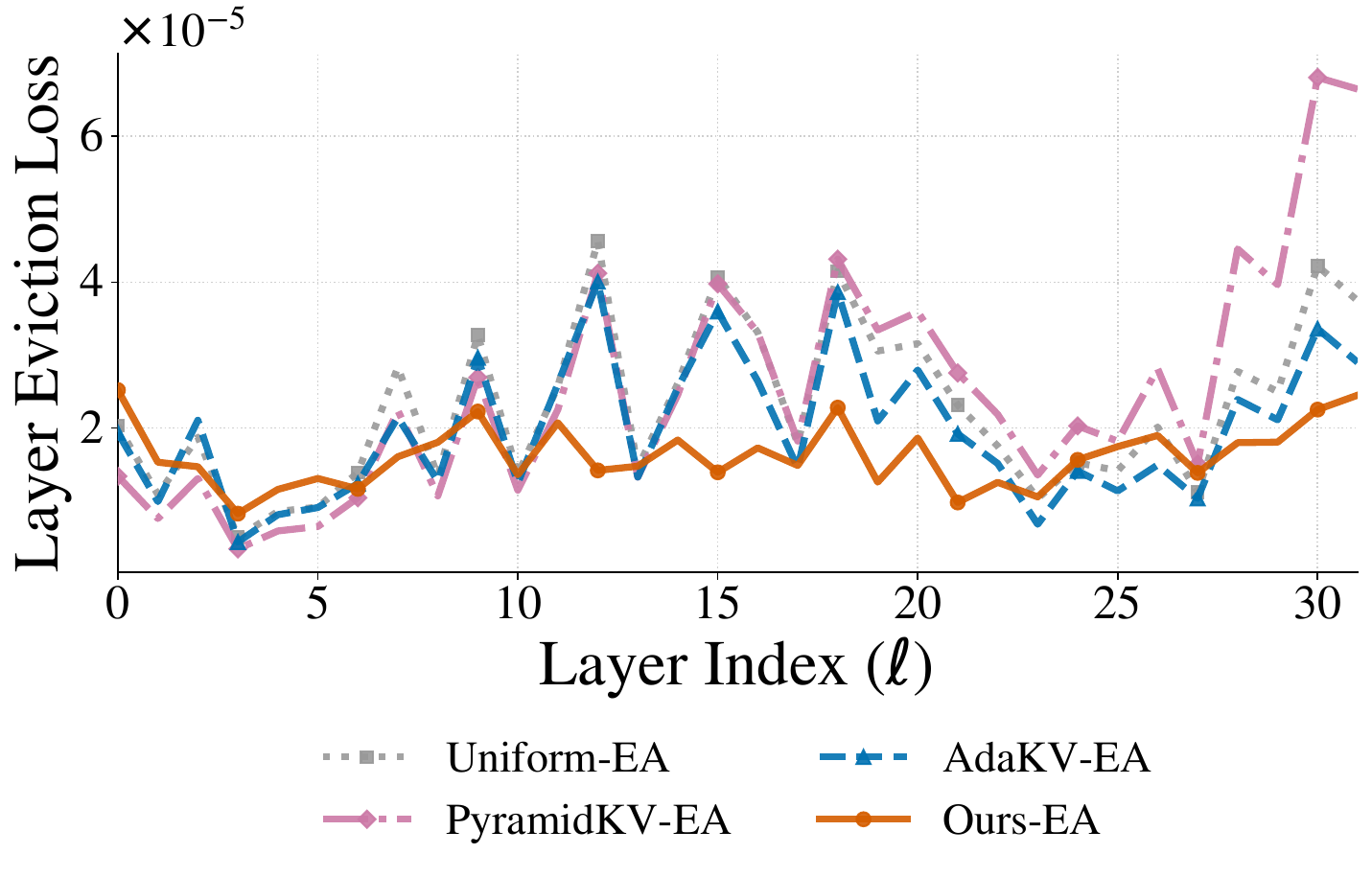}
        \vspace{-0.2in}
        \caption{EA: Layer-wise Aggregation}
        \label{fig:ea_layer}
    \end{subfigure}
    \hfill
    \begin{subfigure}[b]{0.48\textwidth}
        \centering
        \includegraphics[width=\linewidth]{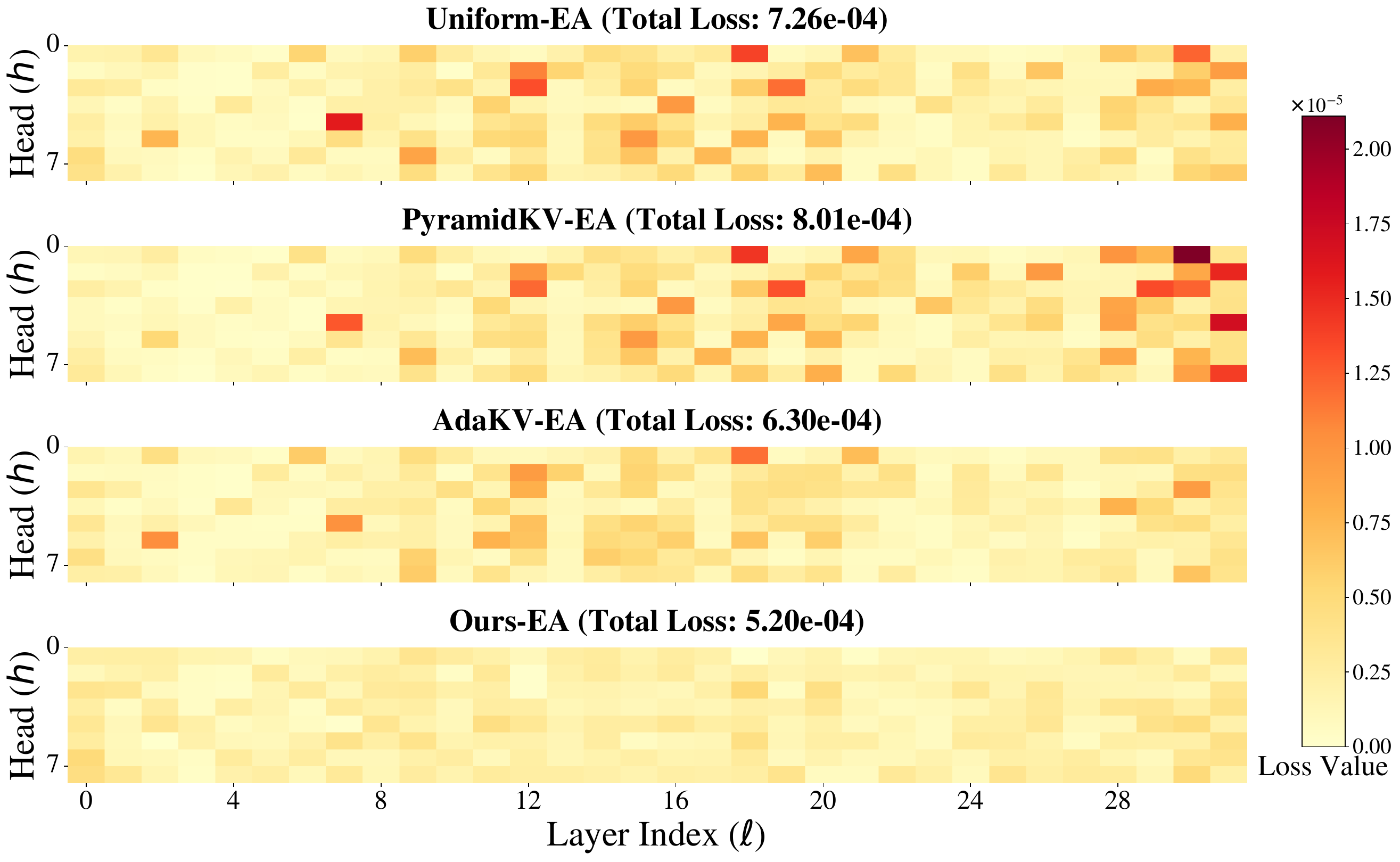}
        \vspace{-0.2in}
        \caption{EA: Per-head Heatmap}
        \label{fig:ea_heatmap}
    \end{subfigure}

    \caption{
        \textbf{Performance of Mistral-7B-v0.3 on HotpotQA (LongBench) across different metrics.} 
        The figure compares the aggregated layer-wise eviction loss (left column) and per-head loss distribution heatmaps (right column) for \textbf{SnapKV} (top row), \textbf{KeyDiff} (middle row), and \textbf{EA} (bottom row) at an 80\% global compression ratio.
    }
    \label{fig:appendix_full_analysis}
\end{figure}

To evaluate the universality of the \textit{optimality gap}, 
we visualize the eviction loss for the Mistral-7B-v0.3 model 
on the HotpotQA task (LongBench) under an 80\% global 
compression ratio. Figure~\ref{fig:appendix_full_analysis} 
illustrates the results across three heuristic metrics: 
SnapKV, KeyDiff, and EA.

\subsection{Comprehensive Head-wise Optimal Allocation Profiles}
\label{app:full_grid}
In this section, we provide the full visualization of the optimal budget allocation profiles for the \textit{Mistral-7B-v0.3} model using the KeyDiff metric. These figures display the mapping from the target global compression ratio to the allocated local compression ratio for each of the 32 layers and 8 heads.

The elements in the visualizations are defined as follows:
\begin{itemize}
    \item The horizontal axis ($x$-axis) represents the Global Compression Ratio ($\sigma \in [0, 1]$).
    \item The vertical axis ($y$-axis) represents the Optimal Local Compression Ratio ($r_{\ell,h}$) for the specific head.
    % \item The \textbf{black solid line} (labeled as `mytest convex') indicates the allocation curve derived from our proposed \textbf{convex-hull optimization} algorithm, calculated using our synthetic calibration data.
    % \item The \textbf{orange dashed line} (labeled as `mytest mckp') represents the theoretical optimal solution computed via the \textbf{Multi-Choice Knapsack Problem (MCKP)} dynamic programming algorithm on the synthetic calibration data.
    \item The black dashed line (labeled as `Our Method') indicates the allocation curve derived from our proposed convex-hull optimization algorithm, calculated using our synthetic calibration data.
    \item The orange dashed line (labeled as `Optimal DP') represents the theoretical optimal solution computed via the Multi-Choice Knapsack Problem (MCKP)dynamic programming algorithm on the synthetic calibration data.
    \item The background colored lines represent the corresponding utility profiles for various downstream datasets in the benchmark.
\end{itemize}

% --- Part 1: Layers 0-10 ---
\begin{figure*}[p]
    \centering
    \includegraphics[width=0.98\textwidth, height=0.9\textheight, keepaspectratio]{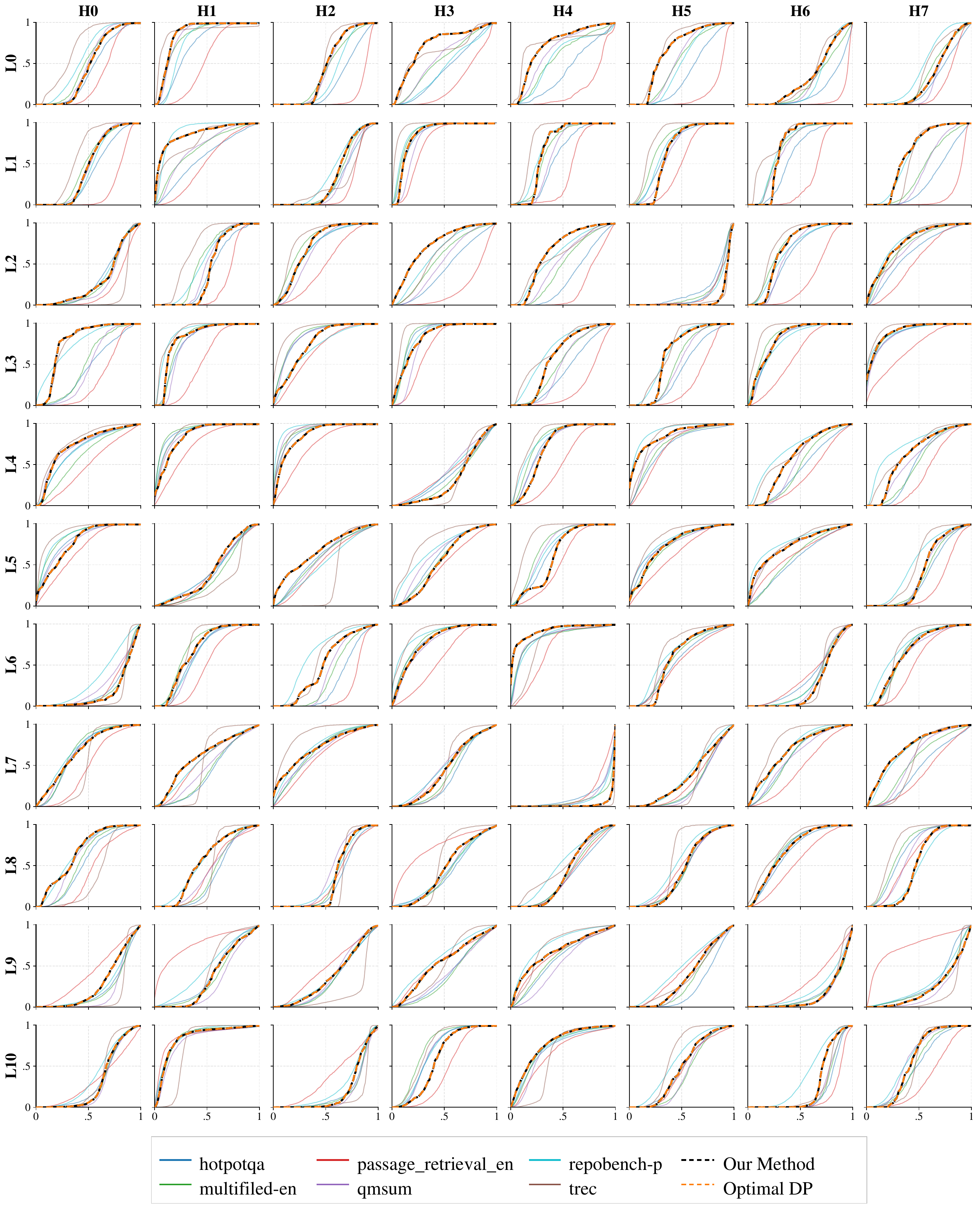}
    \caption{\textbf{Head-wise Optimal Allocation Profiles (Part I: Layers 0 to 10).} Visualization of the optimal local budget distribution for the \textbf{Mistral-7B-v0.3} model on the different tasks (LongBench) using the \textbf{KeyDiff} metric.}
    \label{fig:grid_p1}
\end{figure*}

% --- Part 2: Layers 11-21 ---
\begin{figure*}[p]
    \centering
    \includegraphics[width=0.98\textwidth, height=0.9\textheight, keepaspectratio]{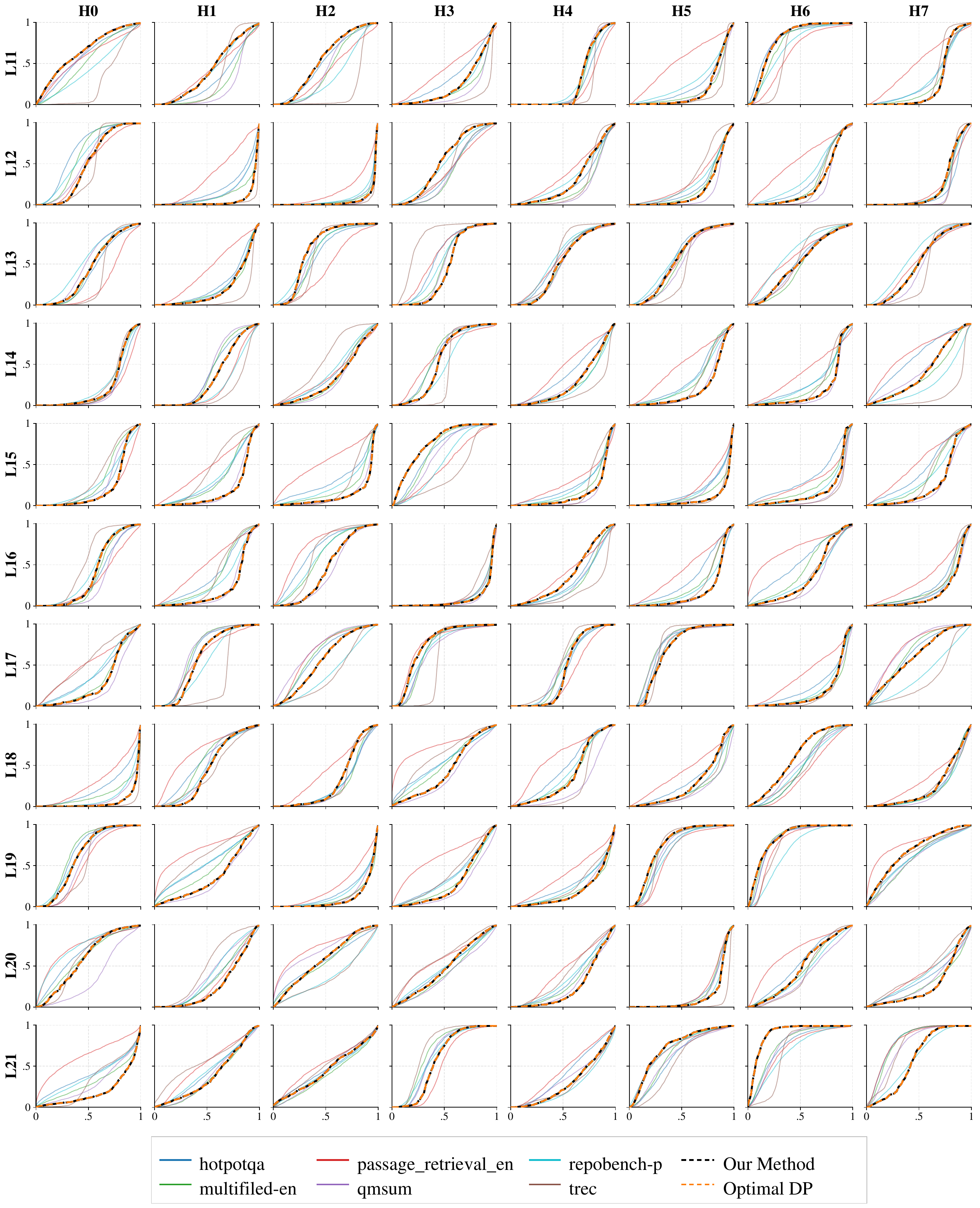}
    \caption{\textbf{Head-wise Optimal Allocation Profiles (Part II: Layers 11 to 21).} Visualization of the optimal local budget distribution for the \textbf{Mistral-7B-v0.3} model on the different tasks (LongBench) using the \textbf{KeyDiff} metric.}
    \label{fig:grid_p2}
\end{figure*}

% --- Part 3: Layers 22-31 ---
\begin{figure*}[p]
    \centering
    \includegraphics[width=0.98\textwidth, height=0.85\textheight, keepaspectratio]{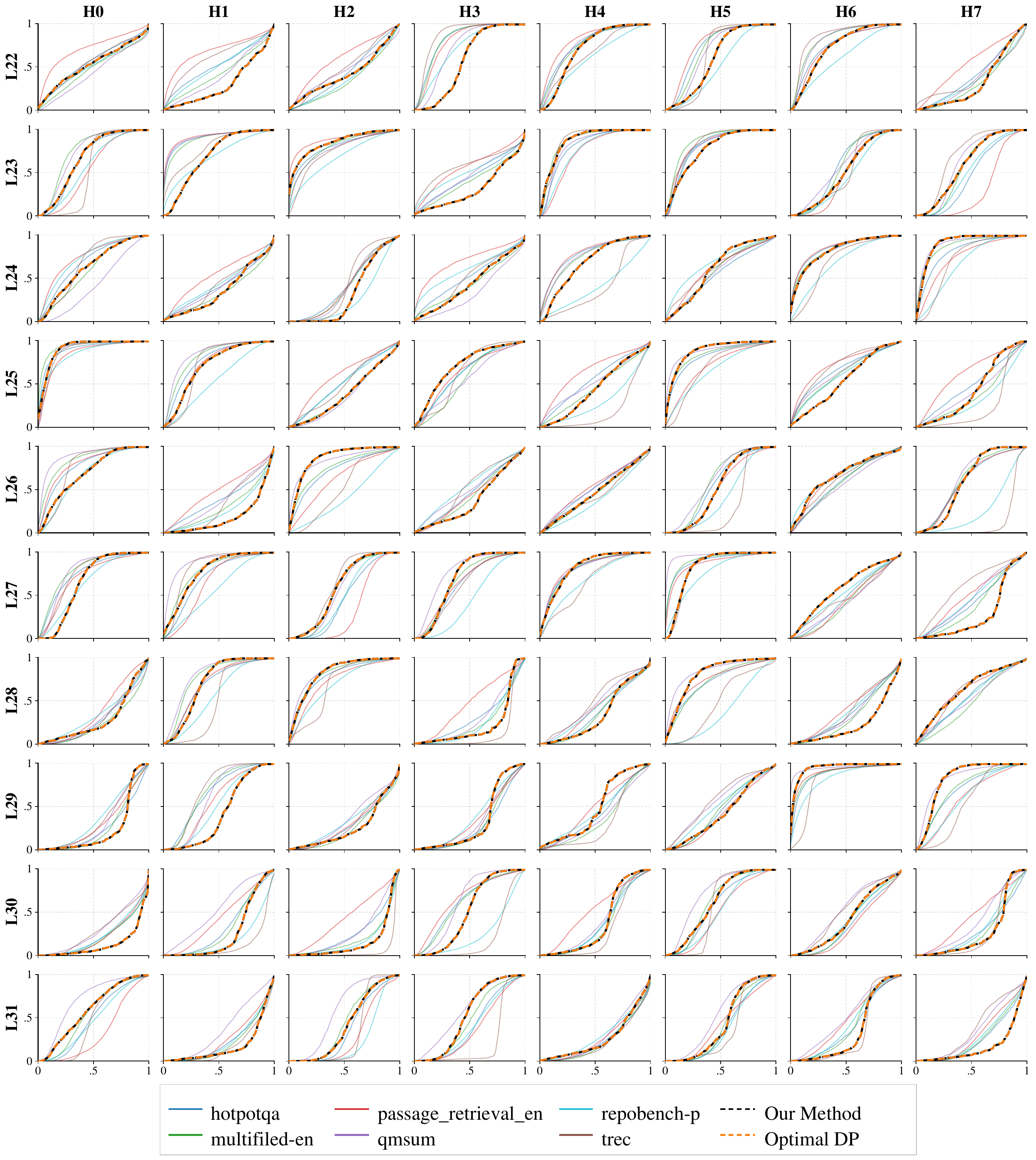}
    \caption{\textbf{Head-wise Optimal Allocation Profiles (Part III: Layers 22 to 31).} Visualization of the optimal local budget distribution for the \textbf{Mistral-7B-v0.3} model on the different tasks (LongBench) using the \textbf{KeyDiff} metric.}
    \label{fig:grid_p3}
\end{figure*}

\clearpage

\subsection{Detailed Scores Of LongBench}
\label{app:longbench_scores}

In this section, we provide a comprehensive breakdown of performance across all 16 datasets in \textit{LongBench}. Table~\ref{tab:longbench_appendix_0.5} presents the results under a 50\% global compression ratio for \textbf{SnapKV} ($\pi_1$) and \textbf{KeyDiff} ($\pi_2$) metrics. To further verify the universality of our approach under more aggressive compression, we also evaluate the performance at an 80\% global compression ratio in Table~\ref{tab:longbench_appendix_0.8}, extending our analysis to include the \textbf{EA} ($\pi_3$) metric. Across these diverse settings, different importance metrics, and various model scales, the results consistently demonstrate that our proposed method remains effective.

\begin{table*}[t]
\centering
\small 
\setlength{\tabcolsep}{1.8pt} % 紧凑的列间距
\renewcommand{\arraystretch}{1.15}

\caption{Detailed scores of 16 datasets on LongBench at an 50\% compression ratio.}
\label{tab:longbench_appendix_0.5}

\resizebox{\textwidth}{!}{%
\begin{tabular}{llccccccccccccccccc}
\toprule
% ================= HEADER =================
\multirow{2}{*}{\textbf{Model}} & \multirow{2}{*}{\textbf{Method}} & 
\multicolumn{3}{c}{Single-Doc QA} & \multicolumn{3}{c}{Multi-Doc QA} & \multicolumn{3}{c}{Summarization} & \multicolumn{3}{c}{Few-shot} & \multicolumn{2}{c}{Synthetic} & \multicolumn{2}{c}{Code} & \multirow{2}{*}{\textbf{Avg}} \\
\cmidrule(lr){3-5} \cmidrule(lr){6-8} \cmidrule(lr){9-11} \cmidrule(lr){12-14} \cmidrule(lr){15-16} \cmidrule(lr){17-18}
& & 
\rotatebox{45}{\scriptsize NrtvQA} & \rotatebox{45}{\scriptsize Qasper} & \rotatebox{45}{\scriptsize MF-en} & 
\rotatebox{45}{\scriptsize Hotpot} & \rotatebox{45}{\scriptsize 2WikiQA} & \rotatebox{45}{\scriptsize Musique} & 
\rotatebox{45}{\scriptsize GovRep} & \rotatebox{45}{\scriptsize QMSum} & \rotatebox{45}{\scriptsize MultiNews} & 
\rotatebox{45}{\scriptsize TREC} & \rotatebox{45}{\scriptsize TriviaQA} & \rotatebox{45}{\scriptsize SAMSum} & 
\rotatebox{45}{\scriptsize PCount} & \rotatebox{45}{\scriptsize PR-en} & 
\rotatebox{45}{\scriptsize Lcc} & \rotatebox{45}{\scriptsize RB-P} & 
 \\
\midrule

% =============================================
% Mistral-7B-Instruct-v0.3
% =============================================
\multirow{12}{*}{\rotatebox[origin=c]{90}{\textbf{Mistral-7B-v0.3}}} 
 & Full-KV  & 27.04 & 38.30 & 49.75 & 49.11 & 36.68 & 27.69 & 34.64 & 25.55 & 26.40 & 76.50 & 88.96 & 47.11 & 5.50 & 97.00 & 65.60 & 60.92 & 47.30 \\
 \cmidrule(lr){2-19}
 
 % --- SnapKV Variants ---
 & \textit{Metric SnapKV ($\pi_1$)} & & & & & & & & & & & & & & & & & \\
 & \hspace{1em} Uniform-$\pi_1$ & 24.51 & 32.14 & 42.98 & 48.71 & 34.72 & 24.64 & 32.07 & 23.68 & 25.10 & 68.50 & 88.91 & 47.18 & 5.50 & 96.50 & 65.36 & 60.51 & 45.06 \\
 & \hspace{1em} Pyramid-$\pi_1$ & 24.40 & 30.30 & 44.34 & 48.54 & 34.10 & 24.18 & 31.74 & 24.06 & 24.43 & 67.50 & 89.21 & 46.83 & 3.50 & 97.50 & 65.63 & 60.34 & 44.79 \\
 & \hspace{1em} Ada-$\pi_1$ & 24.47 & 31.50 & 43.61 & \textbf{50.00} & 35.93 & 25.43 & 31.51 & 24.27 & 25.11 & 72.00 & 88.94 & 47.29 & \textbf{6.50} & 96.50 & 65.35 & 60.87 & 45.58 \\
  & \hspace{1em} \textbf{LU-KV-$\pi_1$ (Ours)} & \textbf{25.81} & \textbf{38.93} & \textbf{50.53} & 49.20 & \textbf{36.82} & \textbf{27.05} & \textbf{34.96} & \textbf{25.64} & \textbf{26.42} & \textbf{76.00} & \textbf{89.45} & \textbf{47.33} & 5.54 & \textbf{98.00} & \textbf{66.14} & \textbf{61.62} & \textbf{47.46} \\
 
 \addlinespace[4pt] 
 
 % --- KeyDiff Variants ---
 & \textit{Metric KeyDiff ($\pi_2$)} & & & & & & & & & & & & & & & & & \\
 & \hspace{1em} Uniform-$\pi_2$ & 24.03 & 35.36 & 49.01 & 47.61 & 36.29 & 25.34 & 31.97 & 24.38 & 25.46 & 56.00 & 88.56 & 46.57 & 4.20 & 95.00 & 56.82 & 60.52 & 44.20 \\
 & \hspace{1em} Pyramid-$\pi_2$ & 27.26 & 34.78 & 46.46 & 45.38 & 35.94 & 25.21 & 31.73 & 24.84 & 25.35 & 68.00 & 89.21 & 47.31 & \textbf{6.50} & \textbf{96.00} & 47.92 & 60.37 & 44.52 \\
 & \hspace{1em} Ada-$\pi_2$ & 26.05 & 37.68 & \textbf{51.31} & 47.04 & 37.79 & \textbf{26.43} & 33.02 & 25.05 & 25.80 & 63.00 & 88.39 & 46.82 & 2.87 & 95.25 & \textbf{65.27} & 60.39 & 45.76 \\
 & \hspace{1em} \textbf{LU-KV-$\pi_2$ (Ours)} & \textbf{27.66} & \textbf{38.91} & 50.92 & \textbf{51.30} & \textbf{39.67} & 24.42 & \textbf{34.57} & \textbf{25.48} & \textbf{26.60} & \textbf{75.00} & \textbf{89.46} & \textbf{47.61} & 3.05 & 95.75 & 63.24 & \textbf{61.14} & \textbf{47.17} \\

\midrule

% =============================================
% Llama-3.1-8B-Instruct
% =============================================
\multirow{12}{*}{\rotatebox[origin=c]{90}{\textbf{Llama-3.1-8B}}} 
 & Full-KV  & 29.39 & 45.17 & 55.74 & 58.31 & 48.12 & 32.57 & 34.53 & 25.30 & 26.91 & 72.50 & 91.78 & 44.32 & 8.47 & 99.50 & 63.43 & 52.59 & 49.29 \\
 \cmidrule(lr){2-19}
 
 % --- SnapKV Variants ---
 & \textit{Metric SnapKV ($\pi_1$)} & & & & & & & & & & & & & & & & & \\
 & \hspace{1em} Uniform-$\pi_1$ & 26.46 & 39.37 & 48.32 & 56.53 & 44.99 & 30.41 & 31.75 & 23.65 & 25.30 & 62.50 & 92.31 & 44.01 & 7.00 & 99.50 & \textbf{66.01} & 53.89 & 47.00 \\
 & \hspace{1em} Pyramid-$\pi_1$ & 28.13 & 33.86 & 48.94 & 55.26 & 46.16 & 31.36 & 30.80 & 24.58 & 24.23 & 60.00 & \textbf{92.53} & 44.03 & 6.59 & 99.50 & 65.98 & 54.44 & 46.65 \\
 & \hspace{1em} Ada-$\pi_1$ & 29.32 & 40.23 & 51.32 & 56.01 & 44.67 & \textbf{32.38} & 31.82 & 24.23 & 25.40 & 68.50 & 91.90 & \textbf{44.19} & 7.89 & 99.50 & 64.92 & 54.31 & 47.91 \\
 & \hspace{1em} \textbf{LU-KV-$\pi_1$ (Ours)} & \textbf{30.57} & \textbf{44.45} & \textbf{55.59} & \textbf{56.91} & \textbf{47.32} & 32.33 & \textbf{34.64} & \textbf{25.40} & \textbf{26.64} & \textbf{71.50} & 91.65 & 44.14 & \textbf{7.95} & \textbf{100.00} & 65.25 & \textbf{55.77} & \textbf{49.38} \\
 
 \addlinespace[4pt] 
 
 % --- KeyDiff Variants ---
 & \textit{Metric KeyDiff ($\pi_2$)} & & & & & & & & & & & & & & & & & \\
 & \hspace{1em} Uniform-$\pi_2$ & 30.58 & 41.13 & 52.39 & 55.78 & 43.72 & 29.92 & 32.73 & 24.68 & 25.07 & 65.00 & \textbf{92.03} & 45.07 & \textbf{7.83} & 99.50 & 55.89 & 54.58 & 47.24 \\
 & \hspace{1em} Pyramid-$\pi_2$ & 30.66 & 41.15 & 52.06 & \textbf{56.13} & 44.73 & 30.01 & 31.85 & 25.11 & 24.98 & 59.00 & 91.78 & \textbf{45.18} & 6.08 & 99.50 & 37.66 & 55.42 & 45.71 \\
 & \hspace{1em} Ada-$\pi_2$ & \textbf{31.24} & 43.64 & 51.53 & 52.68 & \textbf{48.46} & \textbf{31.66} & 34.06 & 24.38 & 26.06 & 70.00 & 91.11 & 44.80 & 7.08 & 99.50 & 63.68 & \textbf{56.85} & \textbf{48.55} \\
 & \hspace{1em} \textbf{LU-KV-$\pi_2$ (Ours)} & 29.78 & \textbf{44.14} & \textbf{54.23} & 54.67 & 44.59 & 27.43 & \textbf{34.71} & \textbf{25.35} & \textbf{26.63} & \textbf{74.00} & 91.89 & 44.52 & 5.79 & 99.50 & \textbf{63.75} & 54.52 & 48.47 \\

\midrule

% =============================================
% Qwen2.5-32B-Instruct
% =============================================
\multirow{12}{*}{\rotatebox[origin=c]{90}{\textbf{Qwen2.5-32B}}} 
 & Full-KV  & 30.68 & 45.93 & 52.13 & 63.00 & 60.75 & 38.71 & 32.43 & 24.51 & 25.06 & 72.00 & 88.71 & 46.01 & 11.50 & 100.00 & 50.72 & 33.98 & 48.51 \\
 \cmidrule(lr){2-19}
 
 % --- SnapKV Variants ---
 & \textit{Metric SnapKV ($\pi_1$)} & & & & & & & & & & & & & & & & & \\
 & \hspace{1em} Uniform-$\pi_1$ & 28.40 & 36.02 & 44.58 & 62.89 & 57.71 & 37.40 & 30.95 & 22.17 & 23.92 & 67.50 & 89.02 & 45.16 & \textbf{13.00} & 99.50 & 55.52 & 34.83 & 46.79 \\
 & \hspace{1em} Pyramid-$\pi_1$ & 25.81 & 22.00 & 36.10 & 57.36 & 49.10 & 33.74 & 28.95 & 21.71 & 21.41 & 58.00 & \textbf{89.03} & 45.55 & 10.00 & 100.00 & \textbf{60.03} & 35.60 & 43.40 \\
 & \hspace{1em} Ada-$\pi_1$ & 27.16 & 34.67 & 43.39 & 63.10 & 57.34 & 36.58 & 30.84 & 22.02 & 23.95 & 70.50 & 88.82 & \textbf{45.67} & 11.50 & 99.75 & 54.38 & 33.63 & 46.46 \\
 & \hspace{1em} \textbf{LU-KV-$\pi_1$ (Ours)} & \textbf{30.73} & \textbf{44.26} & \textbf{51.59} & \textbf{63.21} & \textbf{60.75} & \textbf{39.74} & \textbf{31.96} & \textbf{24.02} & \textbf{24.72} & \textbf{71.50} & 88.38 & 45.40 & 12.00 & 100.00 & 59.30 & \textbf{37.70} & \textbf{49.08} \\
 
 \addlinespace[4pt]
 
 % --- KeyDiff Variants ---
 & \textit{Metric KeyDiff ($\pi_2$)} & & & & & & & & & & & & & & & & & \\
 & \hspace{1em} Uniform-$\pi_2$ & 28.66 & 37.44 & 47.88 & 62.21 & 58.97 & 39.09 & 30.86 & 23.53 & 23.37 & \textbf{75.00} & 85.93 & \textbf{46.15} & 13.25 & 98.00 & 41.03 & 34.73 & 46.63 \\
 & \hspace{1em} Pyramid-$\pi_2$ & 27.84 & 30.57 & 40.25 & 55.92 & 50.81 & 32.99 & 28.69 & 22.47 & 21.50 & 65.50 & 86.66 & 45.14 & 9.67 & 75.08 & 25.86 & \textbf{35.40} & 40.90 \\
 & \hspace{1em} Ada-$\pi_2$ & 29.90 & 42.44 & 51.13 & 60.14 & 59.78 & 39.67 & 31.90 & 23.21 & 23.96 & 72.00 & 87.23 & 44.84 & \textbf{13.50} & 100.00 & 43.33 & 33.71 & 47.30 \\
 & \hspace{1em} \textbf{LU-KV-$\pi_2$ (Ours)} & \textbf{31.63} & \textbf{45.78} & \textbf{51.38} & \textbf{63.37} & \textbf{63.00} & \textbf{43.18} & \textbf{32.46} & \textbf{24.69} & \textbf{25.04} & 74.00 & \textbf{89.26} & 44.47 & 10.50 & 100.00 & \textbf{53.40} & 34.51 & \textbf{49.17} \\
\bottomrule
\end{tabular}%
}
\end{table*}

% ==========================================
% Detailed Results (Place in Appendix)
% ==========================================
\begin{table*}[t]
\centering
\small 
\setlength{\tabcolsep}{1.8pt} % 紧凑的列间距
\renewcommand{\arraystretch}{1.15}

\caption{Detailed scores of 16 datasets on LongBench at an 80\% compression ratio.}
\label{tab:longbench_appendix_0.8}

\resizebox{\textwidth}{!}{%
\begin{tabular}{llccccccccccccccccc}
\toprule
% ================= HEADER =================
\multirow{2}{*}{\textbf{Model}} & \multirow{2}{*}{\textbf{Method}} & 
\multicolumn{3}{c}{Single-Doc QA} & \multicolumn{3}{c}{Multi-Doc QA} & \multicolumn{3}{c}{Summarization} & \multicolumn{3}{c}{Few-shot} & \multicolumn{2}{c}{Synthetic} & \multicolumn{2}{c}{Code} & \multirow{2}{*}{\textbf{Avg}} \\
\cmidrule(lr){3-5} \cmidrule(lr){6-8} \cmidrule(lr){9-11} \cmidrule(lr){12-14} \cmidrule(lr){15-16} \cmidrule(lr){17-18}
& & 
\rotatebox{45}{\scriptsize NrtvQA} & \rotatebox{45}{\scriptsize Qasper} & \rotatebox{45}{\scriptsize MF-en} & 
\rotatebox{45}{\scriptsize Hotpot} & \rotatebox{45}{\scriptsize 2WikiQA} & \rotatebox{45}{\scriptsize Musique} & 
\rotatebox{45}{\scriptsize GovRep} & \rotatebox{45}{\scriptsize QMSum} & \rotatebox{45}{\scriptsize MultiNews} & 
\rotatebox{45}{\scriptsize TREC} & \rotatebox{45}{\scriptsize TriviaQA} & \rotatebox{45}{\scriptsize SAMSum} & 
\rotatebox{45}{\scriptsize PCount} & \rotatebox{45}{\scriptsize PR-en} & 
\rotatebox{45}{\scriptsize Lcc} & \rotatebox{45}{\scriptsize RB-P} & 
 \\
\midrule

% =============================================
% Mistral-7B-Instruct-v0.3
% =============================================
\multirow{16}{*}{\rotatebox[origin=c]{90}{\textbf{Mistral-7B-v0.3}}} 
 & Full-KV  & 27.04 & 38.30 & 49.75 & 49.11 & 36.68 & 27.69 & 34.64 & 25.55 & 26.40 & 76.50 & 88.96 & 47.11 & 5.50 & 97.00 & 65.60 & 60.92 & 47.30 \\
 \cmidrule(lr){2-19}
 
 % --- SnapKV Variants ---
 & \textit{Metric SnapKV ($\pi_1$)} & & & & & & & & & & & & & & & & & \\
 & \hspace{1em} Uniform-$\pi_1$ & 21.37 & 19.10 & 33.64 & 44.22 & 29.19 & 21.93 & 28.20 & 22.06 & 22.20 & 55.00 & 90.07 & 46.79 & 5.00 & 90.00 & 63.37 & 58.59 & 40.67 \\
 & \hspace{1em} Pyramid-$\pi_1$ & 21.77 & 19.38 & 34.92 & 43.67 & 32.21 & 19.97 & 27.77 & 22.36 & 21.80 & 55.00 & 89.21 & 45.85 & 5.50 & 92.00 & 63.53 & 60.07 & 40.94 \\
 & \hspace{1em} Ada-$\pi_1$ & 20.29 & 22.31 & 36.03 & 44.43 & 29.09 & 22.12 & 27.98 & 22.88 & 22.41 & 62.50 & 90.07 & 46.99 & 5.00 & 94.00 & \textbf{64.47} & 59.62 & 41.89 \\
 & \hspace{1em} \textbf{LU-KV-$\pi_1$ (Ours)} & \textbf{25.25} & \textbf{34.91} & \textbf{51.32} & \textbf{48.87} & \textbf{38.10} & \textbf{22.80} & \textbf{33.57} & \textbf{25.02} & \textbf{25.31} & \textbf{71.00} & \textbf{91.32} & \textbf{47.12} & \textbf{5.19} & \textbf{97.50} & 53.76 & \textbf{61.62} & \textbf{45.79} \\
 
 \addlinespace[4pt] 
 
 % --- KeyDiff Variants ---
 & \textit{Metric KeyDiff ($\pi_2$)} & & & & & & & & & & & & & & & & & \\
 & \hspace{1em} Uniform-$\pi_2$ & 22.37 & 24.66 & 38.57 & 41.68 & 32.86 & 18.04 & 28.53 & 22.48 & 23.12 & 42.00 & 88.39 & 46.61 & 3.66 & 62.50 & 34.37 & 59.40 & 36.83 \\
 & \hspace{1em} Pyramid-$\pi_2$ & 22.61 & 26.74 & 38.96 & 39.81 & 34.85 & 16.92 & 28.23 & 22.71 & 22.55 & 42.00 & 89.56 & 46.42 & 5.11 & 63.00 & 26.95 & 58.94 & 36.59 \\
 & \hspace{1em} Ada-$\pi_2$ & 24.70 & 29.65 & 41.47 & 44.47 & 34.48 & 22.39 & 28.78 & 23.26 & 23.65 & 42.50 & \textbf{89.89} & 46.40 & 4.57 & 77.00 & \textbf{55.96} & 59.45 & 40.54 \\
 & \hspace{1em} \textbf{LU-KV-$\pi_2$ (Ours)} & \textbf{25.80} & \textbf{39.78} & \textbf{53.82} & \textbf{48.26} & \textbf{41.33} & \textbf{24.69} & \textbf{33.49} & \textbf{25.52} & \textbf{25.65} & \textbf{69.00} & 88.81 & \textbf{47.14} & \textbf{6.53} & \textbf{97.50} & 51.18 & \textbf{60.89} & \textbf{46.21} \\
 
 \addlinespace[4pt] 
 
 % --- EA Variants ---
 & \textit{Metric EA ($\pi_3$)} & & & & & & & & & & & & & & & & & \\
 & \hspace{1em} Uniform-$\pi_3$ & 12.87 & 30.45 & 40.25 & 28.62 & 23.59 & 11.32 & 21.72 & 22.81 & 24.54 & 2.00 & 13.28 & 17.38 & \textbf{4.67} & 3.55 & 17.44 & 48.62 & 20.19 \\
 & \hspace{1em} Pyramid-$\pi_3$ & 14.50 & 30.96 & 36.90 & 32.05 & 27.57 & 15.42 & 27.10 & 23.75 & 24.87 & \textbf{22.50} & 18.99 & \textbf{25.83} & 2.72 & 30.50 & 20.34 & 37.34 & 24.46 \\
 & \hspace{1em} Ada-$\pi_3$ & 11.08 & 18.20 & 41.28 & 26.25 & 26.43 & 10.27 & 23.24 & 22.27 & 24.43 & 2.50 & 23.75 & 13.86 & 3.32 & 9.83 & \textbf{29.84} & 50.74 & 21.08 \\
 & \hspace{1em} \textbf{LU-KV-$\pi_3$ (Ours)} & \textbf{22.65} & \textbf{36.86} & \textbf{50.81} & \textbf{47.45} & \textbf{36.18} & \textbf{26.65} & \textbf{34.27} & \textbf{25.26} & \textbf{26.78} & 20.00 & \textbf{40.74} & 21.90 & 2.23 & \textbf{97.00} & 22.37 & \textbf{54.52} & \textbf{35.35} \\

\midrule

% =============================================
% Llama-3.1-8B-Instruct
% =============================================
\multirow{16}{*}{\rotatebox[origin=c]{90}{\textbf{Llama-3.1-8B}}} 
 & Full-KV  & 29.39 & 45.17 & 55.74 & 58.31 & 48.12 & 32.57 & 34.53 & 25.30 & 26.91 & 72.50 & 91.78 & 44.32 & 8.47 & 99.50 & 63.43 & 52.59 & 49.29 \\
 \cmidrule(lr){2-19}
 
 % --- SnapKV Variants ---
 & \textit{Metric SnapKV ($\pi_1$)} & & & & & & & & & & & & & & & & & \\
 & \hspace{1em} Uniform-$\pi_1$ & 28.36 & 28.10 & 34.36 & 51.76 & 33.42 & 26.02 & 27.27 & 21.95 & 22.41 & 46.50 & 91.78 & 44.15 & 5.59 & 97.50 & 66.52 & 54.26 & 42.50 \\
 & \hspace{1em} Pyramid-$\pi_1$ & 25.10 & 23.46 & 34.42 & 49.99 & 37.61 & 27.21 & 26.92 & 23.10 & 21.40 & 48.00 & \textbf{91.95} & 44.42 & 6.17 & 98.00 & 64.83 & 55.15 & 42.36 \\
 & \hspace{1em} Ada-$\pi_1$ & 28.27 & 28.43 & 37.38 & 53.24 & 36.21 & 27.66 & 27.46 & 23.09 & 23.06 & 56.50 & 91.76 & \textbf{44.68} & 6.01 & 98.00 & \textbf{66.88} & 55.05 & 43.98 \\
 & \hspace{1em} \textbf{LU-KV-$\pi_1$ (Ours)} & \textbf{29.99} & \textbf{40.10} & \textbf{56.09} & \textbf{56.39} & \textbf{46.94} & \textbf{28.51} & \textbf{32.04} & \textbf{24.32} & \textbf{25.52} & \textbf{65.50} & 89.28 & 43.76 & \textbf{6.43} & \textbf{99.50} & 59.64 & \textbf{59.27} & \textbf{47.70} \\
 
 \addlinespace[4pt] 
 
 % --- KeyDiff Variants ---
 & \textit{Metric KeyDiff ($\pi_2$)} & & & & & & & & & & & & & & & & & \\
 & \hspace{1em} Uniform-$\pi_2$ & 29.06 & 27.88 & 39.53 & 48.54 & 30.91 & 25.13 & 28.43 & 23.57 & 21.45 & 46.50 & 91.53 & 44.29 & \textbf{9.19} & 99.50 & 41.32 & 55.02 & 41.37 \\
 & \hspace{1em} Pyramid-$\pi_2$ & \textbf{30.59} & 30.06 & 39.03 & 49.80 & 35.39 & 26.51 & 27.99 & 23.32 & 21.32 & 41.00 & \textbf{91.83} & 44.30 & 4.95 & 97.50 & 29.25 & 56.26 & 40.57 \\
 & \hspace{1em} Ada-$\pi_2$ & 28.95 & 34.97 & 44.70 & 50.05 & 38.63 & \textbf{28.80} & 30.29 & 23.85 & 22.98 & 60.00 & 90.65 & \textbf{44.90} & 6.99 & 98.00 & \textbf{60.66} & 56.96 & 45.09 \\
 & \hspace{1em} \textbf{LU-KV-$\pi_2$ (Ours)} & 29.57 & \textbf{42.54} & \textbf{53.43} & \textbf{51.84} & \textbf{43.02} & 25.90 & \textbf{33.60} & \textbf{24.79} & \textbf{25.29} & \textbf{67.50} & 90.30 & 44.19 & 6.57 & 99.50 & 58.90 & \textbf{57.61} & \textbf{47.16} \\
 
 \addlinespace[4pt] 
 
 % --- EA Variants ---
 & \textit{Metric EA ($\pi_3$)} & & & & & & & & & & & & & & & & & \\
 & \hspace{1em} Uniform-$\pi_3$ & 30.50 & 37.17 & 41.61 & 48.91 & 38.89 & 25.11 & 29.55 & 23.23 & 24.99 & 47.50 & 89.85 & 42.64 & 9.21 & 91.00 & 55.59 & 53.87 & 43.10 \\
 & \hspace{1em} Pyramid-$\pi_3$ & 29.40 & 35.67 & 40.71 & 48.76 & 37.02 & 24.09 & 28.79 & 23.51 & 24.79 & 42.00 & 90.26 & \textbf{43.11} & \textbf{9.36} & 88.50 & 51.80 & \textbf{58.27} & 42.25 \\
 & \hspace{1em} Ada-$\pi_3$ & 29.93 & 34.50 & 40.53 & 45.49 & 33.99 & 27.80 & 26.62 & 23.42 & 23.62 & 51.00 & \textbf{91.43} & 40.88 & 6.56 & 95.00 & \textbf{63.51} & 58.24 & 43.28 \\
 & \hspace{1em} \textbf{LU-KV-$\pi_3$ (Ours)} & \textbf{32.13} & \textbf{46.81} & \textbf{56.09} & \textbf{53.88} & \textbf{45.64} & \textbf{30.23} & \textbf{34.12} & \textbf{25.07} & \textbf{25.77} & \textbf{71.50} & 90.12 & 42.34 & 6.59 & \textbf{99.50} & 63.38 & 56.45 & \textbf{48.73} \\

\midrule

% =============================================
% Qwen2.5-32B-Instruct
% =============================================
\multirow{16}{*}{\rotatebox[origin=c]{90}{\textbf{Qwen2.5-32B}}} 
 & Full-KV  & 30.68 & 45.93 & 52.13 & 63.00 & 60.75 & 38.71 & 32.43 & 24.51 & 25.06 & 72.00 & 88.71 & 46.01 & 11.50 & 100.00 & 50.72 & 33.98 & 48.51 \\
 \cmidrule(lr){2-19}
 
 % --- SnapKV Variants ---
 & \textit{Metric SnapKV ($\pi_1$)} & & & & & & & & & & & & & & & & & \\
 & \hspace{1em} Uniform-$\pi_1$ & 24.75 & 20.48 & 29.83 & 55.54 & 45.23 & 32.74 & 28.49 & 19.97 & 21.53 & 59.00 & 88.89 & 45.47 & 9.00 & 88.50 & 55.16 & 34.82 & 41.21 \\
 & \hspace{1em} Pyramid-$\pi_1$ & 18.46 & 15.29 & 24.56 & 49.62 & 39.99 & 31.07 & 26.31 & 19.75 & 19.46 & 47.50 & 88.69 & 44.52 & 9.25 & 91.67 & 59.47 & 33.20 & 38.68 \\
 & \hspace{1em} Ada-$\pi_1$ & 25.99 & 21.37 & 29.88 & 54.47 & 45.43 & 30.89 & 28.34 & 19.92 & 21.76 & 62.50 & 88.88 & \textbf{45.71} & 9.50 & 90.75 & 55.38 & 34.55 & 41.58 \\
 & \hspace{1em} \textbf{LU-KV-$\pi_1$ (Ours)} & \textbf{29.41} & \textbf{39.16} & \textbf{50.95} & \textbf{62.82} & \textbf{58.00} & \textbf{39.84} & \textbf{31.34} & \textbf{23.12} & \textbf{24.10} & \textbf{71.00} & \textbf{88.89} & 42.07 & 9.50 & \textbf{100.00} & \textbf{60.21} & \textbf{36.71} & \textbf{47.95} \\
 
 \addlinespace[4pt]
 
 % --- KeyDiff Variants ---
 & \textit{Metric KeyDiff ($\pi_2$)} & & & & & & & & & & & & & & & & & \\
 & \hspace{1em} Uniform-$\pi_2$ & 24.92 & 19.10 & 36.53 & 53.43 & 46.28 & 33.18 & 26.63 & 20.81 & 19.53 & 67.50 & 85.67 & 44.29 & 9.13 & 67.58 & 22.60 & 36.03 & 38.33 \\
 & \hspace{1em} Pyramid-$\pi_2$ & 21.03 & 13.96 & 30.96 & 43.09 & 39.51 & 25.00 & 24.38 & 20.16 & 17.38 & 52.00 & 83.96 & 44.19 & 7.00 & 42.42 & 16.10 & 37.59 & 32.42 \\
 & \hspace{1em} Ada-$\pi_2$ & 26.29 & 26.76 & 43.43 & 55.98 & 48.01 & 35.34 & 28.38 & 21.74 & 20.92 & 66.00 & \textbf{88.58} & \textbf{44.40} & \textbf{11.00} & 86.83 & 42.38 & 31.03 & 42.32 \\
 & \hspace{1em} \textbf{LU-KV-$\pi_2$ (Ours)} & \textbf{31.30} & \textbf{42.88} & \textbf{50.55} & \textbf{61.61} & \textbf{59.67} & \textbf{41.41} & \textbf{31.56} & \textbf{24.01} & \textbf{24.25} & \textbf{74.00} & 88.31 & 41.28 & 7.50 & \textbf{100.00} & \textbf{55.45} & \textbf{38.37} & \textbf{48.26} \\

 \addlinespace[4pt]

 % --- EA Variants ---
 & \textit{Metric EA ($\pi_3$)} & & & & & & & & & & & & & & & & & \\
 & \hspace{1em} Uniform-$\pi_3$ & 24.65 & 32.09 & 36.49 & 54.09 & 47.96 & 32.75 & 31.21 & 22.32 & 23.41 & 70.00 & 74.20 & \textbf{43.98} & \textbf{14.61} & 86.22 & 28.61 & 33.14 & 40.98 \\
 & \hspace{1em} Pyramid-$\pi_3$ & 23.00 & 27.65 & 33.70 & 43.95 & 40.11 & 19.23 & 28.89 & 21.47 & 22.73 & 44.50 & 76.41 & 42.64 & 10.41 & 63.99 & 32.60 & 31.01 & 35.14 \\
 & \hspace{1em} Ada-$\pi_3$ & 30.01 & 29.86 & 45.95 & 60.11 & 54.61 & 40.05 & 31.21 & 22.70 & 24.17 & 72.00 & 82.23 & 43.26 & 8.79 & 88.02 & 46.08 & 35.53 & 44.66 \\
 & \hspace{1em} \textbf{LU-KV-$\pi_3$ (Ours)} & \textbf{30.25} & \textbf{43.34} & \textbf{50.88} & \textbf{64.80} & \textbf{58.72} & \textbf{41.60} & \textbf{33.01} & \textbf{24.74} & \textbf{24.79} & \textbf{77.00} & \textbf{85.85} & 40.32 & 10.12 & \textbf{98.25} & \textbf{57.87} & \textbf{36.06} & \textbf{48.60} \\
\bottomrule
\end{tabular}%
}
\end{table*}

\clearpage

\subsection{Detailed Scores Of RULER}
\label{app:ruler_scores}
In this section, we provide a detailed performance breakdown on the \textbf{RULER} benchmark, evaluating the model across both RULER-16K (Table~\ref{tab:ruler_16k_appendix_0.8}) and RULER-4K (Table~\ref{tab:ruler_4k_appendix_0.8}). These evaluations are conducted at a strict 80\% global compression ratio and include the \textbf{EA} ($\pi_3$) metric to test the robustness of our method in extreme retrieval scenarios.

The results show that traditional baselines, such as \textit{Uniform} and \textit{PyramidKV}, experience significant performance degradation in complex tasks like \textit{multikey} and \textit{variable-tracking} (vt). While \textit{AdaKV} provides some improvement in specific configurations, it remains sensitive to the underlying heuristic metric. This is particularly evident with the \textbf{EA} metric; for example, on \textit{Mistral-7B-v0.3} (RULER-16K), \textit{AdaKV} achieves only 26.28\% average accuracy.

In contrast, our proposed method (\textbf{LU-KV}) consistently achieves superior results across all tasks and metrics. By effectively optimizing the budget allocation, our method significantly boosts retrieval accuracy. Notably, in the aforementioned \textit{Mistral-EA} setting, our method improves the average accuracy to 68.60\%. Similar performance gains are observed for \textit{Llama-3.1-8B} and \textit{Qwen2.5-32B}, confirming the effectiveness of our approach across different model scales and importance metrics.

\begin{table}
\centering
\small 
\setlength{\tabcolsep}{2.5pt} 
\renewcommand{\arraystretch}{1.15}

\caption{Detailed scores of 13 datasets on RULER-16K at an 80\% compression ratio.}
\label{tab:ruler_16k_appendix_0.8}

\resizebox{\textwidth}{!}{%
\begin{tabular}{llcccccccccccccc}
\toprule
% ================= HEADER =================
\multirow{2}{*}{\textbf{Model}} & \multirow{2}{*}{\textbf{Method}} & 
\multicolumn{14}{c}{\textbf{RULER Tasks (16K)}} \\
\cmidrule(lr){3-16}
& & 
\rotatebox{45}{single1} & \rotatebox{45}{single2} & \rotatebox{45}{single3} & 
\rotatebox{45}{multikey1} & \rotatebox{45}{multikey2} & \rotatebox{45}{multikey3} & 
\rotatebox{45}{multivalue} & \rotatebox{45}{multiquery} & 
\rotatebox{45}{vt} & \rotatebox{45}{cwe} & \rotatebox{45}{fwe} & 
\rotatebox{45}{qa-1} & \rotatebox{45}{qa-2} & \rotatebox{45}{\textbf{Avg}} \\
\midrule

% =============================================
% Mistral-7B-Instruct-v0.3
% =============================================
\multirow{15}{*}{\rotatebox[origin=c]{90}{\textbf{Mistral-7B-v0.3}}} 
 & Full-KV  & 94.20 & 96.40 & 99.60 & 97.40 & 95.60 & 76.80 & 89.50 & 88.65 & 96.28 & 82.22 & 87.93 & 71.60 & 50.00 & 86.63 \\
 \cmidrule(lr){2-16}
 
 % --- SnapKV Variants ---
 & \textit{Metric SnapKV ($\pi_1$)} & & & & & & & & & & & & & & \\
 & \hspace{1em} Uniform-$\pi_1$ & 40.40 & 16.20 & 2.40 & 14.20 & 6.20 & 1.00 & 9.65 & 11.00 & 66.92 & 66.96 & 85.53 & 29.80 & 33.60 & 29.53 \\
 & \hspace{1em} Pyramid-$\pi_1$ & 50.00 & 57.00 & 2.40 & 28.00 & 4.80 & 0.20 & 16.15 & 21.55 & 62.32 & 31.94 & 82.20 & 32.00 & 33.00 & 32.43 \\
 & \hspace{1em} Ada-$\pi_1$ & 58.00 & 38.80 & 2.40 & 20.20 & 12.40 & 5.60 & 12.85 & 16.80 & 92.08 & 71.36 & \textbf{86.13} & 33.60 & 37.00 & 37.48 \\
 & \hspace{1em} \textbf{LU-KV-$\pi_1$ (Ours)} & \textbf{70.80} & \textbf{78.80} & \textbf{18.20} & \textbf{83.60} & \textbf{79.20} & \textbf{67.40} & \textbf{67.80} & \textbf{76.25} & \textbf{95.88} & \textbf{78.32} & 84.47 & \textbf{62.00} & \textbf{47.00} & \textbf{69.98} \\
 
 \addlinespace[4pt] 
 
 % --- KeyDiff Variants ---
 & \textit{Metric KeyDiff ($\pi_2$)} & & & & & & & & & & & & & & \\
 & \hspace{1em} Uniform-$\pi_2$ & \textbf{94.60} & 72.80 & 100.00 & 78.80 & 7.40 & 0.80 & 94.80 & 86.10 & 94.16 & 65.56 & \textbf{90.87} & 32.40 & 35.80 & 65.70 \\
 & \hspace{1em} Pyramid-$\pi_2$ & 93.20 & \textbf{96.20} & 99.60 & \textbf{88.20} & 6.60 & 0.60 & 92.00 & 89.75 & \textbf{94.36} & 36.92 & 88.73 & 31.40 & 34.80 & 65.57 \\
 & \hspace{1em} Ada-$\pi_2$ & 92.60 & 91.20 & 97.40 & 87.80 & 6.80 & 1.20 & 88.00 & 86.45 & 91.28 & 75.44 & 86.47 & 36.40 & 36.60 & 67.51 \\
 & \hspace{1em} \textbf{LU-KV-$\pi_2$ (Ours)} & 85.60 & 76.60 & 100.00 & 87.00 & \textbf{90.80} & \textbf{35.20} & \textbf{96.45} & \textbf{92.85} & 92.16 & \textbf{80.78} & 86.80 & \textbf{64.60} & \textbf{46.80} & \textbf{79.66} \\

 \addlinespace[4pt]
 
 % --- EA Variants ---
 & \textit{Metric EA ($\pi_3$)} & & & & & & & & & & & & & & \\
 & \hspace{1em} Uniform-$\pi_3$ & 19.80 & 46.60 & 0.00 & 11.80 & 0.00 & 0.00 & 30.20 & 19.85 & 37.20 & 0.38 & 62.40 & 23.00 & 18.80 & 20.77 \\
 & \hspace{1em} Pyramid-$\pi_3$ & 46.00 & \textbf{51.40} & 0.00 & 20.20 & 0.00 & 0.00 & 28.25 & 17.40 & 34.04 & 29.28 & 48.93 & 32.60 & 29.20 & 25.95 \\
 & \hspace{1em} Ada-$\pi_3$ & 72.00 & 26.20 & 1.80 & 13.20 & 11.00 & 2.40 & 26.70 & 15.70 & 42.12 & 0.34 & \textbf{87.13} & 23.80 & 19.20 & 26.28 \\
 & \hspace{1em} \textbf{LU-KV-$\pi_3$ (Ours)} & \textbf{84.20} & 21.80 & \textbf{58.60} & \textbf{68.80} & \textbf{96.00} & \textbf{55.60} & \textbf{78.45} & \textbf{60.10} & \textbf{93.52} & \textbf{78.96} & 84.73 & \textbf{62.40} & \textbf{48.60} & \textbf{68.60} \\
\midrule

% =============================================
% Llama-3.1-8B-Instruct
% =============================================
\multirow{15}{*}{\rotatebox[origin=c]{90}{\textbf{Llama-3.1-8B}}} 
 & Full-KV  & 100.00 & 100.00 & 100.00 & 99.60 & 100.00 & 99.20 & 98.70 & 99.10 & 99.80 & 88.80 & 89.93 & 81.20 & 57.00 & 93.33 \\
 \cmidrule(lr){2-16}
 
 % --- SnapKV Variants ---
 & \textit{Metric SnapKV ($\pi_1$)} & & & & & & & & & & & & & & \\
 & \hspace{1em} Uniform-$\pi_1$ & 98.00 & 83.60 & 2.60 & 52.20 & 11.20 & 4.40 & 34.90 & 40.50 & 89.40 & 14.22 & 79.00 & 28.80 & 30.80 & 43.82 \\
 & \hspace{1em} Pyramid-$\pi_1$ & 90.60 & 97.80 & 2.40 & 85.40 & 20.60 & 0.80 & 72.70 & 76.85 & 84.64 & 11.10 & 80.33 & 29.40 & 33.00 & 52.74 \\
 & \hspace{1em} Ada-$\pi_1$ & 99.20 & 90.20 & 3.00 & 69.20 & 20.60 & 19.20 & 47.30 & 55.20 & 95.92 & 42.54 & \textbf{86.47} & 32.40 & 33.20 & 53.42 \\
 & \hspace{1em} \textbf{LU-KV-$\pi_1$ (Ours)} & \textbf{100.00} & \textbf{99.80} & \textbf{54.00} & \textbf{99.40} & \textbf{85.20} & \textbf{93.60} & \textbf{98.60} & \textbf{98.80} & \textbf{97.44} & \textbf{61.40} & 84.67 & \textbf{65.40} & \textbf{49.80} & \textbf{83.70} \\
 
 \addlinespace[4pt]
 
 % --- KeyDiff Variants ---
 & \textit{Metric KeyDiff ($\pi_2$)} & & & & & & & & & & & & & & \\
 & \hspace{1em} Uniform-$\pi_2$ & 100.00 & 100.00 & 100.00 & 99.60 & 16.40 & 0.00 & \textbf{99.25} & 99.55 & 98.28 & 59.70 & 86.93 & 38.40 & 43.00 & 72.39 \\
 & \hspace{1em} Pyramid-$\pi_2$ & 100.00 & 100.00 & 100.00 & 99.60 & 11.00 & 0.00 & 98.90 & \textbf{99.60} & \textbf{99.44} & 23.14 & 85.93 & 39.80 & 40.60 & 69.08 \\
 & \hspace{1em} Ada-$\pi_2$ & 100.00 & 100.00 & 100.00 & 99.60 & 26.00 & 0.80 & 98.55 & 99.25 & 98.60 & 78.34 & \textbf{90.93} & 45.60 & 42.00 & 75.36 \\
 & \hspace{1em} \textbf{LU-KV-$\pi_2$ (Ours)} & 100.00 & 100.00 & 100.00 & 99.20 & \textbf{99.20} & \textbf{56.00} & 99.20 & 99.20 & 97.36 & \textbf{80.80} & 87.53 & \textbf{76.20} & \textbf{52.60} & \textbf{88.25} \\

 \addlinespace[4pt]
 
 % --- EA Variants ---
 & \textit{Metric EA ($\pi_3$)} & & & & & & & & & & & & & & \\
 & \hspace{1em} Uniform-$\pi_3$ & 98.60 & 96.20 & 2.40 & 85.80 & 5.20 & 0.00 & 79.45 & 89.75 & 81.00 & 17.44 & 61.33 & 51.60 & 40.20 & 54.54 \\
 & \hspace{1em} Pyramid-$\pi_3$ & 98.80 & 88.80 & 1.20 & 84.80 & 10.00 & 0.00 & 76.65 & 80.65 & 86.80 & 5.88 & 36.20 & 52.00 & 41.00 & 50.98 \\
 & \hspace{1em} Ada-$\pi_3$ & 99.80 & 99.40 & 3.80 & 90.60 & 38.80 & 2.60 & 75.55 & 91.60 & 95.28 & 9.56 & 81.40 & 43.20 & 40.80 & 59.41 \\
 & \hspace{1em} \textbf{LU-KV-$\pi_3$ (Ours)} & \textbf{100.00} & \textbf{100.00} & \textbf{99.60} & \textbf{99.40} & \textbf{99.80} & \textbf{98.60} & \textbf{97.55} & \textbf{98.80} & \textbf{98.04} & \textbf{80.18} & \textbf{88.13} & \textbf{78.40} & \textbf{54.00} & \textbf{91.73} \\
\midrule

% =============================================
% Qwen2.5-32B-Instruct
% =============================================
\multirow{15}{*}{\rotatebox[origin=c]{90}{\textbf{Qwen2.5-32B}}} 
 & Full-KV  & 100.00 & 100.00 & 100.00 & 100.00 & 99.80 & 100.00 & 99.85 & 99.95 & 100.00 & 97.70 & 96.20 & 79.40 & 62.40 & 95.02 \\
 \cmidrule(lr){2-16}
 
 % --- SnapKV Variants ---
 & \textit{Metric SnapKV ($\pi_1$)} & & & & & & & & & & & & & & \\
 & \hspace{1em} Uniform-$\pi_1$ & 97.40 & 55.60 & 3.80 & 25.80 & 4.80 & 2.00 & 14.40 & 19.60 & 99.28 & 87.14 & 94.00 & 28.00 & 39.00 & 43.91 \\
 & \hspace{1em} Pyramid-$\pi_1$ & 83.80 & 36.00 & 2.40 & 19.20 & 2.00 & 0.00 & 13.15 & 14.95 & 93.68 & 56.84 & \textbf{95.73} & 26.40 & 34.60 & 36.83 \\
 & \hspace{1em} Ada-$\pi_1$ & 98.80 & 52.60 & 4.40 & 21.80 & 7.00 & 4.20 & 14.75 & 18.25 & 99.32 & 88.48 & 94.53 & 29.40 & 39.00 & 44.04 \\
 & \hspace{1em} \textbf{LU-KV-$\pi_1$ (Ours)} & \textbf{99.80} & \textbf{99.20} & \textbf{32.00} & \textbf{84.20} & \textbf{71.80} & \textbf{78.40} & \textbf{84.60} & \textbf{85.80} & \textbf{99.72} & \textbf{95.66} & 93.13 & \textbf{65.00} & \textbf{56.80} & \textbf{80.47} \\
 
 \addlinespace[4pt]
 
 % --- KeyDiff Variants ---
 & \textit{Metric KeyDiff ($\pi_2$)} & & & & & & & & & & & & & & \\
 & \hspace{1em} Uniform-$\pi_2$ & 100.00 & 100.00 & 100.00 & 100.00 & 8.00 & 1.00 & 99.40 & 99.95 & 98.92 & 90.36 & \textbf{99.33} & 36.40 & 41.40 & 74.98 \\
 & \hspace{1em} Pyramid-$\pi_2$ & 100.00 & 100.00 & 99.80 & 99.60 & 1.00 & 0.20 & \textbf{99.55} & 99.95 & 84.52 & 69.26 & 98.93 & 30.20 & 35.80 & 70.68 \\
 & \hspace{1em} Ada-$\pi_2$ & 100.00 & 100.00 & 100.00 & 99.80 & 44.20 & 23.00 & 99.00 & 99.90 & 99.88 & 95.34 & 99.07 & 44.00 & 46.40 & 80.81 \\
 & \hspace{1em} \textbf{LU-KV-$\pi_2$ (Ours)} & 100.00 & 100.00 & 100.00 & 100.00 & \textbf{86.60} & \textbf{46.80} & 99.25 & 99.95 & \textbf{100.00} & \textbf{96.02} & 96.20 & \textbf{71.60} & \textbf{59.00} & \textbf{88.88} \\

 \addlinespace[4pt]

 % --- EA Variants ---
 & \textit{Metric EA ($\pi_3$)} & & & & & & & & & & & & & & \\
  & \hspace{1em} Uniform-$\pi_3$ & 93.40 & 97.60 & 6.00 & 97.40 & 0.40 & 0.00 & 94.75 & 98.75 & 99.44 & 73.10 & 80.00 & 49.20 & 48.20 & 64.48 \\
 & \hspace{1em} Pyramid-$\pi_3$ & 81.80 & 69.00 & 0.20 & 55.20 & 0.00 & 0.00 & 56.30 & 61.20 & 96.12 & 10.94 & 39.93 & 39.20 & 41.80 & 42.44 \\
 & \hspace{1em} Ada-$\pi_3$ & \textbf{100.00} & \textbf{100.00} & 22.00 & 99.20 & 96.40 & \textbf{39.40} & 98.40 & 99.55 & 99.88 & 92.04 & \textbf{93.93} & 60.20 & 50.80 & 80.91 \\
 & \hspace{1em} \textbf{LU-KV-$\pi_3$ (Ours)} & 99.60 & 99.80 & \textbf{91.00} & \textbf{100.00} & \textbf{99.40} & 29.40 & \textbf{99.35} & \textbf{99.95} & \textbf{100.00} & \textbf{95.98} & 93.00 & \textbf{79.40} & \textbf{60.60} & \textbf{88.27} \\
\bottomrule
\end{tabular}%
}
\end{table}

\clearpage

\begin{table*}[t]
\centering
\small 
\setlength{\tabcolsep}{2.5pt} 
\renewcommand{\arraystretch}{1.15}

\caption{Detailed scores of 13 datasets on RULER-4K at an 80\% compression ratio.}
\label{tab:ruler_4k_appendix_0.8}

\resizebox{\textwidth}{!}{%
\begin{tabular}{llcccccccccccccc}
\toprule
% ================= HEADER =================
\multirow{2}{*}{\textbf{Model}} & \multirow{2}{*}{\textbf{Method}} & 
\multicolumn{14}{c}{\textbf{RULER Tasks (4K)}} \\
\cmidrule(lr){3-16}
& & 
\rotatebox{45}{single1} & \rotatebox{45}{single2} & \rotatebox{45}{single3} & 
\rotatebox{45}{multikey1} & \rotatebox{45}{multikey2} & \rotatebox{45}{multikey3} & 
\rotatebox{45}{multivalue} & \rotatebox{45}{multiquery} & 
\rotatebox{45}{vt} & \rotatebox{45}{cwe} & \rotatebox{45}{fwe} & 
\rotatebox{45}{qa-1} & \rotatebox{45}{qa-2} & \rotatebox{45}{\textbf{Avg}} \\
\midrule

% =============================================
% Mistral-7B-Instruct-v0.3
% =============================================
\multirow{16}{*}{\rotatebox[origin=c]{90}{\textbf{Mistral-7B-v0.3}}} 
 & Full-KV  & 93.20 & 95.80 & 100.00 & 99.40 & 100.00 & 97.40 & 89.05 & 97.50 & 99.56 & 98.46 & 95.60 & 76.80 & 54.60 & 92.11 \\
 \cmidrule(lr){2-16}
 
 % --- SnapKV Variants ---
 & \textit{Metric SnapKV ($\pi_1$)} & & & & & & & & & & & & & & \\
 & \hspace{1em} Uniform-$\pi_1$ & 37.00 & 13.00 & 2.40 & 18.40 & 5.80 & 0.60 & 11.80 & 17.10 & 25.16 & 84.66 & 82.40 & 40.60 & 35.60 & 28.81 \\
 & \hspace{1em} Pyramid-$\pi_1$ & 35.20 & 16.40 & 2.40 & 19.00 & 4.00 & 0.00 & 11.40 & 17.10 & 28.80 & 42.84 & 75.40 & 35.20 & 33.80 & 24.73 \\
 & \hspace{1em} Ada-$\pi_1$ & 50.40 & 13.00 & 2.40 & 18.20 & 12.80 & 5.80 & 12.70 & 17.25 & 49.84 & 89.54 & 87.00 & 44.60 & 40.20 & 34.13 \\
 & \hspace{1em} \textbf{LU-KV-$\pi_1$ (Ours)} & \textbf{79.00} & \textbf{83.60} & \textbf{12.60} & \textbf{71.20} & \textbf{89.80} & \textbf{93.40} & \textbf{54.50} & \textbf{66.80} & \textbf{93.64} & \textbf{94.80} & \textbf{93.07} & \textbf{68.40} & \textbf{49.60} & \textbf{73.11} \\
 
 \addlinespace[4pt] 
 
 % --- KeyDiff Variants ---
 & \textit{Metric KeyDiff ($\pi_2$)} & & & & & & & & & & & & & & \\
 & \hspace{1em} Uniform-$\pi_2$ & 90.60 & 97.00 & 100.00 & 83.20 & 8.00 & 0.00 & 88.10 & 89.40 & 98.92 & 60.48 & 81.33 & 39.60 & 28.80 & 66.57 \\
 & \hspace{1em} Pyramid-$\pi_2$ & \textbf{94.40} & \textbf{97.20} & 100.00 & 84.80 & 6.00 & 0.00 & 83.80 & 89.30 & 98.84 & 14.72 & 77.53 & 37.60 & 23.80 & 62.15 \\
 & \hspace{1em} Ada-$\pi_2$ & 91.80 & 95.00 & 99.40 & 83.80 & 5.80 & 0.00 & 82.00 & 86.80 & 99.36 & 74.22 & 75.93 & 45.80 & 32.00 & 67.07 \\
 & \hspace{1em} \textbf{LU-KV-$\pi_2$ (Ours)} & 87.40 & 96.00 & 99.40 & \textbf{98.40} & \textbf{98.80} & \textbf{50.20} & \textbf{96.35} & \textbf{97.40} & 99.36 & \textbf{92.28} & \textbf{92.93} & \textbf{72.60} & \textbf{49.20} & \textbf{86.95} \\

 \addlinespace[4pt]
 
 % --- EA Variants ---
 & \textit{Metric EA ($\pi_3$)} & & & & & & & & & & & & & & \\
 & \hspace{1em} Uniform-$\pi_3$ & 66.40 & 57.40 & 0.40 & 44.00 & 0.60 & 0.00 & 56.30 & 33.95 & 57.92 & 71.06 & 79.87 & 49.00 & 41.40 & 42.95 \\
 & \hspace{1em} Pyramid-$\pi_3$ & 72.00 & 51.60 & 0.00 & 38.00 & 0.60 & 0.00 & 39.75 & 28.15 & 80.44 & 57.70 & 64.07 & 46.20 & 38.80 & 39.79 \\
 & \hspace{1em} Ada-$\pi_3$ & \textbf{86.00} & 39.20 & 3.00 & 37.60 & 34.00 & 41.00 & 59.20 & 34.40 & 97.84 & 90.94 & 92.47 & 51.60 & 40.80 & 54.47 \\
 & \hspace{1em} \textbf{LU-KV-$\pi_3$ (Ours)} & 83.80 & \textbf{83.20} & \textbf{98.60} & \textbf{97.20} & \textbf{99.80} & \textbf{97.00} & \textbf{82.05} & \textbf{90.15} & \textbf{99.52} & \textbf{96.50} & \textbf{95.60} & \textbf{75.60} & \textbf{55.60} & \textbf{88.82} \\
\midrule

% =============================================
% Llama-3.1-8B-Instruct
% =============================================
\multirow{16}{*}{\rotatebox[origin=c]{90}{\textbf{Llama-3.1-8B}}} 
 & Full-KV  & 100.00 & 100.00 & 100.00 & 99.80 & 100.00 & 99.80 & 99.90 & 99.90 & 99.88 & 99.68 & 94.93 & 88.00 & 62.60 & 95.73 \\
 \cmidrule(lr){2-16}
 
 % --- SnapKV Variants ---
 & \textit{Metric SnapKV ($\pi_1$)} & & & & & & & & & & & & & & \\
 & \hspace{1em} Uniform-$\pi_1$ & 82.00 & 70.80 & 2.40 & 39.80 & 13.20 & 2.60 & 38.35 & 35.65 & 61.16 & 62.06 & 81.47 & 42.80 & 32.80 & 43.47 \\
 & \hspace{1em} Pyramid-$\pi_1$ & 74.80 & 98.20 & 2.40 & 77.00 & 23.00 & 0.00 & 69.70 & 69.80 & 48.48 & 37.42 & 71.07 & 39.00 & 29.60 & 49.27 \\
 & \hspace{1em} Ada-$\pi_1$ & 86.60 & 69.80 & 2.40 & 44.40 & 17.60 & 13.80 & 37.75 & 38.80 & 81.20 & 89.82 & 87.33 & 48.20 & 36.80 & 50.35 \\
 & \hspace{1em} \textbf{LU-KV-$\pi_1$ (Ours)} & \textbf{99.40} & \textbf{99.60} & \textbf{14.20} & \textbf{99.80} & \textbf{87.20} & \textbf{95.60} & \textbf{98.30} & \textbf{99.80} & \textbf{97.52} & \textbf{96.08} & \textbf{93.53} & \textbf{77.60} & \textbf{55.40} & \textbf{85.69} \\
 
 \addlinespace[4pt]
 
 % --- KeyDiff Variants ---
 & \textit{Metric KeyDiff ($\pi_2$)} & & & & & & & & & & & & & & \\
 & \hspace{1em} Uniform-$\pi_2$ & 100.00 & 100.00 & 100.00 & 99.80 & 14.60 & 0.00 & 98.65 & 99.85 & 98.08 & 39.42 & 82.53 & 38.60 & 27.00 & 69.12 \\
 & \hspace{1em} Pyramid-$\pi_2$ & 100.00 & 99.80 & 100.00 & \textbf{100.00} & 7.80 & 0.00 & 99.25 & 99.80 & \textbf{99.64} & 20.70 & 78.13 & 41.20 & 27.60 & 67.22 \\
 & \hspace{1em} Ada-$\pi_2$ & 100.00 & 99.80 & 100.00 & 99.60 & 35.00 & 0.20 & 99.40 & 99.65 & 99.08 & 68.36 & 82.20 & 54.20 & 36.00 & 74.88 \\
 & \hspace{1em} \textbf{LU-KV-$\pi_2$ (Ours)} & 100.00 & 100.00 & 100.00 & 99.80 & \textbf{99.60} & \textbf{56.80} & \textbf{99.80} & \textbf{99.90} & 99.24 & \textbf{92.50} & \textbf{93.80} & \textbf{83.20} & \textbf{57.20} & \textbf{90.91} \\

 \addlinespace[4pt]
 
 % --- EA Variants ---
 & \textit{Metric EA ($\pi_3$)} & & & & & & & & & & & & & & \\
 & \hspace{1em} Uniform-$\pi_3$ & 98.60 & 91.40 & 1.60 & 92.00 & 13.20 & 0.00 & 90.20 & 93.35 & 81.92 & 43.32 & 73.00 & 55.60 & 45.40 & 59.97 \\
 & \hspace{1em} Pyramid-$\pi_3$ & 99.40 & 81.60 & 1.00 & 89.80 & 22.60 & 0.00 & 86.75 & 92.00 & 97.12 & 26.64 & 58.27 & 57.00 & 44.60 & 58.21 \\
 & \hspace{1em} Ada-$\pi_3$ & 99.80 & 95.00 & 5.00 & 89.20 & 55.40 & 4.60 & 78.65 & 88.65 & 92.92 & 88.94 & 90.40 & 42.60 & 44.80 & 67.38 \\
 & \hspace{1em} \textbf{LU-KV-$\pi_3$ (Ours)} & \textbf{100.00} & \textbf{99.80} & \textbf{100.00} & \textbf{100.00} & \textbf{99.80} & \textbf{99.40} & \textbf{99.90} & \textbf{99.90} & \textbf{98.60} & \textbf{97.46} & \textbf{95.07} & \textbf{83.60} & \textbf{59.60} & \textbf{94.86} \\
\midrule

% =============================================
% Qwen2.5-32B-Instruct
% =============================================
\multirow{16}{*}{\rotatebox[origin=c]{90}{\textbf{Qwen2.5-32B}}} 
 & Full-KV  & 100.00 & 100.00 & 100.00 & 99.80 & 100.00 & 100.00 & 99.95 & 100.00 & 100.00 & 99.86 & 98.60 & 89.40 & 67.60 & 96.55 \\
 \cmidrule(lr){2-16}
 
 % --- SnapKV Variants ---
 & \textit{Metric SnapKV ($\pi_1$)} & & & & & & & & & & & & & & \\
 & \hspace{1em} Uniform-$\pi_1$ & \textbf{95.40} & 39.00 & 3.60 & 27.20 & 4.80 & 0.40 & 25.50 & 22.75 & 92.64 & 96.18 & 86.27 & 56.20 & 41.80 & 45.52 \\
 & \hspace{1em} Pyramid-$\pi_1$ & 80.20 & 13.00 & 2.40 & 13.40 & 1.60 & 0.00 & 13.00 & 12.70 & 57.60 & 63.30 & 66.67 & 34.00 & 31.00 & 29.91 \\
 & \hspace{1em} Ada-$\pi_1$ & 93.60 & 24.00 & 2.40 & 19.60 & 7.00 & 4.00 & 17.55 & 17.40 & 94.80 & 98.50 & 89.00 & 59.00 & 42.60 & 43.80 \\
 & \hspace{1em} \textbf{LU-KV-$\pi_1$ (Ours)} & 93.00 & \textbf{92.20} & \textbf{16.40} & \textbf{81.40} & \textbf{80.60} & \textbf{84.20} & \textbf{91.00} & \textbf{87.90} & \textbf{99.08} & \textbf{99.68} & \textbf{95.40} & \textbf{85.00} & \textbf{63.80} & \textbf{82.28} \\
 
 \addlinespace[4pt]
 
 % --- KeyDiff Variants ---
 & \textit{Metric KeyDiff ($\pi_2$)} & & & & & & & & & & & & & & \\
 & \hspace{1em} Uniform-$\pi_2$ & 100.00 & 100.00 & 100.00 & 99.60 & 4.60 & 0.40 & 99.95 & 100.00 & 76.00 & 78.94 & 88.47 & 49.20 & 34.20 & 71.64 \\
 & \hspace{1em} Pyramid-$\pi_2$ & 100.00 & 100.00 & 99.80 & 99.40 & 1.00 & 0.00 & 99.95 & 99.75 & 26.56 & 8.10 & 75.00 & 35.40 & 24.40 & 59.18 \\
 & \hspace{1em} Ada-$\pi_2$ & 100.00 & 100.00 & 100.00 & 99.80 & 37.60 & 7.60 & 99.95 & 100.00 & 99.68 & 94.72 & 87.53 & 59.40 & 40.60 & 78.99 \\
 & \hspace{1em} \textbf{LU-KV-$\pi_2$ (Ours)} & 100.00 & 100.00 & 100.00 & 99.80 & \textbf{96.20} & \textbf{39.80} & 99.95 & 100.00 & \textbf{100.00} & \textbf{99.20} & \textbf{95.53} & \textbf{87.80} & \textbf{64.20} & \textbf{90.96} \\

 \addlinespace[4pt]

 % --- EA Variants ---
 & \textit{Metric EA ($\pi_3$)} & & & & & & & & & & & & & & \\
& \hspace{1em} Uniform-$\pi_3$ & 86.80 & 92.80 & 3.00 & 91.80 & 0.80 & 0.00 & 91.70 & 91.85 & 99.60 & 90.28 & 82.13 & 62.20 & 52.20 & 65.01 \\
 & \hspace{1em} Pyramid-$\pi_3$ & 71.20 & 49.00 & 0.00 & 33.00 & 0.00 & 0.00 & 30.90 & 35.20 & 92.68 & 27.58 & 38.93 & 49.80 & 41.20 & 36.11 \\
 & \hspace{1em} Ada-$\pi_3$ & \textbf{100.00} & 99.80 & 26.60 & 98.20 & 95.40 & \textbf{69.80} & 99.65 & 99.55 & 99.88 & 99.68 & 94.00 & 73.00 & 59.60 & 85.78 \\
 & \hspace{1em} \textbf{LU-KV-$\pi_3$ (Ours)} & 99.80 & \textbf{100.00} & \textbf{97.20} & \textbf{99.60} & \textbf{100.00} & 51.00 & \textbf{99.90} & \textbf{100.00} & \textbf{100.00} & \textbf{99.86} & \textbf{96.60} & \textbf{86.00} & \textbf{64.60} & \textbf{91.89} \\
\bottomrule
\end{tabular}%
}
\end{table*}

\clearpage

\section{Details about Benchmarks} ~\label{app:benchmark}
To evaluate the long-context capabilities of the models comprehensively, we employ two distinct benchmarks: RULER and LongBench. These benchmarks provide complementary insights, with RULER offering controllable synthetic stress tests and LongBench providing realistic multi-task evaluations.

\subsection{RULER}
RULER~\citep{hsieh_ruler_2024} is a synthetic benchmark designed to evaluate long-context language models beyond the standard retrieval-based ``needle-in-a-haystack'' (NIAH) tests. Unlike simple retrieval tasks, RULER introduces flexible configurations to customize sequence length and task complexity. It categorizes tasks into four distinct domains to test behaviors beyond searching from context:
\begin{itemize}
    \item \textbf{Retrieval:} Extending the vanilla NIAH, this category includes Single NIAH (S-NIAH), Multi-keys NIAH (MK-NIAH), Multi-values NIAH (MV-NIAH), and Multi-queries NIAH (MQ-NIAH). These tasks test the model's robustness against distractors and its ability to retrieve diverse types and quantities of needles.
    \item \textbf{Multi-hop Tracing:} To evaluate coreference chain resolution, RULER utilizes a Variable Tracking (VT) task, requiring the model to trace variable assignment chains across the long context.
    \item \textbf{Aggregation:} This category tests the ability to aggregate relevant information spanning long-range context. Tasks include Common Words Extraction (CWE) and Frequent Words Extraction (FWE), where the model identifies words based on their frequency distribution.
    \item \textbf{Question Answering (QA):} This domain uses augmented versions of SQuAD~\citep{rajpurkar_squad_2018} and HotpotQA~\citep{yang_hotpotqa_2018} with inserted distractors to simulate long-context question answering scenarios.
\end{itemize}

\subsection{LongBench}
Complementing the synthetic nature of RULER, we utilize LongBench~\citep{bai_longbench_2024}, a multi-task benchmark designed to assess long-context understanding in realistic scenarios. In this work, we focus specifically on the 16 English datasets from LongBench, which cover six major task categories. The English subset comprises:
\begin{itemize}
    \item \textbf{Single-Document QA:} Evaluated using NarrativeQA~\citep{kovcisky_narrativeqa_2018}, Qasper~\citep{dasigi_qasper_2021}, and MultiFieldQA-en, requiring models to comprehend long individual documents.
    \item \textbf{Multi-Document QA:} Involves complex reasoning across multiple documents, utilizing HotpotQA~\citep{yang_hotpotqa_2018}, 2WikiMultihopQA~\citep{ho-etal-2wikiqa-2020}, and MuSiQue~\citep{trivedi_2022_musique}.
    \item \textbf{Summarization:} Tests the ability to synthesize long inputs using GovReport~\citep{huang-govreport_2021}, QMSum~\citep{zhong_qmsum_2021}, and MultiNews~\citep{fabbri-multinews-2019}.
    \item \textbf{Few-Shot Learning:} Assesses in-context learning abilities with long-context examples from TREC~\citep{li-roth-trec-2002}, TriviaQA~\citep{joshi-triviaqa-2017}, and SAMSum~\citep{gliwa-etal-2019-samsum}.
    \item \textbf{Synthetic Tasks:} Includes PassageCount and PassageRetrieval-en to isolate specific long-range dependency capabilities.
    \item \textbf{Code Completion:} Evaluates programming context understanding using LCC~\citep{chen_lcc_2021} and RepoBench-P~\citep{liu2023repobench}.
\end{itemize}
\clearpage

\section{Details about Baselines} ~\label{app:baseline}
We adopted the original hyperparameters for the baseline methods as reported in their respective papers, details in Table~\ref{tab:hyperparameters}. 
% SnapKV~\cite{yuhong_SnapKV_2024} employs a max-pooling kernel size of $7$ and observation window of $32$; AdaKV~\cite{feng_adakv_2024} sets $\alpha_\text{safeguard} = 0.20$ and window size $32$; PyramidKV~\cite{cai_pyramidkv_2024} uses window size $8$ and $\beta = 20$; Expected Attention~\cite{devoto_EA_2025} configures $n_\text{future\_positions} = 512$, $n_\text{sink} = 4$, and $\epsilon = 0.02$.

\begin{table}[htbp]
    \centering
    \caption{Hyperparameter configurations for baseline methods, following their original reports.}
    \label{tab:hyperparameters}
    \small
    \begin{tabular}{ll}
        \toprule
        \textbf{Method} & \textbf{Hyperparameters} \\
        \midrule
        SnapKV~\cite{yuhong_SnapKV_2024} & Kernel size = 7, Window size = 32 \\
        AdaKV~\cite{feng_adakv_2024} & $\alpha_{\text{safeguard}} = 0.20$, Window size = 32 \\
        PyramidKV~\cite{cai_pyramidkv_2024} & $\beta = 20$, Window size = 8 \\
        Expected Attention~\cite{devoto_EA_2025} & $n_{\text{future\_positions}} = 512, n_{\text{sink}} = 4, \epsilon = 0.02$ \\
        \bottomrule
    \end{tabular}
\end{table}

\section{LU-KV Implementation Details} \label{app:impl_details}
For offline calibration, we employed an AI-generated novel ($\approx$ 4,000 words) paired with 30 generated questions. We utilized the $L_2$-norm of the projected Value vectors ($|vW_O|_2$) for scoring, applying intra-layer normalization to the results. We configured the attention sink size to 4. The recent token window was set to 1 for KeyDiff and maintained at 32 for SnapKV. Additionally, we imposed a maximum compression threshold of 99\% for any attention head, ensuring a minimum retention rate of 1\%.

\section{Additional Analyses}

% \camblue{We further analyze whether LU-KV is tied to a specific proxy metric or profiling domain. LU-KV only requires an offline utility curve for the chosen metric, so the same allocation solver can be applied to attention-based metrics such as SnapKV, geometry-based metrics such as KeyDiff, and other metric families once their profiled utility curves are available.}

% \camblue{KVZip studies query-agnostic KV cache compression through context reconstruction, and CAKE uses a cascaded strategy to exploit layer-wise cache redundancy. In the extended experiments, LU-KV is evaluated with these methods by applying its allocation step to their corresponding proxy metrics. HeadKV also uses offline calibration, but it learns a static head-importance profile from retrieval and reasoning tasks; LU-KV instead learns metric-specific marginal utility curves and solves a global budget allocation problem for each target compression ratio.}

% \camblue{The offline profiling corpus is intentionally separate from LongBench and RULER. This design makes the reported downstream evaluation a transfer test from the calibration distribution to English long-context tasks, rather than a benchmark-specific tuning procedure.}
This section provides additional analyses on compression-ratio sensitivity, offline-profiling transferability, integration with other KV compression metrics, and comparison with HeadKV. Unless otherwise specified, all additional LongBench results in this section are reported on Llama-3.1-8B-Instruct, with Full Cache average performance of 49.29 under the same evaluation protocol as the main experiments.

\subsection{Compression-Ratio Sensitivity}
Table~\ref{tab:appendix_compression_sweep} evaluates LU-KV under 50\%, 80\%, and 90\% compression ratios on LongBench. LU-KV remains stable across these budgets and degrades more gracefully than PyramidKV and AdaKV under aggressive compression.

\begin{table*}[h]
\centering
\caption{Performance across compression ratios on LongBench. Full cache performance is 49.29.}
\label{tab:appendix_compression_sweep}
% \begin{camblueblock}
\begin{tabular}{lccc}
\toprule
Method & $\sigma$=50\% &$\sigma$=80\%  &$\sigma$=90\% \\
\midrule
PyramidKV (KeyDiff) & 45.71 & 40.57 & 36.56 \\
AdaKV (KeyDiff) & 48.55 & 45.09 & 38.21 \\
\textbf{LU-KV (KeyDiff)} & 48.47 & 47.16 & 44.29 \\
PyramidKV (SnapKV) & 46.65 & 42.36 & 38.02 \\
AdaKV (SnapKV) & 47.91 & 43.98 & 40.20 \\
\textbf{LU-KV (SnapKV)} & \textbf{49.38} & \textbf{47.70} & \textbf{44.69} \\
\bottomrule
\end{tabular}
% \end{camblueblock}
\end{table*}

\subsection{Transferability Analysis of offline Profiling Data} \label{append:Trans_ana_of_offline_profile_data}
able~\ref{tab:appendix_profile_transfer} evaluates whether the offline profiling data transfers across different calibration corpora. The two profiling examples used in this analysis are listed below:

\begin{itemize}[leftmargin=*, nosep]
    \item \textbf{Chinese Novel (4k):} an AI-generated Chinese novel excerpt of approximately 4K tokens, paired with generated questions that query different information segments in the long context.
    \item \textbf{English SAT (150):} a much shorter English SAT Reading excerpt, e.g., a passage beginning with ``The following text is from Edith Wharton's 1905 novel...'', paired with SAT-style comprehension questions.
\end{itemize}

Both profiling examples are out-of-domain relative to LongBench. The comparison between the Chinese Novel profile and the English SAT profile indicates that LU-KV is not tightly coupled to a single profiling corpus, while still benefiting from calibration data that captures long-context retrieval behavior.

\begin{table*}[h]
\centering
\caption{Transferability analysis of offline profiling data on LongBench.}
\label{tab:appendix_profile_transfer}

% \begin{camblueblock}
\setlength{\tabcolsep}{3pt}
% \begin{tabularx}{\textwidth}{@{}lllYYYYYYY@{}}
% \toprule
% Method & \makecell{Compression \\ Ratio$(\sigma)$} & Profiling Data & Avg. ($\Delta$) &\makecell{Single-\\Doc QA} &\makecell{Multi-\\Doc QA} & Summ. & Few-shot & Synthetic & Code \\
% \midrule
% Full Cache & 0 & - & 49.29 & 43.43 & 46.33 & 28.91 & 69.53 & 53.99 & 58.01 \\
% \midrule
% KeyDiff & 0.5 & - & 47.24 ($\downarrow$4.16\%) & 41.37 & 43.14 & 27.49 & 67.37 & \textbf{53.67} & 55.24 \\
% LU-KV & 0.5 & Chinese Novel (4k) & 48.47 ($\downarrow$1.67\%) & 42.72 & 42.23 & 28.90 & \textbf{70.14} & 52.65 & 59.14 \\
% LU-KV & 0.5 & English SAT (150) & \textbf{49.06} ($\downarrow$0.47\%) & \textbf{43.28} & \textbf{43.19} & \textbf{29.03} & 69.87 & 53.53 & \textbf{60.92} \\
% \midrule
% KeyDiff & 0.8 & - & 41.37 ($\downarrow$16.0\%) & 32.16 & 34.86 & 24.48 & 60.77 & \textbf{54.35} & 48.17 \\
% LU-KV & 0.8 & Chinese Novel (4k) & \textbf{47.16} ($\downarrow$4.32\%) & \textbf{41.85} & 40.25 & \textbf{27.89} & \textbf{67.33} & 53.04 & \textbf{58.26} \\
% LU-KV & 0.8 & English SAT (150) & 46.67 ($\downarrow$5.31\%) & 40.66 & \textbf{40.80} & 27.26 & 66.77 & 53.10 & 57.06 \\
% \bottomrule
% \end{tabularx}
\begin{tabularx}{\textwidth}{@{}lllYYYYYYY@{}}
\toprule
Method & \makecell{Compression \\ Ratio$(\sigma)$} & Profiling Data & Avg. & \makecell{Single-\\Doc QA} & \makecell{Multi-\\Doc QA} & Summ. & Few-shot & Synthetic & Code \\
\midrule
Full Cache & 0 & - & 49.29 & 43.43 & 46.33 & 28.91 & 69.53 & 53.99 & 58.01 \\
\midrule
KeyDiff & 0.5 & - & 47.24 & 41.37 & 43.14 & 27.49 & 67.37 & \textbf{53.67} & 55.24 \\
LU-KV & 0.5 & Chinese Novel (4k) & 48.47 & 42.72 & 42.23 & 28.90 & \textbf{70.14} & 52.65 & 59.14 \\
LU-KV & 0.5 & English SAT (150) & \textbf{49.06} & \textbf{43.28} & \textbf{43.19} & \textbf{29.03} & 69.87 & 53.53 & \textbf{60.92} \\
\midrule
KeyDiff & 0.8 & - & 41.37 & 32.16 & 34.86 & 24.48 & 60.77 & \textbf{54.35} & 48.17 \\
LU-KV & 0.8 & Chinese Novel (4k) & \textbf{47.16} & \textbf{41.85} & 40.25 & \textbf{27.89} & \textbf{67.33} & 53.04 & \textbf{58.26} \\
LU-KV & 0.8 & English SAT (150) & 46.67 & 40.66 & \textbf{40.80} & 27.26 & 66.77 & 53.10 & 57.06 \\
\bottomrule
\end{tabularx}
% \end{camblueblock}
\end{table*}

\subsection{Comparison with HeadKV}
We further compare LU-KV with HeadKV, an offline-calibration-based allocation method, under the original fixed-budget setting used by HeadKV. Table~\ref{tab:appendix_headkv_empirical} reports the cross-task performance on LongBench. LU-KV improves the average score for both SnapKV and EA, indicating that metric-specific global allocation provides additional benefit under the same fixed-budget evaluation setting.

\begin{table*}[h]
\centering
\caption{Cross-task comparison with HeadKV on LongBench under the original HeadKV fixed-budget setting.}
\label{tab:appendix_headkv_empirical}
% \begin{camblueblock}
\setlength{\tabcolsep}{3pt}

\begin{tabularx}{\textwidth}{@{}lYYYYYYY@{}}
\toprule
Method & Avg. & Single-Doc QA & Multi-Doc QA & Summ. & Few-shot & Synthetic & Code \\
\midrule
Full Cache & 49.29 & 43.43 & 46.33 & 28.91 & 69.53 & 53.99 & 58.01 \\
HeadKV (SnapKV) & 45.27 & 35.46 & 40.72 & 25.55 & 64.08 & 51.97 & 61.45 \\
LU-KV (SnapKV) & 46.37 & 37.99 & \textbf{42.57} & 26.45 & \textbf{65.02} & 51.04 & \textbf{61.87} \\
HeadKV (EA) & 45.34 & 37.79 & 39.74 & 26.88 & 63.53 & 51.09 & 59.70 \\
LU-KV (EA) & \textbf{47.12} & \textbf{41.32} & 41.75 & \textbf{27.34} & 64.77 & \textbf{53.23} & 60.99 \\
\bottomrule
\end{tabularx}
% \end{camblueblock}
\end{table*}

\subsection{Integration with More KV Compression Metrics}
We further evaluate whether LU-KV can be applied beyond the SnapKV and KeyDiff metrics used in the main experiments. Table~\ref{tab:appendix_kvzip_cake} integrates LU-KV with KVZip and CAKE by keeping their underlying compression metrics while replacing the default budget allocation with LU-KV's global allocation. This separates the effect of the compression metric from the effect of budget allocation. Across both KVZip and CAKE, LU-KV improves the average score under the same compression ratio, with larger gains under more aggressive compression.

Table~\ref{tab:appendix_metric_ablation} summarizes the metric-level ablation across KVZip, CAKE, and EA. The purpose is to test whether LU-KV depends on a particular proxy metric, or whether it can improve different metrics once their offline utility profiles are available. The consistent gains across all three metrics suggest that LU-KV acts as a general allocation layer: it does not replace the underlying metric, but improves how the global KV budget is distributed for that metric.

\begin{table*}[h]
\centering
\caption{Integration with KVZip and CAKE on LongBench. The rows labeled LU-KV apply LU-KV's allocation to the corresponding metric.}
\label{tab:appendix_kvzip_cake}
% \begin{camblueblock}
\setlength{\tabcolsep}{3pt}
% \begin{tabularx}{\textwidth}{@{}llYYYYYYY@{}}
% \toprule
% Method & Ratio & Avg. ($\Delta$) & Single-Doc QA & Multi-Doc QA & Summ. & Few-shot & Synthetic & Code \\
% \midrule
% Full Cache & - & 49.29 & 43.43 & 46.33 & 28.91 & 69.53 & 53.99 & 58.01 \\
% KVZip & 0.5 & 48.90 ($\downarrow$0.79\%) & 43.43 & 43.58 & 28.92 & 69.19 & 51.80 & 61.71 \\
% LU-KV (KVZip) & 0.5 & \textbf{49.16} ($\downarrow$0.26\%) & 43.09 & 43.76 & 28.98 & 69.73 & 52.78 & 62.16 \\
% KVZip & 0.8 & 47.41 ($\downarrow$3.80\%) & 43.77 & 41.41 & 28.48 & 66.55 & 49.21 & 59.80 \\
% LU-KV (KVZip) & 0.8 & \textbf{48.81} ($\downarrow$0.98\%) & 43.31 & 42.57 & 28.67 & 68.58 & 52.50 & 63.25 \\
% KVZip & 0.9 & 40.84 ($\downarrow$17.15\%) & 38.41 & 39.53 & 27.01 & 62.32 & 24.15 & 51.65 \\
% LU-KV (KVZip) & 0.9 & \textbf{44.79} ($\downarrow$9.13\%) & 40.92 & 38.33 & 27.16 & 64.31 & 40.91 & 61.31 \\
% \midrule
% CAKE & 0.5 & 47.24 ($\downarrow$4.15\%) & 42.33 & 41.17 & 27.12 & 66.71 & 53.14 & 58.83 \\
% LU-KV (CAKE) & 0.5 & \textbf{49.73} ($\uparrow$0.89\%) & 43.38 & 46.34 & 28.66 & 69.87 & 54.48 & 60.97 \\
% CAKE & 0.8 & 43.56 ($\downarrow$11.62\%) & 33.29 & 38.53 & 24.57 & 62.07 & 52.32 & 58.49 \\
% LU-KV (CAKE) & 0.8 & \textbf{47.86} ($\downarrow$2.91\%) & 42.94 & 45.78 & 27.29 & 65.16 & 52.97 & 58.12 \\
% CAKE & 0.9 & 39.93 ($\downarrow$18.99\%) & 25.39 & 34.26 & 22.61 & 60.09 & 47.55 & 58.38 \\
% LU-KV (CAKE) & 0.9 & \textbf{45.36} ($\downarrow$7.96\%) & 36.88 & 43.98 & 24.49 & 62.90 & 52.96 & 57.58 \\
% \bottomrule
% \end{tabularx}

\begin{tabularx}{\textwidth}{@{}llYYYYYYY@{}}
\toprule
Method & Ratio & Avg. & Single-Doc QA & Multi-Doc QA & Summ. & Few-shot & Synthetic & Code \\
\midrule
Full Cache & - & 49.29 & 43.43 & 46.33 & 28.91 & 69.53 & 53.99 & 58.01 \\
KVZip & 0.5 & 48.90 & 43.43 & 43.58 & 28.92 & 69.19 & 51.80 & 61.71 \\
LU-KV (KVZip) & 0.5 & \textbf{49.16} & 43.09 & 43.76 & 28.98 & 69.73 & 52.78 & 62.16 \\
KVZip & 0.8 & 47.41 & 43.77 & 41.41 & 28.48 & 66.55 & 49.21 & 59.80 \\
LU-KV (KVZip) & 0.8 & \textbf{48.81} & 43.31 & 42.57 & 28.67 & 68.58 & 52.50 & 63.25 \\
KVZip & 0.9 & 40.84 & 38.41 & 39.53 & 27.01 & 62.32 & 24.15 & 51.65 \\
LU-KV (KVZip) & 0.9 & \textbf{44.79} & 40.92 & 38.33 & 27.16 & 64.31 & 40.91 & 61.31 \\
\midrule
CAKE & 0.5 & 47.24 & 42.33 & 41.17 & 27.12 & 66.71 & 53.14 & 58.83 \\
LU-KV (CAKE) & 0.5 & \textbf{49.73} & 43.38 & 46.34 & 28.66 & 69.87 & 54.48 & 60.97 \\
CAKE & 0.8 & 43.56 & 33.29 & 38.53 & 24.57 & 62.07 & 52.32 & 58.49 \\
LU-KV (CAKE) & 0.8 & \textbf{47.86} & 42.94 & 45.78 & 27.29 & 65.16 & 52.97 & 58.12 \\
CAKE & 0.9 & 39.93 & 25.39 & 34.26 & 22.61 & 60.09 & 47.55 & 58.38 \\
LU-KV (CAKE) & 0.9 & \textbf{45.36} & 36.88 & 43.98 & 24.49 & 62.90 & 52.96 & 57.58 \\
\bottomrule
\end{tabularx}
% \end{camblueblock}
\end{table*}

\begin{table*}[t]
\centering
\caption{Metric ablation analysis on LongBench using LU-KV.}
\label{tab:appendix_metric_ablation}
% \begin{camblueblock}
\begin{tabular}{lccc}
\toprule
Metric & Compression Ratio & Original Avg. & w/ LU-KV \\
\midrule
Full Cache & - & 49.29 & - \\
\midrule
KVZip & 0.5 & 48.90 & \textbf{49.16} \\
KVZip & 0.8 & 47.41 & \textbf{48.81} \\
KVZip & 0.9 & 40.84 & \textbf{44.79} \\
\midrule
CAKE & 0.5 & 47.24 & \textbf{49.73} \\
CAKE & 0.8 & 43.56 & \textbf{47.86} \\
CAKE & 0.9 & 39.93 & \textbf{45.36} \\
\midrule
EA & 0.5 & 48.35 & \textbf{49.37} \\
EA & 0.8 & 43.10 & \textbf{48.73} \\
EA & 0.9 & 36.58 & \textbf{46.28} \\
\bottomrule
\end{tabular}
% \end{camblueblock}
\end{table*}

\end{document}